\documentclass[preprint,12pt]{article}
\usepackage{MyDefinitions}
\usepackage{multirow}
\usepackage{amsmath}
\usepackage{amssymb}

\usepackage{graphicx} 

\newtheorem{example}{Example}
\newtheorem{theorem}{Theorem}
\newtheorem{definition}{Definition}
\newtheorem{remark}{Remark}
\title{A binarized-domains arc-consistency algorithm for $\tcsps$: its computational analysis and its use as a filtering procedure in solution search algorithms}

\author{Amar Isli\\
            University of Sciences and Technology Houari Boumedi\`ene\\
            Faculty of Computer Science\\
            Department of Artificial Intelligence and Data Science\\
            BP 32, Bab Ezzouar\\
            DZ-16111 Algiers\\
           Algeria\\
            E-mail: a\_isli@yahoo.com}

\begin{document}
\maketitle
\begin{abstract}
$\tcsps$ (Temporal Constraint Satisfaction Problems), as defined in \cite{DechterMP91a}, get rid of unary constraints
by binarizing them after having added an "origin of the world" variable. This makes the constraints exclusively
binary; additionally, a $\tcsp$ verifies the properties of node- and arc-consistencies, equivalent, respectively, to strong 1- and 2-consistencies. Path-consistency, equivalent to strong 3-consistency,
the next higher local consistency, solves the consistency problem of a convex $\tcsp$, referred to in
\cite{DechterMP91a} as an $\stp$ (Simple Temporal Problem); more than that, the output of path-consistency applied
to an $n+1$-variable $\stp$ is a minimal and strongly $n+1$-consistent $\stp$. Weaker versions of path-consistency,
aimed at avoiding what is referred to in \cite{SchwalbD97a} as the "fragmentation problem", are used as filtering
procedures in recursive backtracking algorithms for the consistency problem of a general $\tcsp$. In this work, we look
at the constraints between the "origin of the world" variable and the other variables, as the (binarized) domains of
these other variables. With this in mind, we define a notion of arc-consistency for $\tcsps$, which we will refer to as
binarized-domains Arc-Consistency, or bdArc-Consistency for short. We provide an algorithm achieving bdArc-Consistency
for a $\tcsp$, which we will refer to as \mbox{bdAC-3}, for it is an adaptation of Mackworth's \cite{Mackworth77a}
well-known arc-consistency algorithm \mbox{AC-3}.
We show that if an $\stp$ is bdArc-Consistent, and connected, i.e., its "origin of the world" variable is disconnected from
none of the other variables, its binarized domains are minimal. We provide two polynomial backtrack-free procedures: one
for the task of getting a solution from a connected bdArc-Consistent $\stp$; the other for the task of
getting, from a bdArc-Consistent $\stp$, either that it is inconsistent or, in case of consistency, a connected bdArc-Consistent
           $\stp$ refinement.
We then show how to use our results both in a general $\tcsp$ solver and in a $\tcsp$-based job shop scheduler. In particular,
we show a new result on a known class of $\tcsps$, which will make it possible to use \mbox{bdAC-3} as the filtering procedure
of the scheduler, with no risk of making its branching factor worsened by the fragmentation problem. From our work can be
extracted a one-to-all all-to-one shortest paths algorithm of an $\BBR$-labeled directed graph,
as a generalization of Bellman-Ford-Moore's algorithm \cite{Bellman58a,FordF62a,Moore59a}.
We provide an adaptation to $\tcsps$ of the other three local consistency algorithms in \cite{Mackworth77a}: the
arc-consistency algorithm \mbox{AC-1}, and the path-consistency algorithms \mbox{PC-1} and \mbox{PC-2}; and show that
an existing adaptation to $\tcsps$ of \mbox{PC-2} \cite{DechterMP91a,SchwalbD97a} is not correct, as it does not always
terminate. The work also provides an experimental comparison on $\stps$ of \mbox{bdAC-3} with an existing arc-consistency
algorithm, $\acstp$, restricted to $\stps$ \cite{KongLL18a}; an experimental comparison of three $\tcsp$-based job shop
schedulers, two of which use weak versions of \mbox{bdAC-3} as the filtering procedure during the search, the third \cite{SchwalbD97a}
a weak version of path-consistency; and the swi-prolog source codes used by these comparisons. Last but not least, we provide an incremental version of \mbox{bdAC-3}.
\end{abstract}







%
\section{Introduction}\label{szero}
Constraint-based temporal reasoning, or CTR, came as a generalization of discrete constraint satisfaction problems, or discrete CSPs
\cite{Montanari74a,Mackworth77a,Freuder82a}, to continuous variables. Qualitative approaches to CTR include Allen's
\cite{Allen83a} highly influential Interval Algebra, and quantitative approaches to it include the equally influential framework of
Temporal Constraint Satisfaction Problems, or $\tcsps$ \cite{DechterMP91a}. The main research issues
dealt with in CTR include the following two:
\begin{enumerate}
  \item extraction of tractable subclasses, for which polynomial algorithms, incomplete for the general, subsuming
language, become complete \cite{Bessiere96a,BessiereIL96a,BodirskyJ17a,BroxvallJR02a,Isli01a,Ligozat96a,NebelB95a,vanBeek92a}; and
  \item design of complete algorithms, with an exponential worst-case computational complexity, but with a nice computational behavior on
           known classes of real-life problems \cite{LadkinM94a,LadkinR92a,Nebel97a,SchwalbD97a,StergiouK00a,vanBeek92a,vanBeekM96a}.
\end{enumerate}

The constraints of a Temporal Constraint Satisfaction Problem (TCSP) \cite{DechterMP91a} were first defined as
being either unary or binary. Then, in the same article, the definition was modified through the addition of
a new variable referred to as the ''origin of the world'' variable, which is used to binarize the unary constraints. This
addition of an ''origin of the world'' variable makes the constraints of a $\tcsp$ exclusively binary, with the consequence
that a $\tcsp$ verifies then the properties of node- and arc-consistencies.

While the adding of an ''origin of the world'' variable
has the advantage of transforming the solving of an $\stp$, i.e., a $\tcsp$ whose constraints are convex, into a shortest
paths problem, it remains that, because a $\tcsp$ then becomes node- and arc-consistent, it gets easy for one's attention to immediately
skip to the next higher local consistency, path-consistency, which is exactly what seems to have happened in \cite{DechterMP91a}.

In this work, while we will keep the idea of having an ''origin of the world'' variable, and of binarizing the unary constraints,
making the constraints of a $\tcsp$ exclusively binary, we will look
at the constraints between the "origin of the world" variable and the other variables, as the (binarized) domains of
these other variables. With this in mind:
\begin{enumerate}
  \item we define a notion of arc-consistency for $\tcsps$, which we will refer to as
binarized-domains Arc-Consistency, or bdArc-Consistency for short;
  \item we provide an algorithm achieving bdArc-Consistency
for a $\tcsp$, which we will refer to as \mbox{bdAC-3}, for it is an adaptation of Mackworth's \cite{Mackworth77a}
well-known arc-consistency algorithm \mbox{AC-3};
  \item we show that if an $\stp$ is bdArc-Consistent, and connected, i.e., its "origin of the
world" variable is disconnected from none of the other variables, its binarized domains are minimal;
  \item we provide two polynomial backtrack-free procedures:
    \begin{enumerate}
      \item one for the task of getting a solution from a connected bdArc-Consistent $\stp$;
      \item the other for the task of getting, from a bdArc-Consistent $\stp$, either that it is inconsistent or, in case of consistency, a
	connected bdArc-Consistent $\stp$ refinement;
    \end{enumerate}
  \item we then show how to use bdArc-Consistency and weak versions of it as filtering procedures of a general $\tcsp$ solver and
            of a $\tcsp$-based job shop scheduler;
  \item we also provide experimental evaluations:
    \begin{enumerate}
      \item an experimental comparison on randomly gererated $\stps$ of \mbox{bdAC-3} with an existing arc-consistency
                algorithm, $\acstp$, restricted to $\stps$ \cite{KongLL18a}. The swi-prolog source code is included as part of
                the work: cmp-bdAC1-bdAC3-gSTPs.
       \item an experimental comparison of three $\tcsp$-based job shop schedulers, two of which use weak versions of
                \mbox{bdAC-3} (\mbox{wbdAC-3}, weak \mbox{bdAC-3}, and \mbox{lbdAC-3}, loose \mbox{bdAC-3}) as the
                 filtering procedure during the search, the third \cite{SchwalbD97a} a weak version of the path-consistency
                algorithm \mbox{PC-2}, \mbox{lPC-2} (loose \mbox{PC-2}). The one using  \mbox{lbdAC-3} performs better
                than each of the other two. We chose \mbox{lPC-2} because empirical comparisons from the literature
               [Schwalb and Dechter 1997] show that it is the best version of path consistency to use as the filtering procedure
               of a $\tcsp$. The swi-prolog source codes of the three schedulers are also included as part of the work:
               js-LA-wbdac3, js-LA-lbdac3 and js-LA-lpc2.
      \item The four swi-prolog source codes added as part of the work include explaining comments, expected to make
               them easily readable. In particular, the source codes can be used to test the three solvers on machines faster than
               the one used in our experiments. This is expected to be a convincing argument for the extension of pure Logic
               Programming to Constraint Logic Programming over Temporal Domains, as it once was extended to Constraint Logic
               Programming over Finite Domain. Furthermore, available spatial systems and tool boxes implementing existing
               qualitative spatial calculi, such as the toolbox \mbox{SparQ} in \cite{DyllaFWW06a,WallgrunFWDF06a} and the
               \mbox{RCC-8} query answering system in \cite{BennettIC98a}, make it possible to have even better extensions
               of pure Logic Programming, such as Constraint Logic Programming over Spatial and Temporal Domains.
    \end{enumerate}
  \item we provide an incremental version of \mbox{bdAC-3}, suggested by the importance of constraint propagation and real-time
            maintenance of local consistency such as bdArc-consistency for applications such as planning \cite{GhallabA89,GhallabL94a,PlankenWK08a,PlankenWY10a,XuC03a},
            for which it is generally not possible to have the constraints all at once, which arrive rather as packs separated in time. Real-time maintenance
            will allow not to postpone the treatment of a newly arrived pack, which is propagated, until bdArc-consistency is reached again, 
            in real time, as it arrives, without waiting for the arrival of future packs, the whole process, thanks to incrementality, without having to redo the work done for the packs
            arrived before it.
\end{enumerate}
Additionally:
\begin{enumerate}
  \item we extract from our work a one-to-all all-to-one shortest paths algorithm of an $\BBR$-labeled directed graph,
as a generalization of Bellman-Ford-Moore's algorithm \cite{Bellman58a,FordF62a,Moore59a};
  \item we provide an adaptation to $\tcsps$ of the other three local consistency algorithms in \cite{Mackworth77a}: the
arc-consistency algorithm \mbox{AC-1}, and the path-consistency algorithms \mbox{PC-1} and \mbox{PC-2}; and
  \item we show that
an existing adaptation to $\tcsps$ of \mbox{PC-2} \cite{DechterMP91a,SchwalbD97a} is not correct, as it does not always
terminate.
\end{enumerate}
\subsection{Relevance to Artificial Intelligence}
Research topics of Constraint Satisfaction, a well-established area of Artificial Intelligence, include local consistency algorithms and their use as
a filtering procedure in solution search algorithms. The work fits in this topic, as it defines a new local consistency algorithm for TCSPs, namely, a
binarized-domains arc-consistency algorithm; and gives a thorough look at its use as a filtering procedure in solution search algorihms in general,
and in TCSP-based schedulers in particular. Research on discrete CSPs' arc-consistency has had so far no challenging ideas coming from TCSPs,
because the best interesting strong k-consistency known for these was the factor(s) computationally more expensive path-consistency; it will now
have to compete with research on its $\tcsps$' counterpart, binarized-domains arc-consistency.

Another point worth mentioning concerning the accessibility of the work to a wide audience is that, as explained in Section \ref{scheduler}, which
provides a TCSP-based scheduler, there are many approaches to scheduling other than TCSP-based approaches. Our work gives these other
approaches another TCSP-based scheduler they will have to compete with.
\subsection{Outline of the paper}
The rest of the paper is organized as follows. Section \ref{motivation} motivates the work. Section \ref{sone} is devoted to needed background.
Section \ref{stwo} defines our notion of binarized-domains Arc-Consistency (bdArc-Consistency) for $\tcsps$; provides our algorithm \mbox{bdAC-3}
achieving bdArc-Consistency for a general $\tcsp$,  together with a proof of its termination and its computational complexity; shows the minimality result
alluded to above; and ends with an important corollary providing a one-to-all all-to-one shortest paths algorithm of an $\BBR$-labeled directed
graph, as a generalization of Bellman-Ford-Moore's algorithm \cite{Bellman58a,FordF62a,Moore59a}. Sections \ref{bfpone} and \ref{bfptwo}
provide the two polynomial backtrack-free procedures referred to above. Section \ref{sthree} provides a general $\tcsp$ solver using a weak
version of bdArc-Consistency, wbdArc-Consistency, as the filtering procedure during the search, whose correctness is a direct consequence of our
minimality result of Section \ref{stwo}. A $\tcsp$-based job shop scheduler, which is an adaptation of the general $\tcsp$ solver, is described in
Section \ref{scheduler}. Subsection \ref{schedulerSubs1} describes the use in such a scheduler of \mbox{wbdAC-3} as the filtering procedure;
subsection \ref{schedulerSubs2} shows a new result on a known class of $\tcsps$, guaranteeing that \mbox{bdAC-3} can be used as the
filtering procedure in such a scheduler, with no risk of making its branching factor worsened by the fragmentation problem.  Section \ref{relatedwork}
discusses of the most related literature; provides an adaptation to $\tcsps$ of the other three local consistency algorithms in \cite{Mackworth77a}:
the arc-consistency algorithm \mbox{AC-1}, and the path-consistency algorithms \mbox{PC-1} and \mbox{PC-2}; studies the termination of these
adaptations; and shows that an existing adaptation to $\tcsps$ of \mbox{PC-2} \cite{DechterMP91a,SchwalbD97a} does not always terminate.
Sections \ref{expcmpstps} and \ref{expcmpjss} provide the experimental comparisons referred to above. Section \ref{incv} describes the incremental
version of \mbox{bdAC-3}. Section \ref{sfive} summarizes the work and provides some of its possible future directions.
\section{Main motivation}\label{motivation}
We first motivate our interest in binarized-domains arc-consistency.
\subsection{\mbox{bdAC-3}, the $\tcsps$' counterpart of discrete CSPs' \mbox{AC-3}}
The invariance of $\tcsps$ by classical node- and arc-consistencies makes path-consistency, equivalent to
strong \mbox{3-consistency}, the smallest interesting strong \mbox{k-consistency} \cite{Freuder82a}. This situation had
been made possible by the fact that Dechter et al. \cite{DechterMP91a} made no distinction between, on one hand, the constraints
between the ''origin of the world'' variable and the other variables and, on the other, the other constraints.
The invariance of $\tcsps$ by \mbox{arc-consistency}, making the latter useless, is in high contrast with the well-known importance
of arc-consistency for discrete CSPs, for which it is the local consistency the most used as the filtering procedure of solution search
algorithms. This, together with the factor(s) higher computational cost of path-consistency, which $\tcsps$' solvers have been using so far as
the filtering procedure during the search, is certainly what favored the use of discrete CSPs’ solvers, with \mbox{arc-consistency} as a filtering
procedure. This work is expected to reverse the situation and make $\tcsps$ give the importance they deserve to
\mbox{arc-consistency} and to solution search algorithms using it as the filtering procedure, thanks to the
binarized-domains arc-consistency algorithm to be defined. In particular, we expect TCSP-based schedulers with binarized-domains arc-consistecy
as the filtering procedure, such as the one this work provides, to compete with, or even challenge, their discrete CSP-based
counterparts, with classical arc-consistency as the filtering procedure during the search. This expectation is confirmed by our experimental
comparison of Section \ref{expcmpjss} of $\tcsp$-based job shop solvers, the main result of which is as follows:
\begin{enumerate}
  \item the best weak version of \mbox{bdAC-3} to use as the filtering procedure of a $\tcsp$-based job shop scheduler is \mbox{lbdAC-3},
            loose \mbox{bdAC-3};
  \item \mbox{lbdAC-3} is advantageously compared to the best weak version of path consistency, \mbox{lPC-2} (loose
            \mbox{PC-2}), to use as the filtering procedure of a $\tcsp$ solver \cite{SchwalbD97a}.
\end{enumerate}
This main result of the experimental comparison is expected to attract interest in improving the bdArc Consistency algorithm \mbox{bdAC-3}
to be defined, the goal being to keep alive the process of improving the efficiency of the search algorithms using it as the filtering procedure.
This is well-known with discrete CSPs: many improved versions of the arc-consistency algorithm \mbox{AC-3} \cite{Mackworth77a} exist, each
coming as an improvement of its predecessor. 
\subsection{Constraint Logic Programming over Spatial and Temporal Domains}
Constraint Logic Programming over Finite Domains, or \mbox{CLP(FD)} for short, extends pure Logic Programming with algorithmic tools for
the handling of discrete \mbox{CSPs}. It is particularly useful for the solving of combinatorial problems such as scheduling problems and planning
problems.

The $\tcsp$ tools provided in this work, that include the bdArc Consistency algorithm \mbox{bdAC-3}, a general $\tcsp$ solver using \mbox{bdAC-3} as
the filtering procedure, and a job shop scheduler using also \mbox{bdAC-3} as the filtering procedure, pave the way for another extension of pure Logic
Programming, Constraint Logic Programming over Temporal Domains. This would, among other things, offer another way of tackling and solving
combinatorial problems, and open Logic Programming to the world of continuous and temporal domains, of which we know that it has
not delivered all its secrets yet. Such an extension of pure Logic Programming would also serve as a platform for real-time systems and applications,
for which the handling of temporal constraints is crucial.

Even better, if we add the spatial systems and tool boxes available in the literature, such as the toolbox \mbox{SparQ} in \cite{DyllaFWW06a,WallgrunFWDF06a}
or the \mbox{RCC-8} query answering system in \cite{BennettIC98a}, that include
implementations of known qualitative spatial calculi \cite{EgenhoferF91a,Frank92a,Freksa92a,IsliC00a,Ligozat98a,RandellCuiCohn92a}, even better extensions
of pure Logic Programming are within reach, such as Constraint Logic Programming over Spatial and Temporal Domains.
\section{Temporal Constraint Satisfaction Problems}\label{sone}
Temporal Constraint Satisfaction Problems, or $\tcsps$ for short, have been proposed in \cite{DechterMP91a} as an extension of
(discrete) CSPs \cite{Mackworth77a,Montanari74a} to continuous variables.
\begin{definition}[$\tcsp$ \cite{DechterMP91a}]
A $\tcsp$ is a pair $P=(X,C)$ consisting of:
\begin{enumerate}
  \item a finite set $X$ of $n$ variables, $X_1,\ldots ,X_n$, ranging over the
universe of time points; and
  \item a finite set $C$ of Dechter, Meiri and Pearl's
constraints (henceforth $\dmp$ constraints) on the variables.
\end{enumerate}
\end{definition}
A $\dmp$ constraint is either unary or binary. A unary constraint has
the form $X_i\in C_i$, and a binary constraint the form $(X_j-X_i)\in C_{ij}$, where $C_i$ and $C_{ij}$ are subsets of the set $\BBR$ of real
numbers, and $X_i$ and $X_j$ are variables ranging over the set $\BBR$ seen as the universe of time points. A unary constraint $X_i\in C_i$
may be seen as a special binary constraint if we consider an origin of the World (time $0$), represented by a variable $X_0$: $X_i\in C_i$ is
then equivalent to $(X_i-X_0)\in C_{0i}$,with $C_{0i}=C_i$. Unless explicitly stated otherwise, we assume, in the rest of the paper, that the
constraints of a $\tcsp$ are all binary. Furthermore, without loss of generality, we make the assumption that all constraints $(X_j-X_i)\in C_{ij}$
of a $\tcsp$ are such that $i<j$: if this is not the case for a constraint $(X_j-X_i)\in C_{ij}$, we replace it with the equivalent constraint
$(X_i-X_j)\in C_{ij}^\smile$, where $C_{ij}^\smile$ is the converse of $C_{ij}$ (see the definition of converse below).
\begin{definition}[$\stp$ \cite{DechterMP91a}]
An $\stp$ (Simple Temporal Problem) is a $\tcsp$ of which all the
constraints are convex, i.e., of the form $(X_j-X_i)\in C_{ij}$, $C_{ij}$ being a convex 
subset of $\BBR$.
\end{definition}
A universal constraint for $\tcsps$ in general, and for $\stps$ in
particular, is of the form $(X_j-X_i)\in\BBR$, and is equivalent to
``no knowledge'' on the difference $(X_j-X_i)$. An equality
constraint is of the form $(X_j-X_i)\in\{0\}$: it ``forces" variables $X_i$ and $X_j$ to be equal.

We now consider an $n+1$-variable $\tcsp$ $P=(X,C)$, with $X=\{X_0,X_1,$ $\ldots ,X_n\}$, the variable
$X_0$ representing the origin of the World. Without loss of generality, we assume that $P$ has at most one
constraint per pair of variables: if $C$ contains $p$ constraints $(X_j-X_i)\in C_{ij_1}$, ..., $(X_j-X_i)\in C_{ij_p}$
on the same pair $(X_i,X_j)$ of variables, with $p\geq 2$, these can be replaced with the unique constraint
$(X_j-X_i)\in \bigcap _{k=1}^pC_{ij_k}$ and the resulting $\tcsp$ remains equivalent to the original one.

\begin{definition}[network representation]
The network representathon of $P$ is the weighted directed graph of which the
vertices are the variables of $P$, and the edges are the pairs $(X_i,X_j)$ of variables on
which a constraint $(X_j-X_i)\in C_{ij}$ is specified. The weight of edge $(X_i,X_j)$ is the set $C_{ij}$
such that $(X_j-X_i)\in C_{ij}$ is the constraint of $P$ on the pair $(X_i,X_j)$ of
variables.
\end{definition}

\begin{definition}[matrix representation]
The matrix representation
of $P$ is the $(n+1)\times (n+1)$-matrix, denoted by $P$ for simplicity,
defined as follows:
\begin{enumerate}
  \item $P_{ii}=\{0\}$, $\forall i=0\ldots n$;
  \item $P_{ij}=C_{ij}$ and $P_{ji}=\{a\mbox{ s.t. }(-a)\in C_{ij}\}$, for all $i\neq j$ such that a
           constraint $(X_j-X_i)\in C_{ij}$ is specified on $X_i$ and $X_j$;
  \item $P_{ij}=\BBR$, for all other pairs $(i,j)$.
\end{enumerate}
\end{definition}

\begin{definition}[(consistent) instantiation]
An instantiation of $P$ is any n+1-tuple $(x_0,x_1,\ldots ,x_n)\in\BBR ^{n+1}$, representing an assignment of a value to
each variable. A consistent instantiation, or solution, is an instantiation satisfying all the
constraints: for all $i,j$, $(X_i,X_j)=(x_i,x_j)$
satisfies the constraint, if any, specified on the pair $(X_i,X_j)$.
\end{definition}
\begin{definition}[subnetwork]
A $k$-variable subnetwork, $k\leq n+1$, is any
restriction of the network $P$ to $k$ of its variables and the
constraints on pairs of those $k$ variables.
\end{definition}

\begin{definition}[(strong) $k$-consistency]
For all $k=1\ldots (n+1)$, $P$ is $k$-consistent if any solution to any $(k-1)$-variable subnetwork extends to any
$k$-th variable; it is strongly
$k$-consistent if it is $j$-consistent, for all $j\leq k$.
\end{definition}

Strong 1-, 2- and 3-consistencies correspond to \mbox{node-,} arc- and path-consistencies,
respectively \cite{Montanari74a,Mackworth77a}. Strong $(n+1)$-consistency of $P$ corresponds to
decomposability \cite{Dechter92a}. It facilitates the exhibition of a
solution by backtrack-free search \cite{Freuder82a}.

The consistency problem of a TCSP, i.e., the problem of verifying whether it has
a consistent instantiation, is NP-hard. Davis \cite{Davis89a} (cited in \cite{DechterMP91a})  showed that even the subclass
of TCSPs in which the constraints are of length lower than or equal to 2 (i.e.,
they are of the form $(X_j-X_i)\in C_{ij}$, with
$C_{ij}$ being a convex set or a union of two disjoint convex sets) is \mbox{NP-hard} (see also \cite{DechterMP91a}, Theorem 4.1,
Page 73). However, when the constraints of a TCSP are all convex, i.e., when we restrict ourselves to STPs, the consistency problem
is polynomial \cite{DechterMP91a}. Moreover, in the case of STPs the classical path-consistency method \cite{Montanari74a,Mackworth77a}
leads to strong $(n+1)$-consisteny \cite{DechterMP91a}.
\begin{definition}[converse]
The converse of a DMP constraint $(X_j-X_i)\in C_{ij}$ is the equivalent DMP
constraint $(X_i-X_j)\in C_{ij}^\smile$:
$(X_j-X_i)\in C_{ij}\Leftrightarrow (X_i-X_j)\in C_{ij}^\smile$.Clearly, $C_{ij}^\smile =\{-a\mbox{ such that }a\in C_{ij}\}$.
\end{definition}
\begin{definition}[intersection]
The intersection of two DMP constraints $(X_j-X_i)\in C_{ij}^1$ and $(X_j-X_i)\in C_{ij}^2$, on the same
pair of variables, is the DMP constraint $(X_j-X_i)\in C_{ij}$, with $C_{ij}=C_{ij}^1\cap C_{ij}^2$.
\end{definition}
\begin{definition}[composition]
The composition of two DMP
constraints $(X_k-X_i)\in C_{ik}$ and $(X_j-X_k)\in C_{kj}$, written $(X_k-X_i)\in C_{ik}\otimes (X_j-X_k)\in C_{kj}$,
is the constraint $(X_j-X_i)\in C_{ikj}$, on the extreme variables $X_i$ and $X_j$, such that
$C_{ikj}=\{c:\mbox{ }\exists a\in C_{ik},\exists b\in C_{kj}\mbox{ s.t. }c=a+b\}$. We will also say that the composition
of the sets $C_{ik}$ and $C_{kj}$ is the set $C_{ikj}$: $C_{ik}\otimes C_{kj}=C_{ikj}$.
\end{definition}
\begin{definition}[Convex closure]The convex closure of a set $A\subseteq\BBR$, $cc(A)$, is the smallest convex set verifying
$A\subseteq cc(A)$. The convex closure of a $\tcsp$ constraint $(X_j-X_i)\in C_{ij}$ is the $\tcsp$ constraint $(X_j-X_i)\in cc(C_{ij})$.
The convex closure of a $\tcsp$ $P=(X,C)$, $cc(P)$, is the $\stp$ $cc(P)=(X,C')$, with
$C'=\{(X_j-X_i)\in cc(C_{ij}):\mbox{ }(X_j-X_i)\in C_{ij}\mbox{ constraint of $P$}\}$.
\end{definition}
\begin{definition}[weak composition \cite{SchwalbD97a}] The weak composition of
two DMP constraints $(X_k-X_i)\in C_{ik}$ and $(X_j-X_k)\in C_{kj}$, written
$(X_k-X_i)\in C_{ik}\otimes _w(X_j-X_k)\in C_{kj}$, is the composition of their
convex closures; i.e., $(X_k-X_i)\in C_{ik}\otimes _w(X_j-X_k)\in C_{kj}$ is equal
to $(X_k-X_i)\in cc(C_{ik})\otimes (X_j-X_k)\in cc(C_{kj})$.
\end{definition}
\begin{definition}[Minimal partition into convex subsets]Let $S$ be a subet of $\BBR$: $S\subseteq\BBR$. The minimal partition of
$S$ into convex subsets, or $\mbox{mPcs}(S)$ for short, is the partition $\mbox{mPcs}(S)=\{S_1,\ldots ,S_p\}$ of $S$  into
maximal convex subsets; in other words, $\mbox{mPcs}(S)$ is such that:
\begin{enumerate}
  \item $S=S_1\cup\ldots\cup S_p$
  \item for all $i\in\{1,\ldots ,p\}$:
  \begin{enumerate}
    \item $S_i$ is convex, and
    \item for all convex set $S'$, if $S_i\subset S'$ then $S'$ is  not a subset of $S$.
  \end{enumerate}
\end{enumerate}
\end{definition}
Applying PC even to a $\tcsp$ whose constraints are initially of length lower than or equal to 2, may lead to the so-called
``fragmentation problem'' \cite{SchwalbD97a}: the constraints may be ``fragmented''. Schwalb and Dechter \cite{SchwalbD97a}
gave a number of weak versions of PC, which are less effective but have the advantage of getting rid of the fragmentation problem.
One of these versions is ULT (Upper-Lower Tightening), which we will also refer to as weak Path-Consistency, or wPC for
short. It consists of Path-Consistency (PC) in which the usual composition operation is replaced with weak composision.
\begin{definition}[loose intersection \cite{SchwalbD97a}]  Loose intersection of two sets $T$ and $S$, $T\vartriangleleft S$, with $T$ being
the union of disjoint convex sets $I_1,\ldots , I_r$, is the union of sets $J_1,\ldots ,J_r$ with $J_k$ ($k$ from $1$ to $r$) equal to the convex
closure of $I_k\cap S$.
\end{definition}
We refer to the operation $C_{ij}:=C_{ij}\cap C_{ik}\otimes C_{kj}$ (respectively, $C_{ij}:=C_{ij}\cap C_{ik}\otimes _wC_{kj}$ and $C_{ij}:=C_{ij}\vartriangleleft C_{ik}\otimes C_{kj}$)
as the path-consistency (respectively, the weak path-consistency and  the loose path-consistency) operation on triangle $(X_i,X_k,X_j)$.
Applying path consistency (respectively, weak path-consistency and loose path-consistency) to a TCSP consists of repeating, until stability is reached or
inconsistency is detected, the process of applying the path-consistency (respectively, the weak path-consistency and the loose path-consistency) operation
to each triangle $(X_i,X_k,X_j)$ of variables. We will also refer to each of the three operations as a relaxation step.

Example \ref{example} illustrates some of the introduced notions, including the fragmentation problem.
\begin{figure}
\centerline{\includegraphics{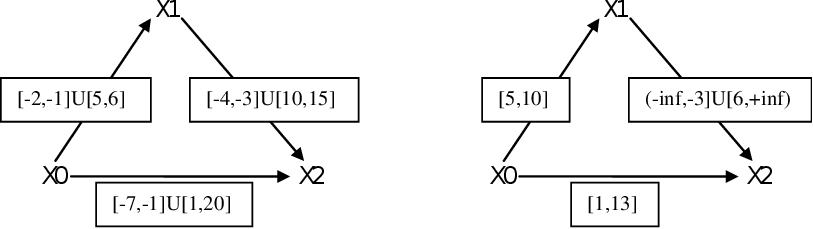}}
\caption{Illustration of the fragmentation problem: the path-consistency operation $C_{02}=C_{02}\cap C_{01}\otimes C_{12}$ fragments the weight of
	edge $(X_0,X_2)$, passing from the union of two convex sets to a union of four convex sets (left), and from one convex set to a union of two
	convex sets (right). In the figure on the right, inf stands for $\infty$ (infinity).}\label{fragprob}
\end{figure}
\begin{example}\label{example}
Let $P=(X,C)$ be a $\tcsp$, with $X=\{X_0,X_1,X_2\}$, $C=\{c_1:(X_1-X_0)\in [-2,-1]\cup [5,6],c_2:(X_2-X_0)\in [-7,-1]\cup [1,20],c_3:(X_2-X_1)\in [-4,-3]\cup [10,15]\}$,
and the network representation given by Figure \ref{fragprob}(left):
\begin{enumerate}
  \item The path-consistency operation on triangle $(X_0,X_1,X_2)$ is $C_{02}=C_{02}\cap C_{01}\otimes C_{12}$, which leads to
           $C_{02}=[-6,-4]\cup [1,3]\cup [8,14]\cup [15,20]$,  for
           $
            \begin{array}{lll}
           C_{01}\otimes C_{12}&=&[-2,-1]\otimes [-4,-3]\cup\\
                                               &&[-2,-1]\otimes [10,15]\cup \\
                                               &&[5,6]\otimes [-4,-3]\cup\\
                                               &&[5,6]\otimes [10,15]\\
                                               &=&[-6,-4]\cup [8,14]\cup [1,3]\cup [15,21]\\
                                               &=&[-6,-4]\cup [1,3]\cup [8,14]\cup [15,21].
            \end{array}
             $
  \item The path-consistency operation, therefore, fragments the weight of edge $(X_0,X_2)$, which was the union of two convex
            sets, and becomes the union of four convex sets.
  \item The convex closures of $C_{01}$ and $C_{12}$, respectively, are $cc(C_{01})=[-2,6]$ and $cc(C_{12})=[-4,15]$. Weak
            composition of $C_{01}$ and $C_{12}$ is therefore $C_{01}\otimes _wC_{12}=cc(C_{01})\otimes cc(C_{12})=[-2,6]\otimes [-4,15]=[-6,21]$.
  \item If, instead of the path-consistency operation, we used the weak path-consistency operation, we would get $C_{02}=[-6,-1]\cup [1,20]$,
           which does not fragment the weight of edge $(X_0,X_2)$.
  \item If, instead of the path-consistency operation, we used the loose path-consistency operation, we would get $C_{02}=[-6,-4]\cup [1,20]$,
           which does not fragment the weight of edge $(X_0,X_2)$ and gives a tighter constraint.
  \item\label{fragprobright} Even when we replace $C_{01}$ and $C_{02}$ with the convex sets $[5,10]$ and $[1,13]$, respectively, and $C_{12}$ with the
	union of two conex sets $(-\infty ,-3]\cup [6,+\infty )$ (see Figure \ref{fragprob}(right)), the fragmentation problem remains:
           $C_{01}\otimes C_{12}=[5,10]\otimes (-\infty ,-3]\cup [5,10]\otimes [6,+\infty )= (-\infty ,7]\cup [11,+\infty )$, and the path-consistency
	operation $C_{02}=C_{02}\cap C_{01}\otimes C_{12}$ transforms therefore $C_{02}$ from the convex set $[1,13]$ into the union of two
	convex sets $[1,7]\cup [11,13]$ .
  \item\label{fragprobright2} Finally, if $C_{12} = [-3 ,-2]\cup [6,12]$ in Figure \ref{fragprob}(right), the fragmentation problem also remains:
           $C_{01}\otimes C_{12}=[5,10]\otimes [-3,-2]\cup [5,10]\otimes [6,12]= [2,8]\cup [11,22]$, and the path-consistency
	operation $C_{02}=C_{02}\cap C_{01}\otimes C_{12}$ transforms $C_{02}$ from $[1,13]$ into $[2,8]\cup [11,13]$ . \cqfd
\end{enumerate}
\end{example}
\begin{remark}\label{remark2}
Let $(X_i,X_k,X_j)$ be a triangle subnetwork of a $\tcsp$ $P$, with $|\mbox{mPcs}(C_{ij})|=n_1$, $|\mbox{mPcs}(C_{ik})|=n_2$ and $|\mbox{mPcs}(C_{kj})|=n_3$:
\begin{enumerate}
  \item the worst-case computational complexity of the path-consistency operation, $C_{ikj}=C_{ij}\cap C_{ik}\otimes C_{kj}$, on triangle $(X_i,X_k,X_j)$  is $O(n_1+n_2n_3)=O(n_2n_3)$;
  \item the worst-case computational complexity of the weak path-consistency operation, $C_{ikj}=C_{ij}\cap C_{ik}\otimes _wC_{kj}$, on triangle $(X_i,X_k,X_j)$  is $O(n_1)$; and
  \item if $C_{ij}$, $C_{ik}$ and $C_{kj}$ are convex, the worst-case computational complexity of both operations is $O(1)$.
\end{enumerate}
See Appendix \ref{appendix1} for details. \cqfd
\end{remark}
A distance graph is a complete labeled (weighted) directed graph $G=(V,E,w)$, where $V$ is the set of vertices (nodes), $E=V\times V$ is the set of edges, and $w$ is a labeling (weight) function from
$E$ to the set $\BBR\cup\{+\infty\}$ verifying $w(V_i,V_i)=0$, for all vertices $V_i$. A rooted distance graph is a pair $(G,S)$, where $G$ is a distance graph and $S$ is a vertex of G.
A path of length $p$, or $p$-path, of a distance graph is a sequence $\pi =(V_{i_0},\cdots ,V_{i_p})$ of vertices, whose edges are all pairs $(V_{i_j},V_{i_{j+1}})$, with
$j$ from 0 to $p-1$. The length of a path is the number of its edges.

Let $\pi =(V_{i_0},\cdots ,V_{i_p})$ be a $p$-path of a distance graph $G=(V,E,w)$. $\pi$ is a circuit if $V_{i_0}= V_{i_p}$;
it is an elementary path if, for all $j,k$ such that $j<k$, if $j\not = 0$ or $k\not = p$ then $V_{i_j}\not = V_{i_k}$; it is an
elementary circuit if it is an elementary path and $V_{i_0}= V_{i_p}$. The weight function $w$ of $G$ can be extended
to paths as follows. The weight of $\pi$  is the sum $w(\pi )=\sum _{j=0}^{p-1}w(V_{i_j},V_{i_{j+1}})$. A (strictly)
negative circuit is a circuit whose weight is strictly negative.

We now define the distance graph and the rooted distance graph of an $\stp$.
\begin{definition}[(rooted) distance graph of an $\stp$]\label{rdg}             
If the $\tcsp$ $P=(X,C)$ is an $\stp$, its distance graph is the distance
graph $G=(X,X\times X,w)$ constructed as follows:
\begin{enumerate}
  \item the set of vertices of $G$ is the set $X$ of variables of $P$
  \item The labeling function $w$ is built as follows:
    \begin{enumerate}
      \item\label{item2} for all variables $X_i$ and $X_j$, with $i<j$, if $P$ has a constraint $(X_j-X_i)\in C_{ij}$, with $C_{ij}$ of the
               form $(-\infty ,b]$, $[a,b]$ or $(a,b]$, then the weight of edge $(X_i,X_j)$ is $b$: $w(X_i,X_j)=b$
      \item\label{item3} for all variables $X_i$ and $X_j$, with $i<j$, if $P$ has a constraint $(X_j-X_i)\in C_{ij}$, with $C_{ij}$ of the
               form $(-\infty ,b)$, $[a,b)$ or $(a,b)$, then the weight of edge $(X_i,X_j)$ is $b^-$: $w(X_i,X_j)=b^-$
      \item\label{item4} for all variables $X_i$ and $X_j$, with $i<j$, if $P$ has a constraint $(X_j-X_i)\in C_{ij}$, with $C_{ij}$ of the
               form $[a,+\infty )$, $[a,b]$ or $[a,b)$, then the weight of edge $(X_j,X_i)$ is $-a$: $w(X_j,X_i)=-a$
      \item\label{item5} for all variables $X_i$ and $X_j$, with $i<j$, if $P$ has a constraint $(X_j-X_i)\in C_{ij}$, with $C_{ij}$ of the
               form $(a,+\infty )$, $(a,b]$ or $(a,b)$, then the weight of edge $(X_j,X_i)$ is $(-a)^-$: $w(X_j,X_i)=(-a)^-$
      \item all other edges $(X_i,X_j)$, with $i\not =j$, are such that $w(X_i,X_j)=+\infty$
    \end{enumerate}
\end{enumerate}
We refer to the rooted distance graph $(G,X_0)$ as the rooted distance graph of the $\stp$ $P$. \cqfd
\end{definition}
\begin{remark}
\begin{enumerate}
  \item If an $\stp$ $P$ contains a constraint $(X_j-X_i)\in [a,b)$, then, from Items \ref{item3} and \ref{item4} of Definition \ref{rdg}, the two edges
	$(X_i,X_j)$ and $(X_j,X_i)$ of the (rooted) distance graph of $P$ are labeled, respectively, with $b^-$ and $-a$. This is so because the constraint is equivalent to the
	following conjunction of linear inequalities: $X_j-X_i<b\wedge X_i-X_j\leq -a$. To have uniform linear inequalities, using exclusively
	$\leq$ (lower than or equal to), we write the inequality $X_j-X_i<b$ as $X_j-X_i\leq b^-$, the minus sign in $b^-$ meaning that the
	upper bound b is not reached. Note that if a distance graph contains an edge $(X,Y)$ labeled with $a$, with $a\in\BBR\cup\BBR ^-$ and $\BBR^-=\{a^-\mbox{ such that }a\in\BBR\}$, this is interpreted as
	$Y-X\leq a$.
  \item In order to be able to apply shortest paths algorithms to a (rooted) distance graph as defined in this work, the addition ($+$) and comparison ($<$) of real numbers
	should be generalized to $\BBR\cup\BBR^-$. This is done as follows, where
	$a,b\in\BBR$:
	\begin{enumerate}
  	  \item $a+b$ and $a<b$ are interpreted in the usual way;
  	  \item $a+b^-=a^-+b=a^-+b^-=(a+b)^-$;
  	  \item $a^-<a$; and
  	  \item if $a<b$ then $a^-<b$,  $a<b^-$ and $a^-<b^-$.
	\end{enumerate}
	Of particular importance for the detection of negative circuits,
	$0^-<0$, meaning that $0^-$ is a (strictly) negative weight. \cqfd
\end{enumerate}
\end{remark}
\begin{theorem}[\cite{Pratt77a,Shostak81a}]\label{shostak}Let $P$ be an $\stp$. $P$ is consistent if and only if its distance graph has no negative circuit. \cqfd
\end{theorem}
The use in \cite{DechterMP91a} of labeled directed graphs to represent a conjunction of linear inequalities is restricted
to large inequalities of the form $X_j-X_i\leq a$, where $a\in\BBR$. An idea similar to the one used in this work, which combines large
and strict inequalities, can be found in \cite{KautzL91a}.
\begin{definition}[d-graph of an $\stp$]The d-graph of an $\stp$ $P=(X,C)$ is the distance graph $G=(X,X\times X,w)$
verifying: $w(X_i,X_j)$ is the weight of the shortest path from $X_i$ to $X_j$ in the distance graph of $P$.
\end{definition}
The d-graph of an $\stp$ can be built from its distance gragh using Floyd-Warshall's all-to-all shortest paths algorithm \cite{AhoHU76a,PapadimitriouS82a} (see also \cite{DechterMP91a}, page 72, Figure 4).

\begin{figure}
\begin{center}
$
\begin{array}{|l|l||l|l|l|}  \cline{3-5}
\multicolumn{2}{l|}{\multirow{2}{*}{}}              &\multicolumn{3}{|c|}{l_2}\\  \cline{3-5}
\multicolumn{2}{l|}{}            &+\infty                &b                  &b^-\\  \hline\cline{3-5}
\multirow{3}{*}{$l_1$}&+\infty     &(-\infty ,+\infty )&[-b,+\infty )&(-b,+\infty )\\  \cline{2-5}
                               &a            &(-\infty ,a]           &[-b,a]          &(-b,a]\\  \cline{2-5}
                               &a^-         &(-\infty ,a)           &[-b,a)         &(-b,a)\\  \hline
\end{array}
$
\end{center}
\caption{The table $M$ used in the construction of the $\stp$ of a rooted distance graph, with $a,b\in\BBR$.}\label{RootedDistanceGraphToSTP}
\end{figure}
\begin{definition}[$\stp$ of a rooted distance graph]Let $(G,V_0)$ be a rooted distance graph, with $G=(V,V\times V,w)$ and
$V=\{V_0,V_1,\ldots ,V_n\}$. The $\stp$ of $(G,V_0)$ is the $\stp$ $P=(X,C)$, with $X=\{X_0,X_1,\ldots ,X_n\}$, constructed as follows:
\begin{enumerate}
  \item initialize $C$ to the empty set: $C=\emptyset$;
  \item for all vertices $V_i$ and $V_j$ of $G$, with $i<j$, such that at least one of the two edges $(V_i,V_j)$ and $(V_j,V_i)$ is not labeled
with $+\infty$:
    \begin{enumerate}
      \item let $l_1=w(V_i,V_j)$ and $l_2=w(V_j,V_i)$;
      \item add to $C$ the constraint $(X_j-X_i)\in M[l_1,l_2]$, where $M[l_1,l_2]$ is as given by the table $M$ of Figure \ref{RootedDistanceGraphToSTP}.
    \end{enumerate}
\end{enumerate}
\end{definition}
For instance, if the edges $(V_1,V_5)$ and $(V_5,V_1)$ are labeled, respectively, with 6 and $(-2)^-$, the constraint $(X_5-X_1)\in (2,6]$ is added to $C$.

\begin{definition}[connected rooted distance graph]A rooted distance graph is connected if its $\stp$ is connected; it is disconnected otherwise.
\end{definition}

In \cite{DechterMP91a}, it has been shown that applying path-consistency to an $\stp$ $P$ is equivalent to applying Floyd-Warshall's all-to-all shortest paths
algorithm \cite{AhoHU76a,PapadimitriouS82a} to the distance graph of $P$. In other words, if $P'$ is the $\stp$ resulting from applying path consistency to $P$ ($P'=PC(P)$),
then $P'$ is exactly the $\stp$ of the d-graph of $P$. Furthermore, $P'$ is minimal and strongly $n+1$-consistent, $n+1$ being the number of variables.
\begin{theorem}[\cite{DechterMP91a}]\label{pcminimality}Let $P$ be an $n+1$-variable $\stp$. If path-consistency applied to $P$ does
not detect an inconsistency, then the resulting $\stp$ $P'$ is minimal and strongly $n+1$-consistent. Furthermore, $P'$ is the $\stp$ of
the d-graph of $P$. \cqfd
\end{theorem}

For the detection of an $\stp$'s inconsistency, which is equivalent to the detection of a negative circuit in its distance graph \cite{Pratt77a,Shostak81a}, we will need
a lower bound of the weight of a finitely-weighted elementary path. Furthermore, for the purpose of evaluating the computational behavior of  the
binarized-domains Arc-Consistency algorithm to be defined, we will additionaly need a upper bound of the same weight. The length, in terms of number of edges, of
an elementary path is bounded by n+1 (the upper bound n+1 is reached when the elementary path is a maximal elementary circuit, traversing each of the n+1 vertices,
before returning to its very first vertex). These are defined as follows.
\begin{definition}\label{pathlbpathubdef}[\mbox{path-lb}, \mbox{path-ub} of an $\stp$]Let $P=(X,C)$ be an $n+1$-variable $\stp$, $G=(X,X\times X,w)$
its rooted distance graph, and $Edges_<(P)$ and $Edges_{\geq}(P)$ the following subsets of $X\times X$:
\begin{enumerate}
  \item[$\bullet$] $Edges_<(P)=$\\
                                                   $\{(X_i,X_j):\mbox{ }i\not =j\mbox{ and }w(X_i,X_j)<0\mbox{ and }w(X_i,X_j)<w(X_j,X_i)\}\cup$\\
                                                   $\{(X_i,X_j):\mbox{ }i<j\mbox{ and }w(X_i,X_j)<0\mbox{ and }w(X_i,X_j)=w(X_j,X_i)\}$,
  \item[$\bullet$] $Edges_{\geq}(P)=$\\
                                                           $\{(X_i,X_j):\mbox{ }i\not =j\mbox{ and }0\leq w(X_i,X_j)<+\infty\mbox{ and }(w(X_j,X_i)=+\infty\mbox{ or }w(X_j,X_i)<w(X_i,X_j))\}\cup$\\
                                                           $\{(X_i,X_j):\mbox{ }i<j\mbox{ and }0\leq w(X_i,X_j)<+\infty\mbox{ and }w(X_j,X_i)=w(X_i,X_j)\}$.
\end{enumerate}
We suppose that the elements of $Edges_<(P)$ (respectively, $Edges_{\geq}(P)$) are sorted in ascending (respectively, descending) order of their weights, and refer to the edge number $i$, $i\geq 1$, as
$Edges_<[P,i]$ (respectively, $Edges_{\geq}[P,i]$). $\mbox{path-lb}(P)$ and $\mbox{path-ub}(P)$ give, respectively, a lower bound and a upper bound of the weight of a finitely-weighted elementary path of
$P$, and are defined as follows:

\begin{equation}
\mbox{path-lb}(P)=
\left\{
    \begin{aligned}
        0\hspace*{4cm}\mbox{ if }Edges_<(P)=\emptyset ,\hspace*{1.05cm}\nonumber\\
        w(Edges_<[P,1])+\Sigma_{i=1}^kw(Edges_<[P,i])\hspace*{1.8cm}\nonumber\\
        \mbox{ if }|Edges_<(P)|=k\leq n,\nonumber\\
       w(Edges_<[P,1])+\Sigma_{i=1}^nw(Edges_<[P,i])\mbox{ otherwise}\nonumber
    \end{aligned}
\right.
\end{equation}

\begin{equation}
\mbox{path-ub}(P)=
\left\{
    \begin{aligned}
      0\hspace*{4cm}\mbox{ if }Edges_{\geq}(P)=\emptyset ,\hspace*{1.05cm}\nonumber\\
      w(Edges_{\geq}[P,1])+\Sigma_{i=1}^kw(Edges_{\geq}[P,i])\hspace*{1.8cm}\nonumber\\
      \mbox{ if }|Edges_{\geq}(P)|=k\leq n,\nonumber\\
      w(Edges_{\geq}[P,1])+\Sigma_{i=1}^nw(Edges_{\geq}[P,i])\mbox{ otherwise}\nonumber
    \end{aligned}
\right.
\end{equation}

\end{definition}
\begin{definition}[\mbox{path-lb}, \mbox{path-ub} of a $\tcsp$]Let $P=(X,C)$ be an $n+1$-variable $\tcsp$: $\mbox{path-lb}(P)=\mbox{path-lb}(cc(P))$ and $\mbox{path-ub}(P)=\mbox{path-ub}(cc(P))$.
\end{definition}
\begin{remark}\label{circuitfreepath}Let $P=(X,C)$ be an $n+1$-variable $\stp$ and $G=(X,X\times X,w)$ its rooted distance graph.
\begin{enumerate}
  \item If the weight $w(u)$ is finite, where $u$ is an elementary path of $G$ in general, or an elementary circuit in particular, then $\mbox{path-lb}(P)\leq w(u)\leq\mbox{path-ub}(P)$.
  \item The lower and upper bounds $\mbox{path-lb}(P)$ and $\mbox{path-ub}(P)$ of Definition \ref{pathlbpathubdef} are reached when the elementary path is a maximal elementary
           circuit, whose length is then $(n+1)$. They can be defined in a much simpler, easier to understand, though not optimized way, which will not compromize their use in establishing
           the termination and the worst-case computational complexity of the binarized-domains arc-consistency algorithm to be defined:
\begin{equation}
\mbox{path-lb}(P)=
\left\{
    \begin{aligned}
        0\hspace*{2.8cm}\mbox{ if }Edges_<(P)=\emptyset ,\nonumber\\
        (n+1)*w(Edges_<[P,1])\mbox{ otherwise}\nonumber
    \end{aligned}
\right.
\end{equation}

\begin{equation}
\mbox{path-ub}(P)=
\left\{
    \begin{aligned}
      0\hspace*{2.8cm}\mbox{ if }Edges_{\geq}(P)=\emptyset ,\nonumber\\
      (n+1)*w(Edges_{\geq}[P,1])\mbox{ otherwise}\nonumber
    \end{aligned}
\right.
\end{equation}
\end{enumerate}
\end{remark}
\section{A binarized-domains arc-consistency algorithm for $\tcsps$}\label{stwo}
\begin{figure}[t]
\begin{enumerate}
  \item[]{\bf procedure }$\mbox{REVISE}(i,j)$\{
  \item\label{revise2}\hskip 0.4cm  DELETE=false;
  \item\label{revise3}\hskip 0.4cm $temp=P_{0i}\cap P_{0j}\otimes P_{ji}$;
  \item\label{revise4}\hskip 0.4cm if($temp\not = P_{0i}$)\{
  \item\label{revise42}\hskip 0.6cm if[$temp\not =\emptyset$ and $(lowerB(temp)>-\mbox{path-lb}(P)$
  \item[]\hskip 1cm or $upperB(temp)<\mbox{path-lb}(P))] temp=\emptyset$;
  \item\label{revise5}\hskip 0.6cm $P_{0i}=temp$;
  \item\label{revise51}\hskip 0.6cm $P_{i0}=temp^\smile$;
  \item\label{revise52}\hskip 0.6cm DELETE=true
  \item\label{revise53}\hskip 0.6cm \}
  \item\label{revise7}\hskip 0.4cm return DELETE
  \item\label{revise72}\hskip 0.4cm \}
\end{enumerate}
\caption{The procedure $\mbox{REVISE}$.}\label{TheREVISEProcedure}
\end{figure}
\begin{figure}[t]
\begin{enumerate}
  \item[]\label{wline3} {\bf procedure }$\mbox{bdAC-3}(P)$\{
  \item\label{wline4}\hskip 0.2cm $Q=\{(i,j):\mbox{ }P\mbox{ }has\mbox{ }a\mbox{ }constraint\mbox{ }on\mbox{ }X_i\mbox{ }and\mbox{ }X_j,$
  \item[]\hskip 1.2cm $\mbox{ }i*j\not =0,\mbox{ }i\not =j\}$;
  \item\hskip 0.2cm $Empty\_domain=false$;
  \item\label{wline5}\hskip 0.2cm {\bf while }$(Q\not =\emptyset$\mbox{ and (not }Empty\_domain))\{
  \item\label{wline6}\hskip 0.8cm select and delete an arc $(k,m)$ from $Q$;
  \item\label{wline7}\hskip 0.8cm {\bf if }$\mbox{REVISE}(k,m)$
  \item\label{wline8}\hskip 1.2cm if $(P_{0k}=\emptyset)$ $Empty\_domain=true$
  \item\label{wline9}\hskip 1.2cm else $Q=Q\cup\{(i,k):\mbox{ $P$ has a constraint}$
   \item[]\hskip 2.0cm \mbox{on $X_i$ and $X_k$, $i\not =0$, $i\not =k$, $i\not =m$}\}
   \item\hskip 0.8cm \}\%endwhile
  \item\label{wline10}\hskip 0.4cm return(not $Empty\_domain$)
  \item\label{wline11}\hskip 0.4cm \}\%end bdAC-3
\end{enumerate}
\caption{The binarized-domains Arc-Consistency algorithm \mbox{bdAC-3}.}\label{bdAC3}
\end{figure}
The transformation of the unary constraints of a $\tcsp$ into binary constraints, after the addition of an "origin of the world" variable $X_0$, makes the
constraints of the $\tcsp$ become exclusively binary. In this work, we look at the constraints between the "origin of the world" variable $X_0$ and the
other variables as representing the (binarized) domains of these other variables. With this in mind, we define a notion of arc-consistency for $\tcsps$, which we
will refer to as binarized-domains arc-consistency, or bdAC for short.
\begin{definition}[binarized-domains arc-consistency]Let $P$ be a $\tcsp$. $P$ is said to verify the binarized-domains arc-consistency, or to be bdArc-Consistent
for short, if and only if for all $i,j\in\{1,\ldots ,n\}$, $i\not = j$, the following holds: $P_{0i}\subseteq P_{0j}\otimes P_{ji}$.
\end{definition}
If $P$ is bdArc-Consistent then, for all $i,j\in\{1,\ldots ,n\}$, $i\not = j$, the following holds: for all $(a_0,a_i)\in\BBR ^2$ such that
$(a_i-a_0)\in P_{0i}$, there exists $a_j\in\BBR$ such that $(a_j-a_0)\in P_{0j}$ and $(a_i-a_j)\in P_{ji}$; in other words, given an
instantiation $X_0=a_0$ of the "origin of the world" variable $X_0$, any instantiation $X_i=a_i$ of variable $X_i$ such that
$(a_i-a_0)$ belongs to the binarized domain $P_{0i}$ of $X_i$, has a supporting value $X_j=a_j$ such that, at the same time,
$(a_j-a_0)$ belongs to the binarized domain $P_{0j}$ of $X_j$ and the instantiation $(X_i,X_j)=(a_i,a_j)$ satisfies the constraints,
if any, on variables $X_i$ and $X_j$.
\begin{definition}\label{reachability}
Let $P$ be a $\tcsp$, and $X_i$ and $X_j$ two variables of $P$. $X_j$ is reachable from $X_i$ if and only if there exists a finite-weight elementary path from
$X_i$ to $X_j$ in the distance graph of $cc(P)$, the convex closure of $P$. $X_i$ and $X_j$ are disconnected from each other if $X_i$ if not reachable from
$X_i$, and $X_j$ is not reachable from $X_i$; $X_i$ and $X_j$ are connected to each other otherwise. $P$ is connectd if $X_0$ is connected to each of the other variables.
\end{definition}
The algorithm of Figure \ref{bdAC3} takes as input a $\tcsp$ $P$, and does the following:
\begin{enumerate}
             \item either, it detects that $P$ is inconsistent; or
             \item it makes $P$ \mbox{bd-ArcConsistent}. 
\end{enumerate}
We refer to the algorithm as \mbox{bdAC-3}, for it is mainly an adaptation of Mackworth's
\cite{Mackworth77a} well-known arc-consistency algorithm \mbox{AC-3}. \mbox{bdAC-3} initializes a queue $Q$ to all pairs $(i,j)$ such that $i\not =0$, $j\not =0$,
$i\not =j$ and $P$ contains a (binary) constraint on $X_i$ and $X_j$. Then it proceeds by taking in turn the pairs in $Q$ for propagation. When a pair $(k,m)$
is taken (and removed) from $Q$, \mbox{bdAC-3} calls the procedure $\mbox{REVISE}$ (Figure \ref{TheREVISEProcedure}) to evenually update $P_{0k}$, which represents the binarized domain of $X_k$,
if it is not a subset of the composition $P_{0m}\otimes P_{mk}$. The main difference of \mbox{bdAC-3} with Mackworth's \cite{Mackworth77a} \mbox{AC-3} lies in
line \ref{revise42} of procedure $\mbox{REVISE}$, which guarantees termination of \mbox{bdAC-3}, and can be explained as follows, where the notations $lowerB(A)$ and
$upperB(A)$, for a set $A\subseteq\BBR$, refer, respectively, to the lower bound and upper bound of $A$:
\begin{enumerate}
  \item[$\bullet$] If $P$ is an $\stp$, $P_{0k}$
is of the form $[lowerB(P_{0k}),upperB(P_{0k})]$, and represents the constraint $lowerB(P_{0k})\leq X_k-X_0\leq upperB(P_{0k})$ on the binarized domain of $X_k$, which  is equivalent to the conjunction of $X_k-X_0\leq upperB(P_{0k})$ (evolution of the shortest path from $X_0$ to $X_k$) and $X_0-X_k\leq -lowerB(P_{0k})$ (evolution of the shortest path from $X_k$ to $X_0$). This means that
$upperB(P_{0k})$ and $-lowerB(P_{0k})$ stand for the current values of the weights of the shortest paths from $X_0$ to $X_k$ and from $X_k$ to
           $X_0$, respectively, in the rooted distance graph of $P$.
	Those of the two values $upperB(P_{0k})$  and $-lowerB(P_{0k})$ that are finite should therefore satisfy the conditions of Theorem \ref{circuitfreepath}. In other words, the
	algorithm should make sure that the evolution of the two weights makes none of them get below the lower bound $\mbox{path-lb}(P)$. Should the condition get violated, this
	would correspond to the detection of a negative circuit in the rooted distance graph of $P$, and \mbox{bdAC-3} would terminate with a negative answer to the consistency
	problem of $P$.
  \item[$\bullet$] If $P$ is a general $\tcsp$, given that $\mbox{path-lb}(P)=\mbox{path-lb}(cc(P))$ and that $P$ is a refinement of its convex closure $cc(P)$, line \ref{revise42} of procedure $\mbox{REVISE}$ still applies.
\end{enumerate}
If the $\mbox{REVISE}$ procedure successfully updates $P_{0k}$, without making it become empty, the pairs $(i,k)$ such that $i\notin\{0,k,m\}$ and $P$ has a constraint on
$X_i$ and $K_k$, reenter the queue $Q$ if they are no longer there.
The algorithm terminates when a binarized domain becomes empty, or when the queue $Q$ becomes empty.

This leads us to the following theorem.
\begin{theorem}
Le $P$ be a $\tcsp$. The \mbox{bdAC-3} algorithm applied to $P$ terminates and, either detects inconsistency of $P$ or computes a bd-Arc Consistent $\tcsp$ equivalent to $P$. \cqfd
\end{theorem}
The range of a $\tcsp$  will be needed for the determinination of the computational complexity of \mbox{bdAC-3}:
\begin{definition}[range of a $\tcsp$]Let $P=(X,C)$ be an n+1-variable $\tcsp$. The range of $P$, $rg(P)$, is defined as $rg(P)=\mbox{path-ub}(P)-\mbox{path-lb}(P)$.
\end{definition}
\begin{theorem}\label{bdAC3complexity}Let $P=(X,C)$ be a $\tcsp$, with $|X|=n+1$ and $|C|=m$. \mbox{bdAC-3} applied to $P$ can be achieved in $O(n^2R)=O(mR)$ relaxation
steps (calls of the procedure REVISE of Figure \ref{TheREVISEProcedure}) and $O(n^2R^3)=O(mR^3)$ arithmetic operations, where $R=rg(P)$ is the range of $P$ expressed
in the coarsest possible time units.
\end{theorem}
{\bf Proof:} The worst case scenario of \mbox{bdAC-3} occurs when, whenever the procedure REVISE updates $P_{0i}$, the length of the set $P_{0i}$ decreases by one time unit.
Furthermore, if $P_{0i}$ has a finite bound, it will be in the set $[\mbox{path-lb}(P),\mbox{path-ub}(P)]$ (Theorem \ref{circuitfreepath}). Therefore, $P_{0i}$ can be
updated at most $\mbox{path-ub}(P)-\mbox{path-lb}(P)=R=O(R)$ times. The pairs $(i,j)$ that enter initially the queue $Q$ of \mbox{bdAC-3} are such that there is a constraint of
$P$ on $X_i$ and $X_j$. A pair $(k,i)$ can reenter the queue $O(R)$ times, whenever $P_{0i}$ has been updated. Because there are $m=O(n^2)$ constraints, the number of
relaxation steps is $O(mR)=O(n^2R)$. A relaxation step consists of a call $\mbox{REVISE}(i,j)$ consisting mainly of the computation of a path consistency operation of the form
$P_{0i}=P_{0i}\cap P_{0j}\otimes P_{ji}$, which needs $O(R^2)$ arithmetic operations since each of the three sets $P_{0i}$, $P_{0j}$ and $P_{ji}$ is a union of at most
$R$ convex subsets. The whole algorithm therefore terminates in $O(mR^3)=O(n^2R^3)$ time. \cqfd

For an $\stp$, a relaxation step is in $O(1)$ (see Remark \ref{remark2}). This justifies the following corollary to Theorem \ref{bdAC3complexity}.
\begin{corollary}\label{stpbdAC3complexity}Let $P=(X,C)$ be an $\stp$, with $|X|=n+1$ and $|C|=m$. \mbox{bdAC-3} applied to $P$ can be achieved in $O(n^2R)=O(mR)$ relaxation
steps (calls of the procedure REVISE of Figure \ref{TheREVISEProcedure}) and $O(n^2R)=O(mR)$ arithmetic operations, where $R=rg(P)$ is the range of $P$ expressed
in the coarsest possible time units. \cqfd
\end{corollary}
\begin{theorem}\label{minimalityresult}Let $P$ be a connected $\stp$. If $P$ is bdArc-Consistent, its (binarized) domains
are minimal: for all $i\in\{1,\ldots ,n\}$, for all $a\in P_{0i}$, there exists a solution $(X_0,X_1,\ldots ,X_i,\ldots ,X_n)=(t_0,t_1,\ldots ,t_i,\ldots ,t_n)$ such that
$t_i-t_0=a$.
\end{theorem}
{\bf Proof:} Let $P$ be a connected bdArc-Consistent $\stp$. Let $(G,X_0)$ be the rooted distance
graph of $P$, with $G=(X,X\times X,w)$. According to Theorem \ref{pcminimality}, showing
that the binarized domains $P_{0i}$, with $i\in\{1,\ldots ,n\}$, are minimal, is equivalent to showing that the labels (weights) $w(X_0,X_i)$ and $w(X_i,X_0)$
are the weights of the shortest paths from $X_0$ to $X_i$ and from $X_i$ to $X_0$, respectively. Suppose that this is not the case; in other words, that for
some $i\in\{1,\ldots ,n\}$, either $w(X_0,X_i)$ is not the weight of the shortest path from $X_0$ to $X_i$, or $w(X_i,X_0)$ is not the weight of the shortest
path from $X_i$ to $X_0$:
\begin{enumerate}
  \item Case 1: $w(X_0,X_i)$ is not the weight of the shortest path from $X_0$ to $X_i$. This would mean the existence of $j$ in $\{1,\ldots ,n\}$, $j\not =i$, and
of a path $<X_{i_0}=X_0,X_{i_1}=X_j,\ldots ,X_{i_k}=X_i>$  from $X_0$ to $X_i$ through $X_j$, whose weight is strictly smaller than $w(X_0,X_i)$:
$w(X_{i_0},X_{i_1})+w(X_{i_1},X_{i_2})+\cdots +w(X_{i_{k-1}},X_{i_k})<w(X_{i_0},X_{i_k})$. But, because the $\stp$ is bdArc-Consistent, we have the following:
\begin{equation}
\left\{
    \begin{aligned}
      w(X_{i_0},X_{i_2})&\leq   &w(X_{i_0},X_{i_1})+w(X_{i_1},X_{i_2})\nonumber\\
      w(X_{i_0},X_{i_3})&\leq   &w(X_{i_0},X_{i_2})+w(X_{i_2},X_{i_3})\nonumber\\
      \cdots                     &\cdots&\cdots\nonumber\\
      w(X_{i_0},X_{i_k})&\leq   &w(X_{i_0},X_{i_{k-1}})+w(X_{i_{k-1}},X_{i_k})\nonumber
    \end{aligned}
\right.
\end{equation}
    Summing the left-hand sides and the right-hand sides of the linear inequalities, we get:
    $w(X_{i_0},X_{i_k})+\displaystyle\Sigma _{l=2}^{k-1}w(X_{i_0},X_{i_l})\leq w(X_{i_{i_0}},X_{i_1})+w(X_{i_1},X_{i_2})+\cdots +w(X_{i_{k-1}},X_{i_k})+\displaystyle\Sigma _{l=2}^{k-1}w(X_{i_0},X_{i_l})$.
    This, in turn, gives $w(X_{i_0},X_{i_k})\leq w(X_{i_{i_0}},X_{i_1})+w(X_{i_1},X_{i_2})+\cdots +w(X_{i_{k-1}},X_{i_k})$, which clearly contredicts our supposition.
  \item Case 2: $w(X_i,X_0)$ is not the weight of the shortest path from $X_i$ to $X_0$. This would mean the existence of $j$ in $\{1,\ldots ,n\}$, $j\not =i$, and
of a path $<X_{i_0}=X_i,X_{i_1}=X_j,\ldots ,X_{i_k}=X_0>$  from $X_i$ to $X_0$ through $X_j$, whose weight is strictly smaller than $w(X_i,X_0)$:
$w(X_{i_0},X_{i_1})+w(X_{i_1},X_{i_2})+\cdots +w(X_{i_{k-1}},X_{i_k})<w(X_{i_0},X_{i_k})$. But, because the $\stp$ is bdArc-Consistent, we have the following:
\begin{equation}
\left\{
    \begin{aligned}
      w(X_{i_0},X_{i_k})&\leq   &w(X_{i_0},X_{i_1})+w(X_{i_1},X_{i_k})\nonumber\\
      w(X_{i_1},X_{i_k})&\leq   &w(X_{i_1},X_{i_2})+w(X_{i_2},X_{i_k})\nonumber\\
      \cdots                     &\cdots&\cdots\nonumber\\
      w(X_{i_{k-2}},X_{i_k})&\leq   &w(X_{i_{k-2}},X_{i_{k-1}})+w(X_{i_{k-1}},X_{i_k})\nonumber
    \end{aligned}
\right.
\end{equation}
    Summing the left-hand sides and the right-hand sides of the linear inequalities, we get:
    $w(X_{i_0},X_{i_k})+\displaystyle\Sigma _{l=1}^{k-2}w(X_{i_l},X_{i_k})\leq w(X_{i_0},X_{i_1})+w(X_{i_1},X_{i_2})+\cdots +w(X_{i_{k-1}},X_{i_k})+\displaystyle\Sigma _{l=1}^{k-2}w(X_{i_l},X_{i_k})$.
    This, in turn, gives $w(X_{i_0},X_{i_k})\leq w(X_{i_0},X_{i_1})+w(X_{i_1},X_{i_2})+\cdots +w(X_{i_{k-1}},X_{i_k})$, which clearly contredicts our supposition. \cqfd
\end{enumerate}

As an immediate consequence of Theorem \ref{minimalityresult}:
\begin{corollary}\label{onetoallalltoone}
Let $(G,V)$ be a connected rooted distance graph. The bdArc-Consistency algorithm \mbox{bdAC-3} can be used to detect the presence of a negative circuit in $G$, if any;
or, otherwise, compute the weights of the shortest paths from $V$ to all vertices of $G$, and the weights of the shortest paths from all vertices of $G$ to $V$.
\end{corollary}
{\bf Proof.}  Proceed as follows:
\begin{enumerate}
  \item Linearly transform the input connected rooted distance graph $(G,V)$ into its connected $\stp$ $P$;
  \item apply the bdArc-Consistency algorithm \mbox{bdAC-3} to $P$;
  \item if \mbox{bdAC-3} detects inconsistency of $P$, $G$ contains a negative circuit; otherwose, linearly transform the $\stp$ resulting from \mbox{bdAC-3}
            into its rooted distance graph, to have the lengths of the shortest paths from $V$ to all vertices, and from all vertices to $V$. \cqfd
\end{enumerate}
\begin{remark}\label{remark3}
\begin{enumerate}
  \item We refer to the procedure $\mbox{REVISE}$ of Figure \ref{TheREVISEProcedure} without its line \ref{revise42} as $\mbox{REVISE}^-$.
  \item We refer to the algorithm $\mbox{bdAC-3}$ (Figure \ref{bdAC3}) in which procedure $\mbox{REVISE}$ is replaced with $\mbox{REVISE}^-$ as $\mbox{bdAC-3}^-$.
  \item $\mbox{bdAC-3}^-$ is a "literal" adaptation of Mackworth's \cite{Mackworth77a} \mbox{AC-3} algorithm to $\tcsps$.
\end{enumerate}
\end{remark}
Appendices \ref{appendix6} and \ref{appendix7} present two detailled examples. The example of Appendix \ref{appendix6} applies the \mbox{bdAC-3} algorithm to an $\stp$,
transforming the latter into a bdArc-Consistent $\stp$ with no variable disconnected from the "origin of the world" variable, without making a binarized domain become empty. The example
of Appendix \ref{appendix7} applies \mbox{bdAC-3} to an $\stp$ whose rooted distance graph presents a negative circuit not reachable from $X_0$, the "origin of the world" variable,
but from which $X_0$ is reachable: \mbox{bdAC-3} detects the negative circuit, and the inconsistency of the $\stp$. In particular, the example of Appendix \ref{appendix7} shows that
\mbox{bdAC-3} without line \ref{revise42} of procedure REVISE, which could be seen as a "literal" adaptation to $\tcsps$ of Mackworth's \cite{Mackworth77a} algorithm \mbox{AC-3},
has no guarantee to always terminate:
\begin{theorem}{}
$\mbox{bdAC-3}^-$ is not guaranteed to terminate. \cqfd
\end{theorem}
\begin{figure}
\begin{enumerate}
  \item[] {\bf Input: }a rooted distance graph $(G,V_0)$, with $G=(V,V\times V,w)$ and $V=\{V_0,V_1,\ldots ,V_n\}$.
  \item[] {\bf Output: }returns false, if $G$ contains a negative circuit reachable from $V_0$, or a negative circuit from which $V_0$ is reachable;
	computes the weights of the shortest paths from $V_0$ to all edges, and from all
	edges to $V_0$, and returns true, otherwise.
  \item[] {\bf procedure }$\mbox{BellmanFord}(G,V_0)$
  \item\label{bf1} {\bf for all} $i\in\{0,\ldots n\}$\{
  \item\label{bf2}\hskip 0.4cm $d\_to(V_i)=+\infty$; $pred(V_i)=NIL$;
  \item\label{bf3}\hskip 0.4cm $d\_from(V_i)=+\infty$; $succ(V_i)=NIL$
  \item\label{bf4}\hskip 0.4cm \}
  \item\label{bf5} $d\_to(V_0)=0$; $d\_from(V_0)=0$;
  \item\label{bf6} {\bf for }($i=1$ to $n$)
  \item\label{bf7}\hskip 0.4cm {\bf for all} edges $(V_i,V_j)$ with finite weight\{
  \item\label{bf8}\hskip 0.8cm {\bf if }$d\_to(V_j)>d\_to(V_i)+w(V_i,V_j)$\{
  \item\label{bf9}\hskip 1.2cm $d\_to(V_j)=d\_to(V_i)+w(V_i,V_j)$; $pred(V_j)=V_i$\}
  \item\label{bf10}\hskip 0.8cm {\bf if }$d\_from(V_i)>w(V_i,V_j)+d\_from(V_j)$\{
  \item\label{bf11}\hskip 1.2cm $d\_from(V_i)=w(V_i,V_j)+d\_from(V_j)$; $succ(V_i)=V_j$\}
  \item\label{bf112}\hskip 0.8cm \}
  \item\label{bf12} {\bf for all} edges $(V_i,V_j)$ with finite weight
  \item\label{bf13}\hskip 0.8cm {\bf if }[($d\_to(V_j)>d\_to(V_i)+w(V_i,V_j)$) {\bf or }
  \item\label{bf14}\hskip 1.4cm($d\_from(V_i)>w(V_i,V_j)+d\_from(V_j)$)] {\bf return false}
  \item\label{bf15} {\bf return }true
\end{enumerate}
\caption{A generalization of Bellman-Ford-Moore's algorithm, from one-to-all to one-to-all all-to-one.}\label{bellmanford}
\end{figure}
\begin{remark}\label{remarkbf}From Corollary \ref{onetoallalltoone}, one can get a generalization of Bellman-Ford-Moore's one-to-all algorithm, so that, given a source node $V_0$, it returns false if there
 is a negative circuit reachable from $V_0$, or from which $V_0$ is reachable; and computes otherwise the weights of the shortest paths from $V_0$ to all nodes, and from all nodes to $V_0$,
before returning true. See Figure \ref{bellmanford} for more details, where:
\begin{enumerate}
  \item $d\_to(V_i)$ and $d\_from(V_i)$, i from $0$ to $n$, both initialized to $+\infty$ (except for $i=0$, for which the initialization is to $0$), are the weights of the shortest paths from $V_0$ to
	$V_i$ and from $V_i$ to $V_0$, respectively;
  \item $pred(V_i)$ and $succ(V_i)$, $i$ from $0$ to $n$, both initialized to $NIL$, are used to record the immediate predecessor and the immediate successor, respectively, in the shortest paths
	from $V_0$ to $V_i$ and from $V_i$ to $V_0$;
  \item the algorithm  proceeds by passes, $n$ passes exactly (line \ref{bf6}), each of which considers once and only once each of the edges $(V_i,V_j)$ whose weights are finite, and does the following:
	\begin{enumerate}
	  \item it checks whether to update the weight of the shortest path from $V_0$ to $V_j$ and the immediate predecessor of node $V_j$ in that shortest path (lines \ref{bf8} and \ref{bf9});
	  \item it checks whether to update the weight of the shortest path from $V_i$ to $V_0$ and the immediate successor of node $V_i$ in that shortest path (lines \ref{bf10} and \ref{bf11});
	\end{enumerate}
  \item after the $n$ passes, the algorithm checks whether there is a negative circuit reachable form $V_0$, or from which $V_0$ is reachable (lines \ref{bf12}, \ref{bf13} and \ref{bf14}), in which case it returns false;
  \item if there is no negative circuit reachable from $V_0$, or from which $V_0$ is reachable, the algorithm returns true (line \ref{bf15});
  \item the algorithm cannot detect negative circuits not reachable from $V_0$, and from which $V_0$ is not reachable;
  \item finally, the worst case computational complexity of the algorithm is clearly the same as the original Bellman-Ford-Moore's one-to-all algorithm, namely $O(mn)$, $m$ being the number of edges with a finite weight, and $n+1$ the number of nodes.
\end{enumerate}
\end{remark}
\section{Searching for a solution of a connected bdArc-Consistent $\stp$: a polynomial backtrack-free procedure}\label{bfpone}
\begin{figure}[t]
\begin{enumerate}
  \item[] {\bf Input: }The matrix representation of an n+1-variable bdArc-Consistent $\stp$ $P$ with no variable disconnected from $X_0$.
  \item[] {\bf Output: }A solution of $P$.
  \item[] {\bf procedure }$backtrack\_free (P)$\{
  \item\label{yline1} {\bf if}($P$ has non-singleton binarized domains)\{
  \item\label{yline2}\hskip 0.4cm select a non-singleton edge $(X_0,X_i)$;
  \item\label{yline3}\hskip 0.4cm Instantiate $(X_0,X_i)$ with an element $a_i$ of $P_{0i}$:
  \item[]\hskip 0.8cm $P_{0i}=\{a_i\};P_{i0}=\{-a_i\}$;
  \item\label{yline4}\hskip 0.4cm $P=\mbox{bdAC-3}(P)$;
  \item\label{yline5}\hskip 0.4cm $backtrack\_free (P)$\}\%endif
  \item\label{yline6} {\bf else write } $solution=(0,a_0,\cdots ,a_n)$
  \item[]\hskip 0.8cm \% $a_i$, with $i$ from $1$ to $n$,  is such that $\{a_i\}$ is the singleton
  \item[]\hskip 0.8cm \% binarized domain $P_{0i}$ of variable $X_i$.
  \item \}
\end{enumerate}
\caption{A polynomial backtrack-free procedure computing a solution of a connected bdArc-Consistent $\stp$.}\label{BacktrackFreeSearch}
\end{figure}
Correctness of the backtrack-free procedure of Figure \ref{BacktrackFreeSearch} is a direct consequence of Theorem \ref{minimalityresult}. If the input
connected bdArc-Consistent $\stp$ $P$ is not a singleton $\stp$ (line \ref{yline1}), we select a non-singleton edge $(X_0,X_i)$ and instantiate it with an element $a_i$
of $P_{0i}$ (lines \ref{yline2} and \ref{yline3}): because $P$ is connected and bdArc-Consistent, the binarized domain $P_{0i}$ is minimal (Theorem \ref{minimalityresult}),
which guarantees that the element $a_i$ chosen to instantiate edge $(X_0,X_i)$ participates in a global solution.  bdArc-Consistency is then applied to the
refinement resulting from the instantiation (line \ref{yline4}). The process needs to be reiterated at most $n$ times before the binarized domains $P_{0i}$,
$i\in\{1,\ldots ,n\}$, become singleton, from which, then, a solution is extracted as $(0,a_0,\cdots ,a_n)$, with $a_i$ such that $P_{0i}=\{a_i\}$ (line \ref{yline6}).

\begin{figure}[t]
\begin{enumerate}
  \item[] {\bf Input: }The matrix representation of a bdArc-Consistent $\stp$ $P$ having variables disconnected from $X_0$.
  \item[] {\bf Output: }Extracts a solution of a connected bdArc-Consistent $\stp$ refinement of $P$, hence of $P$, and returns true, if $P$ consistent; returns false otherwise.
  \item[] {\bf procedure }$connectX0\_in\_bdACstp (P)$\{
\end{enumerate}
\begin{enumerate}
  \item\label{xline13}\hskip 0.2cm select a variable $X_i$ disconnected from $X_0$, and a real number $a$;
  \item\label{xline14}\hskip 0.2cm $P_{0i}=[a,+\infty )$; $P_{i0}=(-\infty ,-a]$;
  \item\label{xline15}\hskip 0.2cm {\bf if}($\mbox{bdAC-3}(P)$)
  \item\label{xline16}\hskip 0.6cm  {\bf if }($P$ has no variable disconnected from $X_0$)\{
  \item[]\hskip 2cm  $backtrack\_free (P)$; {\bf return }true\}
  \item\label{xline17}\hskip 0.6cm  {\bf else }$connectX0\_in\_bdACstp (P)$
  \item\label{xline18}\hskip 0.2cm {\bf else }return false
  \item \}
\end{enumerate}
\caption{The polynomial backtrack-free procedure $connectX0\_in\_bdACstp(P)$.}\label{connectX0}
\end{figure}
\section{Consistency of a bdArc-Consistent $\stp$: a polynomial backtrack-free procedure}\label{bfptwo}
Given a bdArc-Consistent $\stp$ $P$, the backtrack-free procedure\\
$connectX0\_in\_bdACstp(P)$ of Figure \ref{connectX0} checks consistency of $P$: it extracts a solution of a connected
bdArc-Consistent $\stp$ refinement of $P$, hence of $P$, and returns true, if $P$ consistent; it returns false otherwise. Correctness of procedure
$connectX0\_in\_bdACstp(P)$ is guaranteed by Theorem \ref{disconnectedvariables}.
\begin{theorem}\label{disconnectedvariables}
Let $P=(X,C)$ be a bdArc-Consistent $\stp$ with a variable $X_i$, $i\geq 1$, disconnected from $X_0$, and $P'=(X,C\cup\{(X_i-X_0)\in [a,+\infty )\})$ the
$\stp$ obtained from $P$ by adding the constraint $(X_i-X_0)\in [a,+\infty )$, with $a\in\BBR$. $P$ is consistent if and only if $P'$ is consistent.
\end{theorem}
{\bf Proof:} Immediate consequence of Theorem \ref{shostak} and the fact that the rooted distance graphs of $P$ and $P'$ are such that one has a negative circuit if and only if the other has. \cqfd
\section{A $\tcsp$ solver with \mbox{wbdAC-3} as the filtering procedure}\label{sthree}
\begin{figure}
\begin{enumerate}
  \item[] {\bf Input: }The matrix representation of an n+1-variable $\tcsp$ $P$.
  \item[] {\bf Output: }Extracts a solution of a connected bdArc-Consistent $\stp$ refinement of $P$, hence of $P$, and returns true, if $P$ consistent; returns false otherwise.
  \item[] {\bf procedure }$consistent (P)$\{
  \item\label{xline1} {\bf if}(not \mbox{wbdAC-3}$(P)$){\bf return false}
  \item\label{xline2} {\bf else}
  \item\label{xline3}\hskip 0.4cm {\bf if}($P$ has disjunctive edges)\{\%beginif2
  \item\label{xline4}\hskip 0.8cm select a disjunctive edge $(X_i,X_j)$, with $0\leq i<j\leq n$;
  \item\label{xline5}\hskip 0.8cm let $\mbox{mPcs}(P_{ij})=\{P_{ij_1},\ldots ,P_{ij_{n_{ij}}}\}$;
  \item\label{xline6}\hskip 0.8cm {\bf for all }$S\in\mbox{mPcs}(P_{ij}$)\{
  \item\label{xline7}\hskip 1.2cm $P'=P$;
  \item\label{xline72}\hskip 1.2cm $P_{ij}'=S$;
  \item\label{xline73}\hskip 1.2cm $P_{ji}'=S^\smile$;
  \item\label{xline8}\hskip 1.2cm {\bf if}($consistent (P')$){\bf return true}\}\%endfor
  \item\label{xline9}\hskip 0.8cm {\bf return false}\}\%endif2
  \item\label{xline10}\hskip 0.4cm  {\bf else}
  \item\label{xline11}\hskip 0.6cm  {\bf if }($P$ has no variable disconnected from $X_0$)\{
  \item\label{xline112}\hskip 2cm  $backtrack\_free (P)$;
  \item\label{xline113}\hskip 2cm  {\bf return }true\}
  \item\label{xline12}\hskip 0.6cm  {\bf else return }$connectX0\_in\_bdACstp (P)$
  \item \}
\end{enumerate}
\caption{The recursive procedure $consistent(P)$.}\label{TheTCSPSolver}
\end{figure}
We define a refinement of a TCSP $P$ to be any TCSP $P'$  on the same set of variables such
that for all constraint $(X_j-X_i)\in P_{ij}^{'}$ of $P'$, the corresponding constraint
$(X_j-X_i)\in P_{ij}$ of $P$ verifies $P_{ij}^{'}\subseteq P_{ij}$. A subset $S$ 
of $P_{ij}$ is a subweight of edge $(X_i,X_j)$; it is a convex subweight if it is convex; it is a maximal
convex subweight, if it is a convex subweight and, for all convex set $S'$ such that $S\subset S'$,
$S'$  is not a subweight of $(X_i,X_j)$. A
refinement is convex if it is an STP \cite{DechterMP91a}. The solver can now be described as follows (see Figure \ref{TheTCSPSolver}).
As the filtering procedure during the search, it uses a weak version of the
bdArc-Consistency algorithm \mbox{bdAC-3}, which we refer to as \mbox{wbdAC-3}, and consists of replacing composition with
weak composition in the $\mbox{REVISE}$ procedure of Figure \ref{TheREVISEProcedure}, the aim being to avoid the ``fragmentation
problem'' \cite{SchwalbD97a}. If $P$ is the input $\tcsp$, the recursive procedure $consistent(P)$ is called, which works as follows. The
filtering procedure \mbox{wbdAC-3} is applied (the very first application of the filtering, at the root of the search space, consists of the
preprocessing step). If the \mbox{wbdAC-3} filtering detects an inconsisteny (line \ref{xline1}), by reducing a binarized domain to the
empty set, a failure (dead end)  is reached and the procedure:
\begin{enumerate}
  \item returns false to the consistency problem of the input $\tcsp$ if the current node is the root of the search space; or
  \item backtracks by returning false to the parent node, otherwise, as a sign that the exploration has not found a solution yet and
            should be pursued.
\end{enumerate}
Otherwise, if $P$ has no disjunctive edge (line \ref{xline10}) then:
\begin{enumerate}
  \item the result of the \mbox{wbdAC-3} filtering is a bdArc-Consistent $\stp$, which is not necessarily consistent since it might contain a negative circuit disconnected from $X_0$;
  \item in such a case, the procedure does the following:
  \begin{enumerate}
  \item if $P$ has no variable disconnected from $X_0$ then $P$ is consistent (Theorem \ref{minimalityresult}), and the procedure extracts a solution (call of $backtrack\_free$)
	before returning true (lines \ref{xline112} and \ref{xline113});
  \item if $P$ has variables that are disconnected from $X_0$ (line \ref{xline12}), these might hide a negative circuit, of which $X_0$ cannot be a vertex. The backtrack-free search
	procedure $connectX0\_in\_bdACstp(P)$ is called, which extracts a solution of a connected
bdArc-Consistent $\stp$ refinement of $P$ before returning true, if $P$ consistent, and returns false otherwise: $consistent (P)$ returns the boolean result returned by
	$connectX0\_in\_bdACstp(P)$, which indicates whether a solution has already been found.
  \end{enumerate}
\end{enumerate}
If the result of the \mbox{wbdAC-3} filtering is not an STP then there remain disjunctive edges (line \ref{xline3}). A
disjunctive edge is selected  (lines \ref{xline4} and \ref{xline5}) and instantiated with one of its maximal convex subweights (lines \ref{xline7}, \ref{xline72} and \ref{xline73}):
$P'$ is the refinement of $P$ resulting from the instantiation. The recursive call $consistent(P')$ is then made (line \ref{xline8}). If $consistent(P')$ returns true, $consistent(P)$ returns true.
If $consistent(P')$ returns false, the next maximal convex subweight, if any, of the
edge being instantiated is chosen, and the recursive call $consistent(P')$ is made again. If all
subweights have been chosen then $consistent(P)$ returns false (line \ref{xline9}); in other words:
\begin{enumerate}
  \item it returns false to the consistency problem of the input $\tcsp$ if the current node is the root of the search space; or
  \item backtracks by returning false to the parent node, otherwise, as a sign that the exploration has not found a solution yet and
            should be pursued.
\end{enumerate}
Completeness of the solver is guaranteed by completeness of bdArc-Consistency for connected $\stps$ (Theorem \ref{minimalityresult}).

The next section describes how, from the general $\tcsp$ solver, one can extract a $\tcsp$-based job shop scheduler with \mbox{wbdAC-3}
as the filtering procedure during the search. Then it shows that the use in such a solver of \mbox{bdAC-3} as the filtering procedure during
the search, instead of \mbox{wbdAC-3}, is possible, despite the fact that it can lead to a fragmentation of the binarized domains, as illustrated by Example
\ref{example}(\ref{fragprobright}-\ref{fragprobright2}) and the accompanying Figure \ref{fragprob}(right).
\section{\mbox{A $\tcsp$-based job shop scheduler with wbdAC-3} or \mbox{bdAC-3} as the filtering procedure}\label{scheduler}
Many approaches to scheduling are known in the literature, among which
mathematical programming, timed Petri nets \cite{CarlierCG84a} and discrete
CSPs \cite{DincbasSvH90a,SadehSX95a}. We focus on job shop scheduling problems.
\subsection{\mbox{wbdAC-3} as the filtering procedure}\label{schedulerSubs1}
A scheduling $\tcsp$ is a $\tcsp$ $P=(X,C)$, with $X=\{X_0,X_1,\ldots ,X_n\}$, $X_0$ being the
''origin of the world'' variable standing for a global release date, $n$ being the number of (non-preemptive) tasks:
\begin{enumerate}
  \item For $i\in\{1,\ldots ,n\}$:
    \begin{enumerate}
      \item variable $X_i$ stands for the starting date of task $i$;
      \item the duration of task $i$ is $d_i$, with $d_i>0$
	(the tasks are durative, in the sense that their durations are not null; furthermore, they have fixed durations, known in advance); and
      \item the release and due dates of task $i$ are, respectively, $rd_i$ and $dd_i$. If no release (respectively, due) date is given for a task,
	we suppose that it is equal to $0$ (respectively, $+\infty$).
    \end{enumerate}
  \item A
conjunctive (or precedence) constraint between tasks $i$ and $j$ has the form $(X_j-X_i)\in [d_i,+\infty )$.
  \item A disjunctive
constraint between tasks $i$ and $j$ has the form $(X_j-X_i)\in (-\infty ,-d_j]\cup [d_i,+\infty )$.
  \item A release (respectively,
due) date constraint has the form $(X_i-X_0)\in [rd_i,+\infty )$ (respectively, $(X_i-X_0)\in (-\infty,dd_i-d_i]$).
  \item Finally, $X_0$
standing for a global release date, we add the $n$ constraints $(X_i-X_0)\in [0,+\infty )$, $i\in\{1,\ldots ,n\}$.
\end{enumerate}
\begin{figure}
\begin{enumerate}
  \item[]\label{zline1} {\bf Input: }The matrix representation of an n+1-variable scheduling $\tcsp$ $P$.
  \item[]\label{zline2} {\bf Output: }The optimum $z$ of $P$ and a solution $s$ realizing it.
  \item[]\label{zline3} {\bf Method: }initialize $z$: $z=+\infty$; then call $optimum (P)$.
  \item[]\label{zline4} {\bf procedure }$optimum (P)$\{
  \item\label{zline5} {\bf if }(not \mbox{wbdAC-3}$(P)$ {\bf or }$z\leq OLB (P))$ return
  \item\label{zline6} {\bf else}
  \item\label{zline7}\hskip 0.3cm {\bf if}($P$ has disjunctive edges)\{\%beginif2
  \item\label{zline8}\hskip 0.6cm select a disjunctive edge $(X_i,X_j)$, with $1\leq i<j\leq n$;
  \item\label{zline9}\hskip 0.6cm let $A=P_{ij}=A_1\cup A_2$;
  \item\label{zline10}\hskip 0.6cm {\bf for}($k=1\mbox{ to }2$)\{
  \item\label{zline12}\hskip 1.2cm $P'=P$; $P_{ij}'=A_k$; $P_{ji}'=A_k^\smile$;
  \item\label{zline13}\hskip 1.2cm $optimum (P')$
  \item\label{zline132}\hskip 1.2cm \}\%endfor
  \item\hskip 0.6cm \}\%endif2
  \item\label{zline14}\hskip 0.3cm {\bf else }\{\%check whether the optimum should be updated ...
  \item\label{zline142}\hskip 0.6cm $solP=(0,lowerB(P_{01}),\ldots ...,lowerB(P_{0n}))$;
  \item\label{zline143}\hskip 0.6cm $optP=dur(solP)$;
  \item\label{zline144}\hskip 0.6cm {\bf if }$z>optP$\{\%update ...
  \item\label{zline145}\hskip 0.9cm $s=solP$; $z=optP$\}\}
  \item \}
\end{enumerate}
\caption{The $\tcsp$-based job-shop scheduler.}\label{jsscheduler}
\end{figure}
Let $s=(t_0,t_1,\ldots ,t_n)$ be a solution of $P$, assigning time $t_i$ to variable $X_i$.
We define
the latency $lat(s)$ of $s$ as
the amount of time separating stimulus and response, the stimulus
being given at time $t_0$, and the response obtained at the effective beginning of the very first task: $lat(s)=\displaystyle min_{i=1}^{n}t_i-t_0$.
The duration, or makespan, of $s$ is defined as follows:
\begin{enumerate}
  \item as $dur(s)=\displaystyle max_{i=1}^{n}(t_i+d_i)-t_0$, if we consider the latency of $s$ as part of its duration;
  \item as $dur(s)=\displaystyle max_{i=1}^{n}(t_i+d_i)-\displaystyle min_{i=1}^{n}t_i$, otherwise.
\end{enumerate}
The optimum of $P$,
$t_{opt}^P$, is defined as $t_{opt}^P=\displaystyle min_{s\in sol(P)}dur(s)$, with $sol(P)$ being the set of
solutions of $P$.

The scheduler we propose (Figure \ref{jsscheduler}) minimises the makespan and initializes the optimum $z$ to $+\infty$; it is an
adaptation of the $\tcsp$ solver of Figure \ref{TheTCSPSolver}. The lower bound of
the optimum of $P$, $OLB(P)$, is defined as follows, where $a_i$, $i$ from $1$ to $n$, is the lower bound of the binarized
domain $P_{0i}$:
\begin{enumerate}
  \item as $OLB(P)=\displaystyle max_{i=1}^{n}(a_i+d_i)$, if we consider the latency of a solution to $P$ as part of its duration;
  \item as $OLB(P)=\displaystyle max_{i=1}^{n}(a_i+d_i)-\displaystyle max(\{0\}\cup\{rd_i:\mbox{ $i$ from $1$ to $n$ and $rd_i$ finite}\})$, otherwise.
\end{enumerate}
\begin{theorem}\label{minimality1}let $P$ be an n+1-variable scheduling $\tcsp$, $P'$ an $\stp$ refinement of $P$, and $P''$ the $\stp$ refinement resulting from
$\mbox{bdAC-3}$ (or, indeed, from $\mbox{wbdAC-3}$) applied to $P'$: $P''=\mbox{bdAC-3}(P')=\mbox{wbdAC-3}(P')$. The optimum
$t_{opt}^{P'}$ of $P'$ and a solution $s$ realizing it can be extracted from the lower bounds $a_i$, $i$ from 1 to n, of the binarized domains
$P^{''}_{0i}$ of $P''$, as follows: $s=(0,a_1,\ldots ,a_n)$, $t_{opt}^{P'}=dur(s)$.
\end{theorem}
{\bf Proof:} Let $P$, $P'$ and $P''$ be as stated in the theorem. Given the nature of a scheduling $\tcsp$ ($X_0$ is a global release date),
$X_0$ is disconnected from none of the other variables. Because the binarized domains of a connected bdArc-Consistent $\stp$ are minimal (Theorem
\ref{minimalityresult}), a direct consequence of results in \cite{DechterMP91a} (Corollaries 3.2 and 3.4, Pages 69 and 70) is that a solution
$s$ realizing the optimum  $t_{opt}^{P'}$ is given by the lower bounds $a_i$, $i$ from 1 to n, of the binarized domains $P^{''}_{0i}$ of $P''$
as follows: $s=(0,a_1,\ldots ,a_n)$, $t_{opt}^{P'}=dur(s)$. \cqfd
\begin{theorem}\label{minimality2}let $P$ be a scheduling $\tcsp$. $t_{opt}^P\geq OLB(P)$.
\end{theorem}
{\bf Proof:} Let $P$ be a scheduling $\tcsp$, and denote by $a_i$ the lower bound of the binarized domain $P_{0i}$, $i$ from 1 to n: $a_i=lowerB(P_{0i})$.
Clearly, because the optimum of $P$ is the best of the optima of its maximal $\stp$ refinements that are consistent, it is sufficient to show that $t_{opt}^{P'}\geq OLB(P)$, for
all such refinements $P'$. We consider such a refinement $P'$ and denote by $P''$ the $\stp$ refinement of $P$ resulting from \mbox{bdAC-3} (or, indeed,
from \mbox{wbdAC-3}) applied to $P'$: $P''=\mbox{bdAC-3}(P')=\mbox{wbdAC-3}(P')$. From Theorem \ref{minimality1}, we can extract from the lower bounds $b_i$, $i$ from 1 to n, of the binarized domains
$P^{''}_{0i}$ of $P''$, a solution $solP'$ realizing the optimum
$t_{opt}^{P'}$ of $P'$, and the optimum $t_{opt}^{P'}$  itself: $solP'=(0,b_1,\ldots ,b_n)$ and $t_{opt}^{P'}=dur(solP')$. Let us now compare $t_{opt}^{P'}$ with $OLB(P)$:
\begin{enumerate}
  \item Case 1: the latency of a solution is part of its duration. In this case, $OLB(P)=\displaystyle max_{i=1}^{n}(a_i+d_i)$ and the duration of a solution
	$s=(t_0,t_1,\ldots ,t_n)$ is $dur(s)=\displaystyle max_{i=1}^{n}(t_i+d_i)-t_0$. Thus $t_{opt}^{P'}=dur(solP')=\displaystyle max_{i=1}^{n}(b_i+d_i)$.
	Given that $a_i\leq b_i$, for all $i$ from 1 to n, because $P''$ is a refinement of $P$, we clearly have $t_{opt}^{P'}\geq OLB(P)$.
  \item Case 2: the latency of a solution is not part of its duration. In this case, $OLB(P)=\displaystyle max_{i=1}^{n}(a_i+d_i)-\displaystyle max(\{0\}\cup\{rd_i:\mbox{ $i$
	from $1$ to $n$ and $rd_i$ finite}\})$ and the duration of a solution $s=(t_0,t_1,\ldots ,t_n)$ is $dur(s)=\displaystyle max_{i=1}^{n}(t_i+d_i)-\displaystyle min_{i=1}^{n}t_i$.
Thus $t_{opt}^{P'}=dur(solP')=\displaystyle max_{i=1}^{n}(b_i+d_i)-\displaystyle min_{i=1}^{n}b_i$:
	\begin{enumerate}
	  \item Given that $a_i\leq b_i$, for all $i$ from 1 to n, because $P''$ is a refinement of $P$, we clearly have $\displaystyle max_{i=1}^{n}(b_i+d_i)\geq\displaystyle max_{i=1}^{n}(a_i+d_i)$.
	  \item The quantity $\displaystyle max(\{0\}\cup\{rd_i:\mbox{ $i$ from $1$ to $n$ and $rd_i$ finite}\})$ stands for the latest release date of the $n$ tasks (equal to 0 if no
	release date is given), whereas the quantity $\displaystyle min_{i=1}^{n}b_i$ stands for the starting date of the very first task of the solution realizing the optimum of $P'$. Clearly,
	$\displaystyle min_{i=1}^{n}b_i\leq \displaystyle max(\{0\}\cup\{rd_i:\mbox{ $i$ from $1$ to $n$ and $rd_i$ finite}\})$, for otherwise one would be able to use a global translation to
	construct from $solP'$ another solution $solP'2$ of $P'$ whose duration is strictement smaller than that of $solP'$, which would contradict the fact that $dur(solP')$ is the optimum of
	$P'$: $solP'2=solP'-d=(0,b_1-d,\ldots ,b_i-d,\ldots ,b_n-d)$, with $d=\displaystyle min_{i=1}^{n}b_i-\displaystyle max(\{0\}\cup\{rd_i:\mbox{ $i$ from $1$ to $n$ and
	$rd_i$ finite}\})$. A global translation of a solution remains a solution, as long as the release dates and the due dates are not violated: the other constraints (precedence constraints
	and disjunctive constraints) are on distances separating variables, and their satisfiability is invariant by global translation. \cqfd
	\end{enumerate}
\end{enumerate}
The result of Theorem \ref{minimality2}, $t_{opt}^P\geq OLB(P)$, is used by the scheduler to backtrack whenever $z\leq OLB(P)$ (line \ref{zline5}), for it has then the guarantee that the current
refinement $P$ cannot improve the global optimum $z$. Furthermore, whenever the \mbox{wbdAC-3} filtering leads to an $\stp$ refinement $P$ (line \ref{zline14}), $P$ is bdArc-Consistent, for an $\stp$
closed by wbdArc-Consistency is also closed by bdArc-Consistency (the convex closure of a convex set is the set itself, meaning that weak composition, when convexity is met, is equal to classical
composition); and in this case:
\begin{enumerate}
  \item from Theorem \ref{minimality1}, a solution $solP$ realizing the optimum of refinement $P$ is given by the lower bounds of the binarized domains $P_{0i}$, $i$ from 1 to n, and the optimum
	$optP$ of $P$ is equal to the duration $dur(solP)$ of $solP$ (lines \ref{zline142} and \ref{zline143});
  \item the global optimum $z$ and the solution $s$ realizing it are updated if $z>optP$ (line \ref{zline145}). 
\end{enumerate}
\subsection{\mbox{bdAC-3} as the filtering procedure}\label{schedulerSubs2}
The relaxation step of the \mbox{bdAC-3} algorithm, consisting of a call of the $\mbox{REVISE}(i,j)$ procedure of Figure \ref{TheREVISEProcedure}, and reducing mainly to the
path-consistency operation $P_{0i}=P_{0i}\cap P_{0j}\otimes P_{ji}$, may result in a fragmentation of the binarized domain $P_{0i}$, as illustrated by Example
\ref{example}(\ref{fragprobright}-\ref{fragprobright2}) and the accompanying Figure \ref{fragprob}(right). To avoid such a fragmentation, we used in the general $\tcsp$ solver of Section \ref{sthree}, a weak
version of the \mbox{bdAC-3} algorithm, \mbox{wbdAC-3}, as the filtering procedure. Given that the disjunctive binarized domains are also instantiated in the $\tcsp$ solver (Figure
\ref{TheTCSPSolver}, line \ref{xline4}), avoiding the fragmentation of the binarized domains means avoiding a blowup of the branching factor (the maximal number of choices when instantiating
an entry of the matrix representation).

With the job shop scheduler (Figure \ref{jsscheduler}), however, the disjunctive binarized domains are not instantiated (see line \ref{zline8}). It is true that even if we included the instantiation
of disjunctive binarized domains in the scheduler (replacement of $1\leq i<j\leq n$ with $0\leq i<j\leq n$ in line \ref{zline8}), this would have no effect, given that the binarized domains are
initially convex, and \mbox{wbdAC-3} does not fragment convex sets. But the happy ending is that, we can replace \mbox{wbAC-3} with \mbox{bdAC-3} in the solver and keep the
rest as it is (in particular, keep $1\leq i<j\leq n$ as it is in line \ref{zline8}, so that the instantiation of disjunctve binarized domains remains excluded), without affecting the main results.
At the same time, the main advantage of excluding the instantiation of disjunctive binarized domains is computational, for the replacement of \mbox{wbdAC-3}
with \mbox{bdAC-3} leads to no risk of a blowup of the branching factor, which remains equal to $2$, given that the disjunctive entries that do not correspond to binarized domains,
are all of the form $(-\infty ,a]\cup [b,+\infty )$, with $a<0<b$. Theorem \ref{bdac3f} below shows that replacing \mbox{wbdAC-3} with \mbox{bdAC-3} and keeping excluded the
instantiation of disjunctive binarized domains, keeps the solver exact; this consists of replacing lines \ref{zline5}, \ref{zline6}, \ref{zline7} and \ref{zline8} of the scheduler (Figure
\ref{jsscheduler}) as follows:
\begin{enumerate}
  \item[\ref{zline5}.] {\bf if }(not \mbox{bdAC-3}$(P)$ {\bf or }$z\leq OLB (P))$ return
  \item[\ref{zline6}.] {\bf else}
  \item[\ref{zline7}.]\hskip 0.3cm {\bf if}($P$ has disjunctive edges other than binarized domains)\{\%beginif2
  \item[\ref{zline8}.]\hskip 0.6cm select such a disjunctive edge $(X_i,X_j)$, with $1\leq i<j\leq n$;
\end{enumerate}
We first need a new material.
\begin{definition}\label{ctcsp}The set of (topologically) closed subsets of $\BBR$ is the smallest set verifying the following:
\begin{enumerate}
  \item for all $a\in\BBR$, $(-\infty ,a]$ and $[a,+\infty )$ are closed subsets of $\BBR$;
  \item for all $a,b\in\BBR$, with $a\leq b$, $[a,b]$ is a closed subset of $\BBR$;
  \item if $A$ and $B$ are closed subsets of $\BBR$ then so is $A\cup B$.
\end{enumerate}
A closed $\tcsp$, or $c\tcsp$, is a $\tcsp$ whose constraints are of the form $(X_j-X_i)\in C_{ij}$, $C_{ij}$ being a closed subset of $\BBR$. A \mbox{STAR} $\tcsp$
\cite{SchwalbD97a} is a $c\tcsp$ of which all constraints $(X_j-X_i)\in C_{ij}$ such that $0\notin\{i,j\}$ are convex.
\end{definition}
\begin{theorem}\label{bdac3f}Let $P$ be a scheduling $\tcsp$ and $P'$ a refinement of $P$ that is a \mbox{STAR} $\tcsp$. If $P'$ is bd-arc consistent then $s=(0,lowerB(P^{'}_{01}),\ldots ,lowerB(P^{'}_{0n}))$
realizes the optimum of $P'$, and the optimum of $P'$ is $dur(s)$.
\end{theorem}
{\bf Proof: }Let $P$ and $P'$ be as stated in the theorem, and denote by $P''$ the convex closure of $P'$. Clearly, $P''$ differs from $P'$ only at the binarized domains: $P^{''}_{0i}=cc(P^{'}_{0i})$. Furthermore,
$P''$ is a bd-arc consistent scheduling $\stp$. Therefore, $s=(0,lowerB(P^{''}_{01}),\ldots ,lowerB(P^{''}_{0n}))$ is a solution realizing the optimum of $P''$, and the optimum of $P''$ is $dur(s)$ (Theorem
\ref{minimality1}). But gven that $P'$ is a \mbox{STAR} $\tcsp$, we have $lowerB(P^{''}_{0i})=lowerB(P^{'}_{0i})$, for all $i$ from 1 to n.Therefore $s=(0,lowerB(P^{'}_{01}),\ldots ,lowerB(P^{'}_{0n}))$ is a
solution realizing the optimum of $P'$, and the optimum of $P'$ is $dur(s)$. \cqfd
\section{Related work and additional adaptations}\label{relatedwork}
We now compare our results to related work, mainly work on arc consistency and path consisteny for $\tcsps$. We also provide additional adaptations to $\tcsps$, which
will concern the other local consistency algorithms in \cite{Mackworth77a}. In particular, this corrects an existing adaptation to $\tcsps$ of \mbox{PC-2}
\cite{DechterMP91a,SchwalbD97a}, whose non-termination is shown with an example.

First, it is not the main issue of the paper, but it is worth reminding the definition of a $\tcsp$ being path-consistent:
\begin{definition}[path-consistency]Let $P$ be an n+1-variable $\tcsp$. $P$ is said to be path-consistent if for all $i,j,k\in\{0,1,\ldots ,n\}$, the following holds: $P_{ij}\subseteq P_{ik}\otimes P_{kj}$.
\end{definition}
\subsection{Arc consistency}
Related work on arc-consistency for $\tcsps$ include \cite{CervoniCO94a,CestaO96a,KongLL18a,PlankenWK08a}, but the focus has always been restricted to
$\stps$. The most related remains \cite{KongLL18a}, which defines an arc-consistency
algorithm for $\stps$, $\acstp$, which can be seen as an adaptation of Bellman-Ford-Moore one-to-all shortest paths algorithm \cite{Bellman58a,FordF62a,Moore59a}; this
can also be seen as an adaptation to $\stps$ of Mackworth's \cite{Mackworth77a} arc-consistency algorithm \mbox{AC-1}. Beyond the fact that
the algorithm is defined only for $\stps$, the other criticism is that for applications needing incrementality such as planning, which seems to have been the authors' targetted application,
it is clear that the use of an adaptation of Mackworth's \cite{Mackworth77a} \mbox{AC-3} instead of \mbox{AC-1}, in case of arc-consistency, and an
adaptation of Mackworth's \cite{Mackworth77a} \mbox{PC-2} instead of \mbox{PC-1}, in case of path-consistency, is much better, mainly because these use
a queue allowing them, when new knowledge comes in, to continue, incrementally, on what had been done before, and not to redo it. Allen's well-known Interval Algebra
\cite{Allen83a} in general, and its convex part in particular, whose most targetted application is planning, uses an adaptation of \mbox{PC-2}.

\subsection{Path consistency}
The most related work, worth comparing, is Decther et al.'s \cite{DechterMP91a}, where, in particular, the authors show the equivalence between, on one hand, applying path-consistency
to an $\stp$ and, on the other, applying an all-to-all shortest paths algorithm, such as Floyd-Warshall's \cite{AhoHU76a,PapadimitriouS82a}, to the corresponding distance graph. Dechter et
al. \cite{DechterMP91a}, as explained in the introduction, didn't focus on node- and arc-consistencies, because the adding of an ''origin of the world'' variable led them to the
conclusion that a $\tcsp$ was already node- and arc-consistent. In the present work, we have looked at the constraints between the ''origin of the world'' variable and the other
variables, as the binarized domains of these other variables. This led us to the definition of the notion of binarized-domains Arc-Consistency, which is computationally one factor
less expensive. We showed, in particular, that applying bdArc-Consistency to an $\stp$ was equivalent to applying a one-to-all all-to-one shortest paths algorithm to the corresponding rooted
distance graph. The bdArc-Consistency algorithm we have defined, \mbox{bdAC-3}, detects negative circuits reachable from $X_0$, or from which $X_0$ is reachable.

Another related work is Schwalb and Dechter's \cite{SchwalbD97a}, where, in particular, the authors propose the use of weak versions of path-consistency
as filtering procedures of general $\tcsp$ solvers. One of these weaker versions is ULT (Upper-Lower Tightning), which applies path-consistency to the convex closure of the original $\tcsp$ $P$
and intersects the result with $P$. We have discussed the use of a similar idea in a general $\tcsp$ solver with the filtering procedure consisting of a weak version, \mbox{wbdAC-3}, of the
binarized-domains Arc-Consistency algorithm \mbox{bdAC-3} proposed in this work: composition is replaced with weak
composition.
\subsection{Adaptation to $\tcsps$ of \mbox{AC-1}, \mbox{PC-1} and \mbox{PC-2}}
\begin{figure}
\begin{enumerate}
  \item[] {\bf Input: }The matrix representation of a $\tcsp$ $P$.
  \item[] {\bf Output: }False, indicating that inconsistency of $P$ has been detected; or true, indicating that $P$ has been made bdArc-consistent.
  \item[] {\bf procedure }$\mbox{bdAC-1}(P)$\{
  \item\label{bline1}\hskip 0.2cm $Q=\{(i,j):\mbox{ }P\mbox{ }has\mbox{ }a\mbox{ }constraint\mbox{ }on\mbox{ }X_i\mbox{ }and\mbox{ }X_j,$
  \item[]\hskip 1.2cm $\mbox{ }i*j\not =0,\mbox{ }i\not =j\}$;
  \item\label{bline2}\hskip 0.2cm {\bf repeat}
  \item\label{bline3}\hskip 0.4cm CHANGE=false;
  \item\label{bline4}\hskip 0.4cm {\bf for} each $(i,j)\in Q$
  \item\label{bline5}\hskip 0.6cm if $\mbox{REVISE}(i,j)$\{\%begin if1
  \item\label{bline7}\hskip 0.8cm if($P_{0i}=\emptyset$)return false;
  \item\label{bline8}\hskip 0.8cm CHANGE=true
  \item\label{bline8}\hskip 0.8cm \}\%end if1
  \item\label{bline11}\hskip 0.2cm {\bf until }$\neg$CHANGE
  \item\label{bline12}\hskip 0.2cm return true
  \item\label{bline13}\hskip 0.2cm \}\%end bdAC1
\end{enumerate}
\caption{The binarized-domains Arc-Consistency algorithm \mbox{bdAC-1}.}\label{bdAC1tcsp}
\end{figure}

\begin{figure}
\begin{enumerate}
\item[] {\bf Input: }The matrix representation of a $\tcsp$ $P$.
  \item[] {\bf Output: }False, indicating that inconsistency of $P$ has been detected; or true, indicating that $P$ has been made path-consistent.
  \item[] {\bf procedure }$\mbox{PC-1}(P)$\{
  \item\label{cline1}\hskip 0.2cm {\bf repeat}
  \item\label{cline2}\hskip 0.4cm CHANGE=false;
  \item\label{cline3}\hskip 0.4cm {\bf for }($k=0$ to $n$)
  \item\label{cline4}\hskip 0.6cm {\bf for }($i,j=0$ to $n$)\{
  \item\label{cline5}\hskip 0.8cm $temp=P_{ij}\cap P_{ik}\otimes P_{kj}$;
  \item\label{cline6}\hskip 0.8cm {\bf if}($temp=\emptyset$)return false;
  \item\label{cline7}\hskip 0.8cm {\bf if}($temp\not = P_{ij}$)\{$P_{ij}=temp$;
  \item\label{cline72}\hskip 3.6cm CHANGE=true\}
  \item\label{cline8}\hskip 0.8cm \}
  \item\label{cline9}\hskip 0.2cm{\bf until }$\neg$CHANGE
  \item\label{cline10}\hskip 0.2cm return true
  \item\label{cline11}\hskip 0.2cm \}\%end PC1
\end{enumerate}
\caption{The Path-Consistency algorithm \mbox{PC-1}.}\label{PC1tcsp}
\end{figure}
\begin{figure}[t]
\begin{enumerate}
  \item[]{\bf procedure }$\mbox{REVISE\_PC-2}(i,k,j)$\{
  \item\label{previse1}\hskip 0.4cm  DELETE=false;
  \item\label{previse2}\hskip 0.4cm $temp=P_{ij}\cap P_{ik}\otimes P_{kj}$;
  \item\label{previse3}\hskip 0.4cm if($temp\not = P_{ij}$)\{
  \item\label{previse4}\hskip 0.6cm if[$temp\not =\emptyset$ and $(lowerB(temp)>-\mbox{path-lb}(P)$
  \item[]\hskip 1cm or $upperB(temp)<\mbox{path-lb}(P))] temp=\emptyset$;
  \item\label{previse5}\hskip 0.6cm $P_{ij}=temp$;
  \item\label{previse52}\hskip 0.6cm $P_{ji}=temp^\smile$;
  \item\label{previse6}\hskip 0.6cm DELETE=true
  \item\label{previse7}\hskip 0.6cm \}
  \item\label{previse8}\hskip 0.4cm return DELETE
  \item\label{previse9}\hskip 0.4cm \}
\end{enumerate}
\caption{The procedure $\mbox{REVISE\_PC-2}$ used by the path-consistency algorithm $\mbox{PC-2}$ of Figure \ref{PC2tcsp}.}\label{revisePC2}
\end{figure}
\begin{figure}
\begin{enumerate}
  \item[]\label{aline1} {\bf Input: }The matrix representation of a $\tcsp$ $P$.
  \item[]\label{aline2} {\bf Output: }False, indicating that inconsistency of $P$ has been detected; or true, indicating that $P$ has been made path-consistent.
  \item[] {\bf procedure }$\mbox{PC-2}(P)$\{
  \item\label{aline3}\hskip 0.2cm $Q=\{(i,k,j):\mbox{ }(i<j)\mbox{ and }(k\notin\{i,j\})\mbox{ and }P_{ik}\subset\BBR\mbox{ and }P_{kj}\subset\BBR\}$;
  \item\label{aline4}\hskip 0.2cm $Empty\_domain=false$;
  \item\label{aline5}\hskip 0.2cm {\bf while }$(Q\not =\emptyset$\mbox{ and (not }Empty\_domain))\{
  \item\label{aline6}\hskip 0.4cm select and delete a triple $(i,k,j)$ from $Q$;
  \item\label{aline7}\hskip 0.4cm {\bf if }$\mbox{REVISE\_PC-2}(i,k,j)$
  \item\label{aline8}\hskip 0.6cm if $(P_{ij}=\emptyset)$ $Empty\_domain=true$
  \item\label{aline9}\hskip 0.6cm else $Q=Q\cup$
  \item[]\hskip 0.8cm $\{(i,j,m):\mbox{ $(i<m)$ and $(m\not =j)$ and $P_{jm}\subset\BBR$}\}\cup$
  \item[]\hskip 0.8cm $\{(m,i,j):\mbox{ $(m<j)$ and $(m\not =i)$ and $P_{mi}\subset\BBR$}\}$
   \item\hskip 0.4cm \} \%endwhile
  \item\label{aline10}\hskip 0.2cm return(not $Empty\_domain$)
  \item\label{aline11}\hskip 0.2cm \} \%end PC2
\end{enumerate}
\caption{The path-consistency algorithm $\mbox{PC-2}$.}\label{PC2tcsp}
\end{figure}
We now present an adaptation to $\tcsps$ of the other local-consistency algorithms in \cite{Mackworth77a} and consider their termination. We show, in particular, that an existing adaptation to $\tcsps$
of \mbox{PC-2} \cite{DechterMP91a,SchwalbD97a} is not guaranteed to always terminate.

More specifically, in addition to the \mbox{bdAC-3} algorithm, we provide in Figures \ref{bdAC1tcsp}, \ref{PC1tcsp} and \ref{PC2tcsp} adaptations to $\tcsps$ of the other three local-consistency algorithms in \cite{Mackworth77a}:
the arc-consistency algorithm \mbox{AC-1} and the two path-consistency algorithms \mbox{PC-1} and \mbox{PC-2}. The adaptations are named, respectively, \mbox{bdAC-1},
\mbox{PC-1} and \mbox{PC-2}. Like \mbox{bdAC-3}, when applied to an $\stp$, \mbox{bdAC-1} detects, if any, negative circuits reachable from $X_0$, or from which $X_0$
is reachable, in the corresponding rooted distance graph; it makes the $\stp$ bdArc-Consistent otherwise. \mbox{PC-1} and \mbox{PC-2}, when applied to an $\stp$, both detect, if any,
negative circuits  in the corresponding distance graph; they make the $\stp$ path-consistent otherwise. Both of \mbox{bdAC-1} and \mbox{PC-2} use the condition on
negative circuits one can extract from Theorem \ref{circuitfreepath} (line \ref{revise42} of the $\mbox{REVISE}$ procedure of Figure \ref{TheREVISEProcedure}, for \mbox{bdAC-1};
line \ref{previse4} of the $\mbox{REVISE\_PC-2}$ procedure of Figure \ref{revisePC2}, for \mbox{PC-2}).
\begin{remark}\label{remark4}
\begin{enumerate}
  \item We refer to the procedure $\mbox{REVISE\_PC-2}$ of Figure \ref{revisePC2} without its line \ref{previse4} as $\mbox{REVISE\_PC-2}^-$.
  \item  We refer to the algorithm $\mbox{bdAC-1}$ (Figure \ref{bdAC1tcsp}) in which procedure $\mbox{REVISE}$ is replaced with $\mbox{REVISE}^-$ as
	$\mbox{bdAC-1}^-$, and to the algorithm $\mbox{PC-2}$ (Figure \ref{PC2tcsp}) in which procedure $\mbox{REVISE\_PC-2}$ is replaced with
	$\mbox{REVISE\_PC-2}^-$ as $\mbox{PC-2}^-$.
\end{enumerate}
\end{remark}
\subsection{Termination of the adaptations}
\mbox{PC-1} proceeds by passes, and each pass considers all possible triples of variables. It is a "literal" adaptation from \cite{Mackworth77a}, and is guaranteed to terminate and to
detect, when applied to an $\stp$, the presence of negative circuits in the corresponding distance graph. As illustrated by Example \ref{nontermination}, this is not the case for $\mbox{bdAC-1}^-$ and $\mbox{PC-2}^-$, which
are "literal" adaptations of Mackworth's \cite{Mackworth77a} \mbox{AC-1} and \mbox{PC-2}, respectively, and which, deprived of line \ref{revise42} of Procedure $\mbox{REVISE}$
for the former, and of line \ref{previse4} of Procedure $\mbox{REVISE\_PC-2}$ for the latter, have no guarantee to terminate even when the input is an $\stp$.

\begin{example}\label{nonterminationexample}[non termination of $\mbox{bdAC-1}^-$ and $\mbox{PC-2}^-$]\label{nontermination}
We consider the $\stp$ $P=(X,C)$ with:
\begin{enumerate}
  \item $X=\{X_0,X_1,X_2,X_3\}$ and
  \item $C=\{c_1:(X_1-X_0)\in [30,+\infty ),c_2:(X_2-X_1)\in [-20,-10],c_3:(X_3-X_1)\in (-\infty ,4],c_4:(X_3-X_2)\in [40,50]\}$.
\end{enumerate}
\begin{figure}
\centerline{\includegraphics[width=3cm]{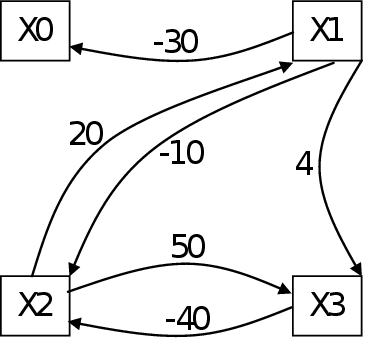}}
\caption{The distance graph of the $\stp$ of Example \ref{nontermination}.}\label{stpdistancegraph2}
\end{figure}
The rooted distance graph of $P$ (see Figure \ref{stpdistancegraph2}) does have a negative ciruit, $(X_1,X_3,X_2,X_1)$, that
is not reachable from $X_0$, but from which $X_0$ is reachable. \mbox{bdAC-1} and $\mbox{bdAC-1}^-$ proceed by passes, each of which is guaranteed to terminate
since it considers at most once each ordered pair $(X_i,X_j)$ such that there is a constraint on $X_i$ and $X_j$. Each of the two versions of bdArc-Consistency terminates
when, either a binarized domain becomes empty, or a whole pass makes no change in the binarized domains. But for the version $\mbox{bdAC-1}^-$, the order in which
the pairs in its queue are propagated might make it impossible for the two terminating conditions to occur. Initially, the binarized domains are as follows:
$P_{01}=[30,+\infty )$, $P_{02}=\BBR$ and $P_{03}=\BBR$. For each of the two versions, we suppose that each of its passes propagates the pairs on which there is a
constraint in the order $(X_1,X_2)$, $(X_1,X_3)$, $(X_2,X_1)$, $(X_2,X_3)$, $(X_3,X_1)$, $(X_3,X_2)$. For both versions, the evolution of the binarized domains for the first three passes
are as follows:
\begin{enumerate}
  \item after Pass 1: $P_{01}=[30,+\infty )$, $P_{02}=[10,+\infty )$ and $P_{03}=[50,+\infty )$
  \item after Pass 2: $P_{01}=[46,+\infty )$, $P_{02}=[26,+\infty )$ and $P_{03}=[66,+\infty )$
  \item after Pass 3: $P_{01}=[62,+\infty )$, $P_{02}=[42,+\infty )$ and $P_{03}=[82,+\infty )$
\end{enumerate}
This means that each pass increases the (finite) lower bound of each binarized domain by 16, which corresponds to the absolute value of the weight of the negative circuit
$(X_1,X_3,X_2,X_1)$. Given that $\mbox{path-lb}(P)=-80$, \mbox{bdAC-1} will detect the negative circuit after the third pass. However, $\mbox{bdAC-1}^-$
will fail to do so, and will not terminate.

Let's now look at how the two adaptations \mbox{PC-2} and $\mbox{PC-2}^-$ of Mackworth's \cite{Mackworth77a} \mbox{PC-2} will behave on the $\stp$. Keep in mind that their
unique difference lies in line \ref{previse4} of Procedure $\mbox{REVISE\_PC-2}$, which is absent in $\mbox{PC-2}^-$. Both versions will initialize their queue as follows:\\
$Q=\{(X_0,X_1,X_2),(X_0,X_1,X_3),(X_1,X_2,X_3)\}$. We take the triple $(X_0,X_1,X_2)$ for propagation. The resulting modifications are as follows:
\begin{enumerate}
  \item $P_{02}=[10,+\infty )$, $P_{20}=(-\infty ,-10]$, 
  \item $Q=\{(X_0,X_1,X_3),(X_1,X_2,X_3),(X_0,X_2,X_1),$
  \item[] $(X_0,X_2,X_3),(X_1,X_0,X_2)\}$
\end{enumerate}
We then take the triple $(X_0,X_2,X_3)$ for propagation. The modifications are as follows:
\begin{enumerate}
  \item $P_{03}=[50,+\infty )$, $P_{30}=(-\infty ,-50]$,
  \item $Q=\{(X_0,X_1,X_3),(X_1,X_2,X_3),(X_0,X_2,X_1),$
  \item[] $(X_1,X_0,X_2),(X_0,X_3,X_1),(X_0,X_3,X_2),$
  \item[] $(X_1,X_0,X_3),(X_2,X_0,X_3)\}$
\end{enumerate}
The situation of the binarized domains is now the same as that after Pass 1 of \mbox{bdAC-1}. We now repeat the following sequence: propagate triple $(X_0,X_3,X_1)$,
then triple $(X_0,X_1,X_2)$, finally triple $(X_0,X_2,X_3)$. Each pass of this repeat loop will increase  by 16 the lower bound of each of the binarized domains. After the second
pass, the binarized domain $P_{03}$ becomes $[82,+\infty )$, and \mbox{PC-2} will then detect the presence of a negative circuit, and terminate.
$\mbox{PC-2}^-$, however, will keep repeating the loop indefinitely, and will not terminate. The details of the first two passes of the repeat loop are given in Figure \ref{repeatloop}.
\begin{figure}
\begin{center}
           $
           \begin{array}{|l||l|}\hline
           \mbox{Pass}      &\mbox{Modifications}\\  \hline\hline
           1&P_{01}=[46,+\infty ),\mbox{ }P_{10}=(-\infty ,-46],\\
             &Q=\{(X_0,X_1,X_3),(X_1,X_2,X_3),(X_0,X_2,X_1),\\
             &\hskip 0.9cm (X_1,X_0,X_2),(X_0,X_3,X_2),(X_1,X_0,X_3),\\
             &\hskip 0.9cm (X_2,X_0,X_3),(X_0,X_1,X_2)\}\\  \cline{2-2}
            &P_{02}=[26,+\infty ),\mbox{ }P_{20}=(-\infty ,-26],\\
             &Q=\{(X_0,X_1,X_3),(X_1,X_2,X_3),(X_0,X_2,X_1),\\
             &\hskip 0.9cm (X_1,X_0,X_2),(X_0,X_3,X_2),(X_1,X_0,X_3),\\
             &\hskip 0.9cm (X_2,X_0,X_3),(X_0,X_2,X_3)\}\\  \cline{2-2}
            &P_{03}=[66,+\infty ),\mbox{ }P_{30}=(-\infty ,-66],\\
             &Q=\{(X_0,X_1,X_3),(X_1,X_2,X_3),(X_0,X_2,X_1),\\
             &\hskip 0.9cm (X_1,X_0,X_2),(X_0,X_3,X_2),(X_1,X_0,X_3),\\
             &\hskip 0.9cm (X_2,X_0,X_3),(X_0,X_3,X_1)\}\\    \hline
           2&P_{01}=[62,+\infty ),\mbox{ }P_{10}=(-\infty ,-62],\\
             &Q=\{(X_0,X_1,X_3),(X_1,X_2,X_3),(X_0,X_2,X_1),\\
             &\hskip 0.9cm (X_1,X_0,X_2),(X_0,X_3,X_2),(X_1,X_0,X_3),\\
             &\hskip 0.9cm (X_2,X_0,X_3),(X_0,X_1,X_2)\}\\  \cline{2-2}
        &P_{02}=[42,+\infty ),\mbox{ }P_{20}=(-\infty ,-42],\\
             &Q=\{(X_0,X_1,X_3),(X_1,X_2,X_3),(X_0,X_2,X_1),\\
             &\hskip 0.9cm (X_1,X_0,X_2),(X_0,X_3,X_2),(X_1,X_0,X_3),\\
             &\hskip 0.9cm (X_2,X_0,X_3),(X_0,X_2,X_3)\}\\  \cline{2-2}
            &P_{03}=[82,+\infty ),\mbox{ }P_{30}=(-\infty ,-82],\\
             &Q=\{(X_0,X_1,X_3),(X_1,X_2,X_3),(X_0,X_2,X_1),\\
             &\hskip 0.9cm (X_1,X_0,X_2),(X_0,X_3,X_2),(X_1,X_0,X_3),\\
             &\hskip 0.9cm (X_2,X_0,X_3),(X_0,X_3,X_1)\}\\    \hline
           \end{array}
           $
\end{center}
\caption{The first two passes of the repeat loop of Example \ref{nonterminationexample}.}\label{repeatloop}
\end{figure}
\end{example}
We now summarize the discussion:
\begin{theorem}
\mbox{bdAC-1}, \mbox{PC-1} and \mbox{PC-2} (Figures \ref{bdAC1tcsp}, \ref{PC1tcsp} and \ref{PC2tcsp}) always terminate. \cqfd
\end{theorem}
\begin{theorem}
$\mbox{bdAC-1}^-$ and $\mbox{PC-2}^-$ are not guaranteed to terminate. \cqfd
\end{theorem}
\section{Experimental comparison of behaviors on $\stps$}\label{expcmpstps}
Our first experimental comparison is dedicated to the behavior on $\stps$ of \mbox{bdAC-3} with that of the arc-consistency algorithm $\acstp$ for $\stps$ in \cite{KongLL18a}.
\subsection{Implementation details}
Our experiments, of this and the next sections, were implemented in free, non-commercial swi-prolog (AMD64, Multi-threaded, version 9.0.4) and carried
out on a 64-bit HP Pavilion dv7 Notebook PC with an Intel Core i7 processor, a 2.3 GHz frequency per CPU and 4 GB memory.
\subsection{Further focus on $\stps$}

\begin{figure}
\begin{enumerate}
  \item[] {\bf Input: }The matrix representation of an n+1-variable $\stp$ $P$.
  \item[] {\bf Output: }False, indicating that inconsistency of $P$ has been detected; or true, indicating that $P$ has been made bdArc-consistent.
  \item[] {\bf procedure }$\mbox{bdAC-1stp}(P)$\{
  \item\label{bline1}\hskip 0.2cm $Q=\{(i,j):\mbox{ }P\mbox{ }has\mbox{ }a\mbox{ }constraint\mbox{ }on\mbox{ }X_i\mbox{ }and\mbox{ }X_j,\mbox{ }i*j\not =0,\mbox{ }i\not =j\}$
  \item\label{bline12}\hskip 0.2cm $COUNT =0 $;
  \item\label{bline2}\hskip 0.2cm {\bf repeat}
  \item\label{bline3}\hskip 0.4cm COUNT = COUNT+1 ; CHANGE=false;
  \item\label{bline4}\hskip 0.4cm {\bf for} each $(i,j)\in Q$
  \item\label{bline5}\hskip 0.6cm if $\mbox{REVISE}^-(i,j)$\{\%begin if1
  \item\label{bline7}\hskip 0.8cm if($P_{0i}=\emptyset$)return false;
  \item\label{bline8}\hskip 0.8cm CHANGE=true\}\%end if1
  \item\label{bline11}\hskip 0.2cm {\bf until }$\neg$CHANGE or COUNT = n+1
  \item\label{bline12}\hskip 0.2cm {\bf if }(CHANGE {\bf and }COUNT = n+1) {\bf then return }false {\bf else }return true
  \item\label{bline13}\hskip 0.2cm \}\%end bdAC-1stp
\end{enumerate}
\caption{The binarized-domains Arc-Consistency algorithm \mbox{bdAC-1stp}.}\label{bdAC1stp}
\end{figure}
\begin{figure}
\begin{enumerate}
  \item[] {\bf procedure }$\mbox{bdAC-3stp}(P)$\{
  \item\label{bline1}\hskip 0.2cm $Q=\{(i,j):\mbox{ }P\mbox{ }has\mbox{ }a\mbox{ }constraint\mbox{ }on\mbox{ }X_i\mbox{ }and\mbox{ }X_j,\mbox{ }i*j\not =0,\mbox{ }i\not =j\}$
  \item\label{bline12}\hskip 0.2cm $COUNT =0 $;
  \item\label{bline2}\hskip 0.2cm {\bf repeat}
  \item\label{bline3}\hskip 0.4cm COUNT = COUNT+1 ; CHANGE=false; $\mbox{Qnextpass}=\emptyset$;
  \item\label{bline4}\hskip 0.4cm {\bf while}$(Q\not =\emptyset )\{$
  \item\label{bline5}\hskip 0.6cm take a pair $(i,j)$ from $Q$ for propagation; $Q=Q\backslash\{(i,j)\}$;
  \item\label{bline5}\hskip 0.6cm if $\mbox{REVISE}^-(i,j)$\{\%begin if1
  \item\label{bline7}\hskip 0.8cm if($P_{0i}=\emptyset$)return false;
  \item\label{bline1}\hskip 0.8cm $\mbox{Qnextpass}=\mbox{Qnextpass}\cup\{(k,i):$
  \item[]\hskip 1.6cm $P\mbox{ }has\mbox{ }a\mbox{ }constraint\mbox{ }on\mbox{ }X_k\mbox{ }and\mbox{ }X_i,(k,i)\notin Q,k\notin\{0,i,j\}\}$
  \item\label{bline8}\hskip 0.8cm CHANGE=true\}\}\%end if1 \%end while
  \item\label{bline82}\hskip 0.4cm $Q=\mbox{Qnextpass}$;
  \item\label{bline11}\hskip 0.2cm {\bf until }$\neg$CHANGE {\bf or } COUNT = n+1
  \item\label{bline12}\hskip 0.2cm {\bf if }(CHANGE {\bf and }COUNT = n+1) {\bf then return }false {\bf else return} true
  \item\label{bline13}\hskip 0.2cm \}\%end bdAC-3stp
\end{enumerate}
\caption{The binarized-domains Arc-Consistency algorithm \mbox{bdAC-3stp}.}\label{bdAC3stp}
\end{figure}

$\acstp$ is the very first arc-consistency algorithm for $\stps$, and its behavior on $\stps$ has been, unsurprisingly, advantageously compared in \cite{KongLL18a} to that of $P^3C$
\cite{PlankenWK08a}, one of many weak versions of path-consistency used for $\stps$ in planning \cite{PlankenWK08a,PlankenWY10a,XuC03a}. $\acstp$, again, is basically an adaptation of Mackworth's
\mbox{AC-1} algorithm or, equivalenty, of Bellman-Ford-Moore's algorithm \cite{Bellman58a,FordF62a,Moore59a}. We first highlight the main weaknesses of $\acstp$, of which \mbox{bdAC-3}
does not suffer:
\begin{enumerate}
  \item lack of incrementality, as alluded to in the introduction.
  \item lack of general convexity : the constraints of an $\stp$ are assumed in \cite{KongLL18a} to be of the form $(X_j-X_i)\in C_{ij}$, with $C_{ij}$ of the form $[a,b]$, $a$
	and $b$ in $\BBR$. This means that $C_{ij}$ should have finite (left and right) bounds, and be topologically closed.
  \item the domains of the variables also suffer from lack of general convexity: a domain has the form $[a,b]$, $a$ and $b$ in $\BBR$.
  \item the fact that $\acstp$ is restricted to $\stps$ makes its use as a true filtering procedure in search algorithms for general $\tcsps$ difficult. Indeed, the only way for such a use, as a
	verification procedure rather than a filtering procedure,  is with a naive, generate-and-test, solution search algorithm, that would keep generating an $\stp$ refinement until one that is
	consistent is found (consistency), or the whole search space is explored without finding a consistent $\stp$ refinement (inconsisteny), with the consistency of a generated $\stp$ verified with
          $\acstp$.
\end{enumerate}
The use of line \ref{revise42} in the \mbox{REVISE} procedure of Figure \ref{TheREVISEProcedure}, meant to check whether the bounds of $temp$ both lie in the interval
$[\mbox{path-lb}(P),-\mbox{path-lb}(P)]$, is not needed when the input $\tcsp$ is an $\stp$. It can be replaced with $\mbox{REVISE}^-$. The use, indeed, of the genuine idea in
Bellman-Ford-Moore's algorithm \cite{Bellman58a,FordF62a,Moore59a}, guarantees that, should there be a negative circuit in the rooted distance graph of $P$, reachable
from the source vertex, it would be detected by the $nth$ pass at the latest. For $\stps$, therefore, the $\mbox{bdAC-1}$ procedure of  Figure \ref{bdAC1tcsp} is
equivalent to the  $\mbox{bdAC-1stp}$ procedure of Figure \ref{bdAC1stp}. From \mbox{bdAC-1stp}, we can get \mbox{bdAC-3stp}, another form of the
\mbox{bdAC-3} procedure for $\stps$ (see Figure \ref{bdAC3stp}): whenever a binarized domain $P_{0i}$ is updated, the edges $(k,i)$ to be added to the queue
are dealt with so that $(k,i)$ is propagated by the current pass if it is in the queue of the current pass (Q), and added to \mbox{Qnextpass} so that it will be propagated
by the next pass, if it is no more in the queue $Q$ of the current pass.

Therefore, we can, and do, bring the experimental comparison on $\stps$ of \mbox{bdAC-3} with the arc-consistency algorithm $\acstp$ in \cite{KongLL18a}, to that on $\stps$ of
\mbox{bdAC-3} with \mbox{bdAC-1}.
\subsection{Experimental comparison of \mbox{bdAC-3stp} with \mbox{bdAC-1stp}}
As motivated by the previous subsection, we transform  the experimental comparison on $\stps$ of \mbox{bdAC-3} with the arc-consistency algorithm $\acstp$ in \cite{KongLL18a}, to
that on $\stps$ of \mbox{bdAC-3} with \mbox{bdAC-1}. The comparison went through two phases, the results of the first having made it clear that the best way to compare the
two algortihms was to do it on trivially consistent $\stps$. The rest of the subsection describes the two phases.
\subsubsection{Phase 1}
Phase 1 of our experimental comparison went as follows:
\begin{enumerate}
  \item We considered two sizes (numbers of variables), 51 and 101.
  \item For each of the two sizes, we considered tight $\stps$ and coarse $\stps$: the constraints of a tight $\stp$ are of the form $(X_j-X_i)\in [a,b]$, with $-5\leq a\leq b\leq 5$,
           whereas the constraints of a coarse $\stp$ are of the form $(X_j-X_i)\in [a,b]$, with -(Nv div 2)$\leq$ a$\leq$ b$\leq$ Nv div 2, $Nv$ being the number of variables, and
           $Nv$ div 2 the integer division of Nv by 2. We looked at the interval [-5,5] as being tight, and at the interval [-(Nv div 2),Nv div 2], because the number Nv of variables
           is large, as being coarse, in the sense that the resulting constraints are permissive and give more freedom to the differences $(X_j-X_i)$ of variables, than do tight
           constraints. This led us to the following four cases:
    \begin{enumerate}
      \item Case 1: $\stps$ with 51 variables and tight constraints.
      \item Case 2: $\stps$ with 51 variables and coarse constraints.
      \item Case 3: $\stps$ with 101 variables and tight constraints.
      \item Case 4: $\stps$ with 101 variables and coarse constraints.
    \end{enumerate}
  \item For each of the four cases, we considered nine densities, 10 to 90 with a step of 10:
    \begin{enumerate}
      \item The density of an $\stp$ is the percentage of pairs $(X_i,X_j)$ of variables, with $0<i<j<Nv$, on which there is a constraint, $Nv$ being the number of variables.
      \item More formally, if the $\stp$ is $P=(X,C)$, with $|X|=Nv$ and $X=\{X_0,X_1,\ldots ,X_{Nv-1}\}$, the maximum number of pairs $(X_i,X_j)$ such that
              $1\leq i<j\leq Nv-1$, is clearly $(Nv-2)(Nv-1)/2$, and the density is given by $D=\frac{2\times Nc}{(Nv-2)\times (Nv-1)}\times 100=\frac{200\times Nc}{(Nv-2)\times (Nv-1)}$,
              where $Nc$ is the number of constraints of the form $(X_j-X_i)\in C_{ij}$, with $1\leq i<j\leq Nv-1$. So, from $D$ and $Nv$, we can get
              $Nc=\frac{D\times (Nv-2)\times (Nv-1)}{200}$.
      \item For $Nv=51$, $Nc=122$ if $D=10$ and $Nc=1102$ if $D=90$.
      \item For $Nv=101$, $Nc=495$ if $D=10$ and $Nc=4455$ if $D=90$.
    \end{enumerate}
  \item For each of the four cases, we randomly generated ten $\stps$ per density.
  \item All generated $\stps$ were inconsistent, and the performances of the two algorithms were exactly the same (the same number of calls of the procedure REVISE).
\end{enumerate}
\subsubsection{Phase 2}
As alluded to above, the results of the first phase clearly suggested that we had to jump to a second phase, and get the two algorithms fed with consistent $\stps$, making them go all
the way through, from the initial $\stp$ to the resulting bdArc Consistent $\stp$. The second phase, therefore, was conducted in the same way as the first, except that the number of
sizes was five instead of just two: five sizes from 51 to 251 with a step of 50, tight and coarse constraints, nine densities for each of the resulting ten case, and ten $\stps$ randomly
generated per case and per density. The random generation of the $\stps$ feeding the two algorithms also changed, which was made in a way that guaranteed that the $\stps$ were always consistent. We considered the class $\stps ^{0}$
of $\stps$ whose constraints are of the form $(X_j-X_i)\in [a,b]$, with $[a,b]$ in the set $S^{0}=\{[c,d]\mbox{ such that }-\infty <c\leq 0\leq d<+\infty\}$. The class $\stps ^{0}$ is
clearly closed under bdArc Consistency, because the set $S^{0}$ is closed under the three operations of converse, intersection and composition:
\begin{figure}[t]
\centerline{\includegraphics[scale=0.43]{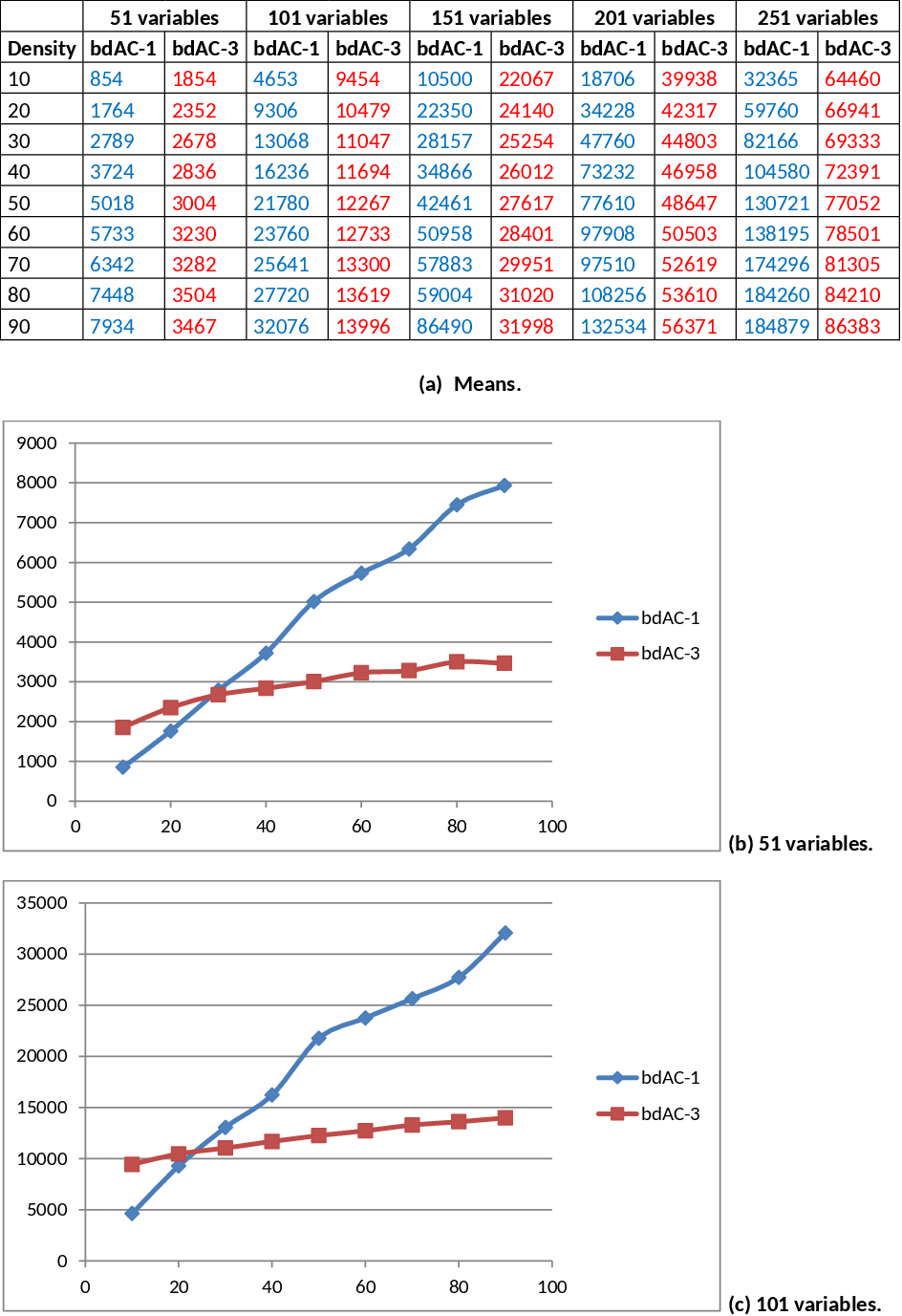}
\includegraphics[scale=0.43]{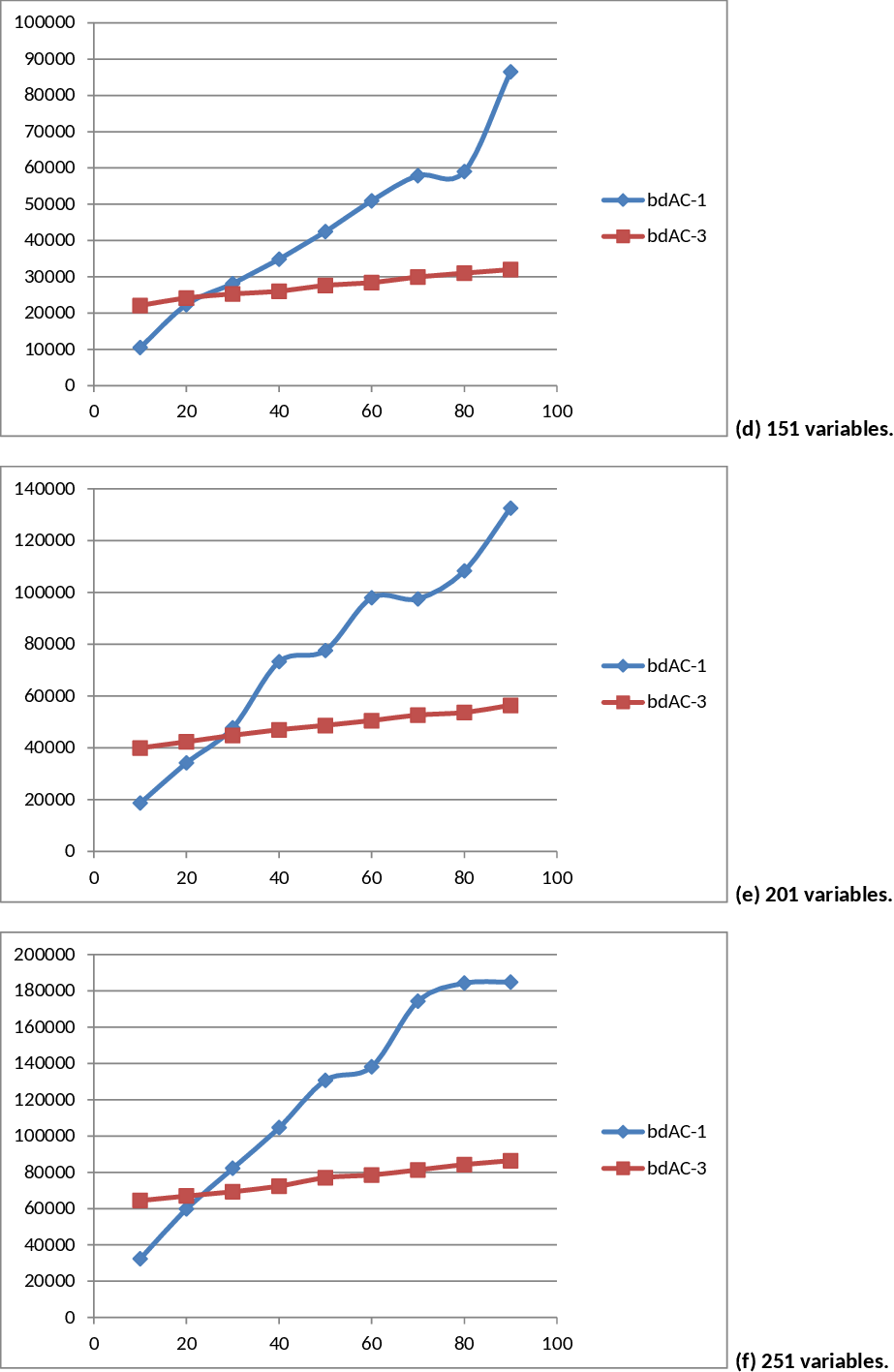}}
\caption{Results of phase 2 of our experimental comparison of \mbox{bdAC-3stp} with \mbox{bdAC-1stp} on trivially consistent {\bf tight} $\stps$ randomly generated: the
	number of calls of procedure $\mbox{REVISE}^{-}$ (y-axis) as a function of the density (x-axis).}\label{expcmp-t}
\end{figure}
\begin{figure}[t]
\centerline{\includegraphics[scale=0.43]{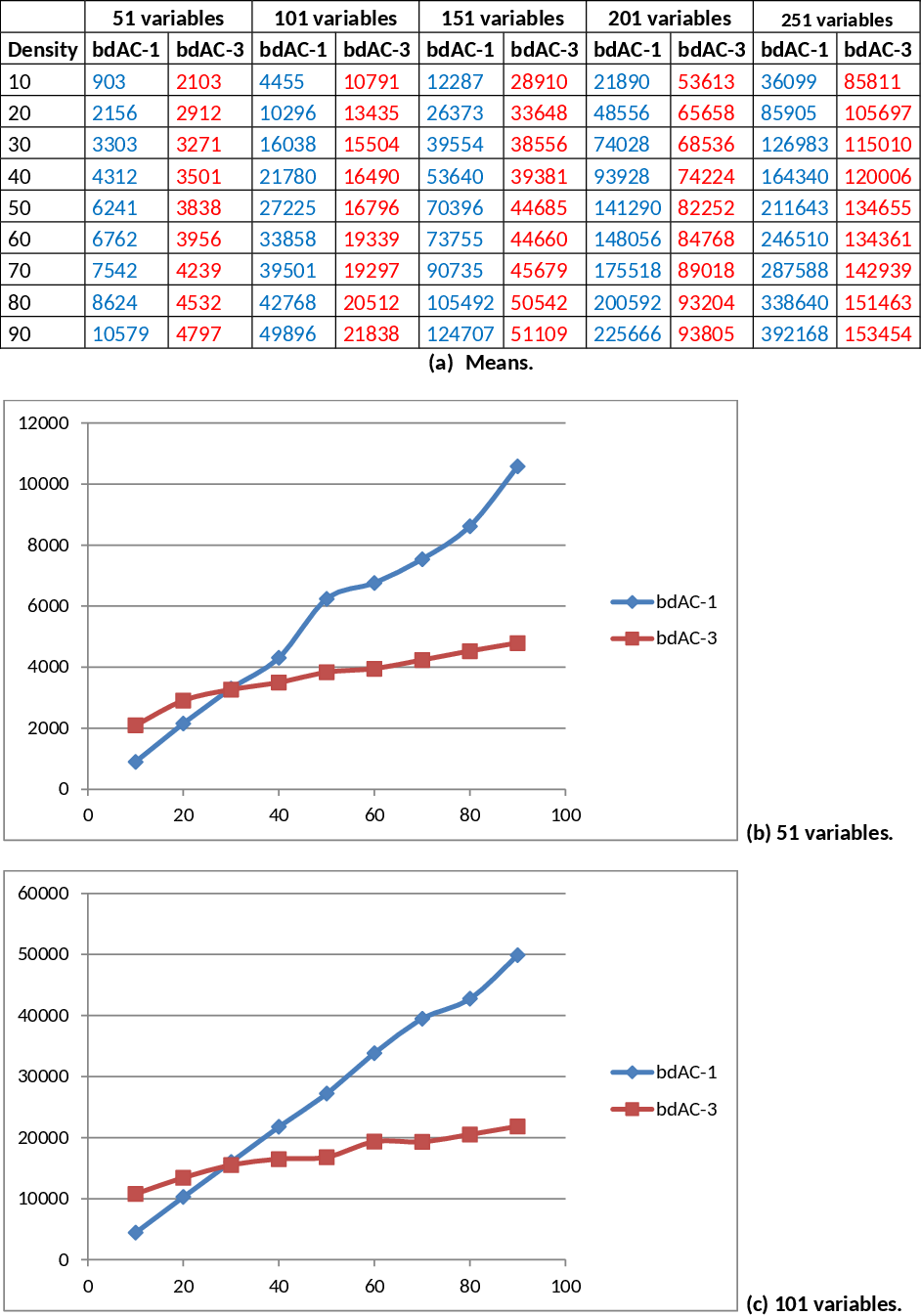}
\includegraphics[scale=0.43]{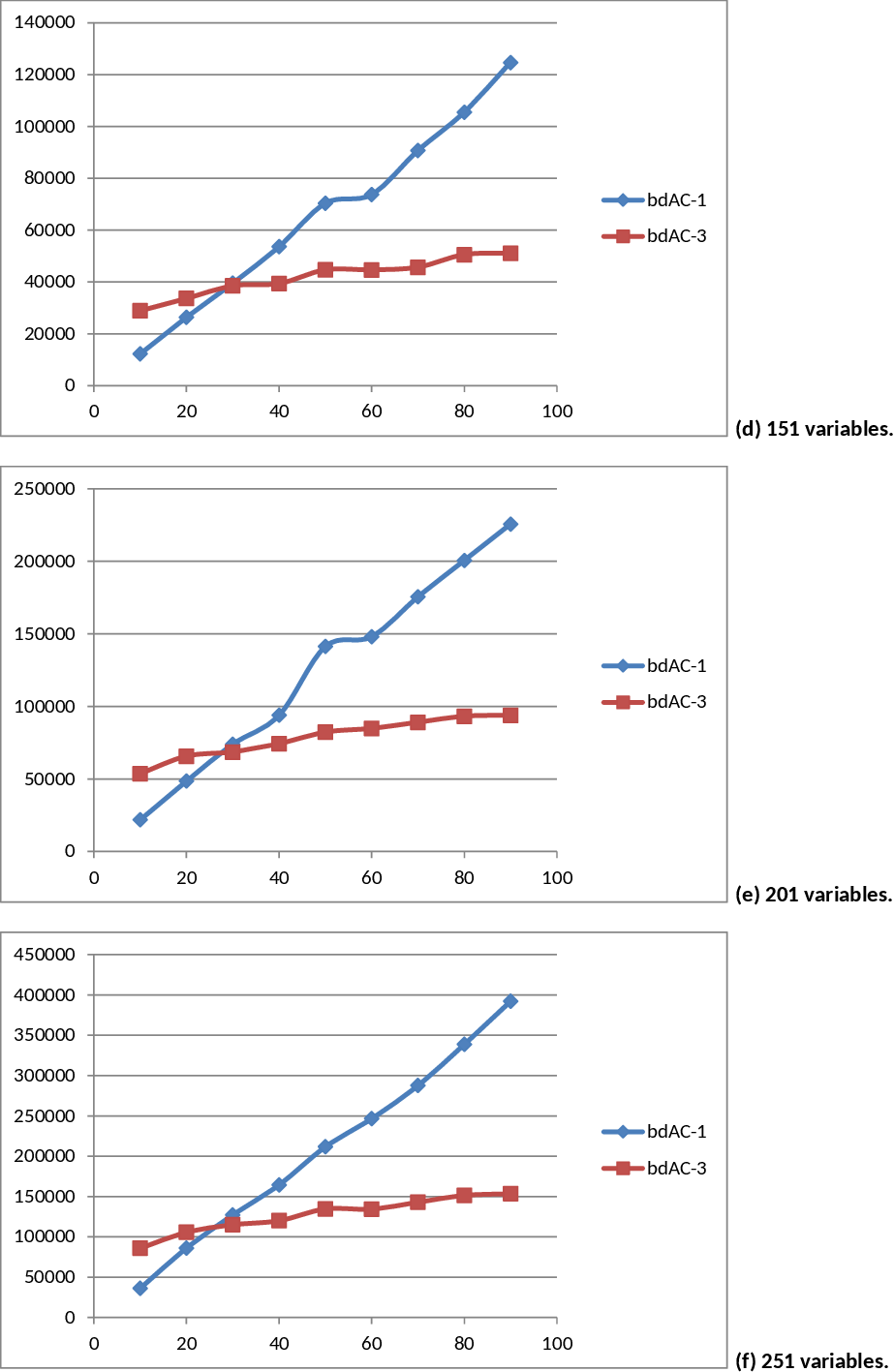}}
\caption{Results of phase 2 of our experimental comparison of \mbox{bdAC-3stp} with \mbox{bdAC-1stp} on trivially consistent {\bf coarse} $\stps$ randomly generated: the
	number of calls of procedure $\mbox{REVISE}^{-}$ (y-axis) as a function of the density (x-axis).}\label{expcmp-c}
\end{figure}
\begin{enumerate}
  \item For all constraints $(X_j-X_i)\in [a,b]$ such that $-\infty <a\leq 0\leq b<+\infty$, we have $-\infty <-b\leq 0\leq -a<+\infty$.
  \item For all constraints $(X_j-X_i)\in [a,b]$ and $(X_j-X_i)\in [c,d]$ such that $-\infty <a\leq 0\leq b<+\infty$ and $-\infty <c\leq 0\leq d<+\infty$, we have
          $-\infty <\mbox{max}(a,c)\leq 0\leq\mbox{min}(b,d)<+\infty$.
  \item For all constraints $(X_k-X_i)\in [a,b]$ and $(X_j-X_k)\in [c,d]$ such that $-\infty <a\leq 0\leq b<+\infty$ and $-\infty <c\leq 0\leq d<+\infty$, we have
          $-\infty <a+c\leq 0\leq b+d<+\infty$.
\end{enumerate}
Therefore, bdAC-1 and bdAC-3 applied to an STP of $\stps ^{0}$ will always produce a bdArc Consistent STP of the class $\stps ^{0}$. Furthermore, an $\stp$ of the class $\stps ^{0}$
is trivially consistent, for it has $(X_0,X_1,\ldots ,X_n)=(0,0,\ldots ,0)$ as a trivial solution. Therefore, the terminating condition will never come from a binarized domain made empty.
The terminating condition, indeed, will come from the queue(s) used and the number of passes:
\begin{enumerate}
  \item At most $Nv$ passes to make sure that the $\stp$ has been made bdArc Consistent.
  \item For \mbox{bdAC-1}, if a whole pass finishes with no change, the $\stp$ has been made bdArc Consistent.
  \item For \mbox{bdAC-3}, if a whole pass finishes with $\mbox{Qnextpass} = []$, the $\stp$ has been made bdArc Consistent, \mbox{Qnextpass} being the list of edges to be
           processed by the next pass.
\end{enumerate}
 Hence, we have found how to feed the two algorithms with trivially consistent $\stps$: by randomly generating $\stps$ from the class $\stps ^{0}$. For this purpose, the random generation of
constraints went as follows:
\begin{enumerate}
  \item For tight $\stps$, the constraints were of the form $(X_j-X_i)\in [a,b]$, with $-5\leq a\leq 0\leq b\leq 5$.
  \item For coarse $\stps$, the constraints were of the form $(X_j-X_i)\in [a,b]$, with -(Nv div 2)$\leq$ a$\leq 0\leq$ b$\leq$ Nv div 2.
\end{enumerate}
The results of phase 2 are summarized in Figures \ref{expcmp-t} for tight $\stps$, \ref{expcmp-c} for coarse $\stps$:
\begin{enumerate}
  \item For each of the two figures, part (a) (the table) summarises, for the corresponding case (tight/coarse),  the results of the five sizes of our experimental comparison, and provides for
	each of the five sizes and each of the nine densities, and for each of the two algorithms, the mean of  the number of calls of procedure $\mbox{REVISE}^{-}$ of the ten $\stps$
	randomly generated.
  \item For each of the two figures, the remaining parts, (b) to (f), provide a graphical illustration of the results, plotting, for each of the five sizes, and each of the two algorithms, the
	number of calls of procedure $\mbox{REVISE}^{-}$ (y-axis) as a function of the density (x-axis).
\end{enumerate}
The results neatly show, both for tight $\stps$ and coarse $\stps$, that $\mbox{bdAC-3stp}$ is the winner for all five sizes, and that, starting from density 30, it takes a clear lead over
$\mbox{bdAC-1stp}$.
\section{Experimental comparison of $\tcsp$-based job shop solvers}\label{expcmpjss}
First, this remark:
\begin{remark}\label{remark5}
\begin{enumerate}
  \item We refer to the procedure $\mbox{REVISE}$ of Figure \ref{TheREVISEProcedure} with weak composition in line \ref{revise3} instead of classical composition as $\mbox{w-REVISE}$
           ($temp=P_{0i}\cap P_{0j}\otimes _wP_{ji}$).
  \item We refer to the procedure $\mbox{REVISE}$ of Figure \ref{TheREVISEProcedure} with loose intersection in line \ref{revise3} instead of classical intersection as $\mbox{l-REVISE}$
          ($temp=P_{0i}\vartriangleleft P_{0j}\otimes P_{ji}$).
  \item We refer to the algorithm $\mbox{bdAC-3}$ (Figure \ref{bdAC3}) in which procedure $\mbox{REVISE}$ is replaced with $\mbox{w-REVISE}$ (respectively, $\mbox{l-REVISE}$) as
           $\mbox{wbdAC-3}$, for weak $\mbox{bdAC-3}$ (respectively, $\mbox{lbdAC-3}$, for loose $\mbox{bdAC-3}$).
  \item We refer to the procedure $\mbox{REVISE\_PC-2}$ of Figure \ref{revisePC2} with weak composition in line \ref{previse2} instead of classical composition as
	$\mbox{w-REVISE\_PC-2}$.
  \item We refer to the procedure $\mbox{REVISE\_PC-2}$ of Figure \ref{revisePC2} with loose intersection in line \ref{previse2} instead of classical intersection as
	$\mbox{l-REVISE\_PC-2}$.
  \item  We refer to the algorithm $\mbox{PC-2}$ (Figure \ref{PC2tcsp}) in which procedure $\mbox{REVISE\_PC-2}$ is replaced with
	$\mbox{w-REVISE\_PC-2}$ (respectively,  $\mbox{l-REVISE\_PC-2}$) as $\mbox{wPC-2}$, for weak $\mbox{PC-2}$ (respectively,  $\mbox{lPC-2}$, for loose $\mbox{PC-2}$).
\end{enumerate}
\end{remark}
\subsection{Path consistency and its unfragmenting weak versions}
The fragmentation problem \cite{SchwalbD97a} associated with the behavior of path consistency on a $\tcsp$  led to the use of weak versions of it as the filtering procedure in
solution search algorithms. The experimental comparison in \cite{SchwalbD97a} of two of these weak versions,  \mbox{ULT} and \mbox{LPC}, as filtering procedures in a $\tcsp$ solver
was much advantageous to \mbox{LPC}. \mbox{ULT} is very similar to $\mbox{wPC-2}$: \mbox{ULT} applied to a $\tcsp$ $P$ stands for first applying path consistency to the convex closure
of $P$, then intersecting the result with $P$; $\mbox{wPC-2}$ applied to $P$ stands for applying to $P$ path consistency in which composition is replaced with weak composition.
\subsection{Arc consistency versus path consistency in discrete CSPs}
$\tcsp$ solution search algorithms using a binarized-domains arc consistency algorithm, such as $\mbox{bdAC-3}$, as the filtering procedure, are expected to drastically
overcome general $\tcsp$ solvers \cite{DechterMP91a,SchwalbD97a} and $\tcsp$-based schedulers \cite{BelhadjiI98a} using path-consistency or a weak version of it as the
filtering procedure. A plausible explanation to this expected future for the newborn notion of \mbox{binarized-domains} arc-consistency, is that, in the case of classical binary
CSPs, arc-consistency algorithms and path-consistency algorithms, the first of which have a lower filtering power but are computationally cheaper, are both known since the
seventies \cite{Montanari74a,Mackworth77a}, and the place of arc-consistency is known to be more advantageous than that of path-consistency as the filtering procedure of
solution search algorithms. A visible witness of this advantageous place of arc-consistency, is that, research on improving efficiency of arc-consistency has attracted tremendous
efforts since its very first algorithms, \mbox{AC-1} and \mbox{AC-3} \cite{Mackworth77a}, that led to various improved versions of the original algorithms.
\subsection{The experiments}
With the above considerations in mind, our experimental comparison implemented three $\tcsp$-based job shop schedulers:
\begin{enumerate}
  \item $\jswbdac3$: this is an implementation of the job-shop scheduler described in Section \ref{scheduler} (Figure \ref{jsscheduler}), that uses $\mbox{wbdAC-3}$ as the filtering procedure
	during the search.
  \item $\jslbdac3$: this is an implementation of the job-shop scheduler described in Section \ref{scheduler} (Figure \ref{jsscheduler}), in which $\mbox{lbdAC-3}$ is used as the filtering
	procedure during the search, instead of $\mbox{wbdAC-3}$.
  \item $\jslpc2$: this is an implementation of the job-shop scheduler described in Section \ref{scheduler} (Figure \ref{jsscheduler}), in which $\mbox{lpc-2}$ is used as the filtering
	procedure during the search, instead of $\mbox{wbdAC-3}$.
\end{enumerate}
Each of the three solvers has been fed with the well-known job shop instance \mbox{ft06} \cite{FisherT63a}:

\begin{scriptsize}
$
\begin{array}{|l|}\hline
\mbox{Instance ft06=}\\
\mbox{[[ 6, 6],}\\
\mbox{ [2, 1, 0, 3, 1, 6, 3, 7, 5, 3, 4, 6],}\\
\mbox{ [1, 8, 2, 5, 4, 10, 5,10, 0, 10, 3, 4],}\\
\mbox{ [2, 5, 3, 4, 5, 8, 0, 9, 1, 1, 4, 7],}\\
\mbox{ [1, 5, 0, 5, 2, 5,3, 3, 4, 8, 5, 9],}\\
\mbox{ [2, 9, 1, 3, 4, 5, 5, 4, 0, 3, 3, 1],}\\
\mbox{ [1, 3, 3, 3, 5,9, 0, 10, 4, 4, 2, 1]]}\\  \hline
\end{array}
$
\end{scriptsize}\\
The first sublist gives the number of jobs and the number of machines, both equal to 6. Each of the remaining six 12-number sublists describes one of the six jobs: the first two numbers describe the first task of the job
(a number from 0 to 5 giving the machine on which the task is to be executed, and another number giving the duration of the task), ..., the last two numbers describe the last task of the job.
The optimum is initialized to the sum of all durations, $197$. The solver then goes through a number of optimum updates until the final, true value is reached. When an update occurs, the solver outputs the corresponding
new value of the triple (Opt,Time,Sol), Opt being the new value of the optimum, Time
the run time in the form [hours, minutes, seconds], and Sol a solution realizing the new temporary value of the optimum, in the form $(X_0,X_1,\ldots ,X_n)$, with $X_0=0$. However, we provide all three parameters
for three updates only, including the very first and the very last; for the remaining updates, we provide only the first two parameters, Opt and Time. The results of the experiments are as follows:
\subsubsection{$\jswbdac3$}
\begin{enumerate}
  \item Initialization of the optimum to the sum of durations: Opt = 197
  \item Very first optimum update:
    \begin{enumerate}
      \item Opt = 170, Time: [hours, minutes, seconds] = [0,0,0]
      \item $Sol = (X_0,X_1,X_2,\cdots ,X_{36})
	       = (0,115,145,148,154,161,164,102,$
      \item[] $110,115,125,135,145,64,69,84,92,101,102,42,54,59,64,67,75,$
      \item[] $30,39,42,47,51,54,0,3,6,15,25,29)$
    \end{enumerate}
  \item Another optimum update:
    \begin{enumerate}
      \item Opt = 113, Time: [hours, minutes, seconds] = [0,0,7]
      \item $Sol = (X_0,X_1,X_2,\cdots ,X_{36})
	      = (0,58,88,91,97,104,107,45,53,58,$
      \item[] $68,78,88,18,23,27,35,44,45,3,8,13,18,29,37,0,9,12,17,25,28,$
      \item[] $0,3,6,15,25,29)$
    \end{enumerate}
  \item Very last optimum update:
    \begin{enumerate}
      \item Opt = 55, Time: [hours, minutes, seconds] = [1,19,50]
      \item $Sol = (X_0,X_1,X_2,\cdots ,X_{36})
	       = (0,22,27,30,36,43,49,0,8,13,28,$
      \item[] $40,50,0,5,9,18,27,30,8,13,23,28,37,46,13,22,25,38,50,54,13,$
      \item[] $16,19,30,45,49)$
    \end{enumerate}
  \item Run time: [hours, minutes, seconds] = [2,52,29] (almost three hours)
  \item All optimum updates in the form [Opt,[hours, minutes, seconds]], without the corresponding solutions: \\
\begin{scriptsize}
$
\begin{array}{|l|l|l|l|l|}\hline
 [170,[0,0,0]]&[157, [0,0,0]]&[144, [0,0,1]]&[141,[0,0,2]]&[138,[0,0,3]]\\  \hline
[135,[0,0,4]]&[132,[0,0,4]]&[129,[0,0,4]]&[125,[0,0,5]]&[124,[0,0,6]]\\  \hline
[121,[0,0,6]]&[118,[0,0,7]]&[115,[0,0,7]]&[113,[0,0,7]]&[111,[0,0,8]]\\  \hline
[110,[0,0,9]]&[108,[0,0,14]]&[107,[0,0,14]]&[104,[0,0,15]]&[103,[0,0,16]]\\  \hline
[102,[0,0,17]]&[101,[0,0,19]]&[96,[0,0,43]]&[95,[0,0,50]]&[94,[0,1,2]]\\  \hline
[92,[0,1,8]]&[90,[0,2,48]]&[89,[0,2,58]]&[88,[0,3,56]]&[86,[0,4,50]]\\  \hline
 [85,[0,4,50]]&[84,[0,4,54]]&[81,[0,4,56]]&[80,[0,4,57]]&[79,[0,4,59]]\\  \hline
[78,[0,7,56]]&[77,[0,8,22]]&[76,[0,8,24]]&[75,[0,15,49]]&[74,[0,15,51]]\\  \hline
[73,[0,15,53]]&[72,[0,19,53]]&[69,[0,20,41]]&[68,[0,20,49]]&[67,[0,21,2]]\\  \hline
[66,[0,21,7]]&[65,[0,21,15]]&[64,[0,26,28]]&[63,[0,26,46]]&[62,[0,39,0]]\\  \hline
[61,[0,44,37]]&[59,[0,45,15]]&[58,[1,1,28]]&[56,[1,4,44]]&[55,[1,19,50]]\\  \hline
\end{array}
$
\end{scriptsize}
\end{enumerate}
\subsubsection{$\jslbdac3$}
\begin{enumerate}
  \item Initialization of the optimum to the sum of durations: Opt = 197
  \item Very first optimum update:
    \begin{enumerate}
      \item Opt = 170, Time: [hours, minutes, seconds] = [0,0,0]
      \item $Sol = (X_0,X_1,X_2,\cdots ,X_{36})
                        = (0,115,145,148,154,161,164,102,$
      \item[] $110,115,125,135,145,64,69,84,92,101,102,42,54,59,64,67,75,$
      \item[] $30,39,42,47,51,54,0,3,6,15,25,29)$
    \end{enumerate}
  \item Another optimum update:
   \begin{enumerate}
      \item Opt = 113, Time: [hours, minutes, seconds] = [0,0,2] 
      \item $Sol = (X_0,X_1,X_2,\cdots ,X_{36})
                        = (0,58,88,91,97,104,107,45,53,58,$
      \item[] $68,78,88,18,23,27,35,44,45,3,8,13,18,29,37,0,9,12,17,25,28,$
      \item[] $0,3,6,15,25,29)$
    \end{enumerate}
  \item Very last optimum update:
    \begin{enumerate}
      \item Opt = 55, Time: [hours, minutes, seconds] = [0,26,21]
      \item $Sol = (X_0,X_1,X_2,\cdots ,X_{36})
                        = (0,22,27,30,36,43,49,0,8,13,28,$
      \item[] $40,50,0,5,9,18,27,30,8,13,23,28,37,46,13,22,25,38,50,54,13,$
      \item[] $16,19,30,45,49)$
    \end{enumerate}
  \item Run time: [hours, minutes, seconds] = [0,57,32] (less than an hour)
  \item All optimum updates in the form [Opt,[hours, minutes, seconds]], without the corresponding solutions:\\
    \begin{scriptsize}
    $
    \begin{array}{|l|l|l|l|l|}\hline
          [170,[0,0,0]]&[157,[0,0,0]]&[144,[0,0,0]]&[141,[0,0,1]]&[138,[0,0,1]]\\  \hline
      [135,[0,0,1]]&[132,[0,0,1]]&[129,[0,0,1]]&[125,[0,0,2]]&[124,[0,0,2]]\\  \hline
      [121,[0,0,2]]&[118,[0,0,2]]&[115,[0,0,2]]&[113,[0,0,2]]&[111,[0,0,3]]\\  \hline
      [110,[0,0,3]]&[108,[0,0,5]]&[107,[0,0,5]]&[104,[0,0,5]]&[103,[0,0,5]]\\  \hline
      [102,[0,0,5]]&[101,[0,0,6]]&[96,[0,0,15]]&[95,[0,0,17]]&[94,[0,0,21]]\\  \hline
      [92,[0,0,23]]&[90,[0,0,54]]&[89,[0,0,57]]&[88,[0,1,16]]&[86,[0,1,32]]\\  \hline
      [85,[0,1,33]]&[84,[0,1,34]]&[81,[0,1,35]]&[80,[0,1,35]]&[79,[0,1,36]]\\  \hline
      [78,[0,2,33]]&[77,[0,2,41]]&[76,[0,2,42]]&[75,[0,5,12]]&[74,[0,5,12]]\\  \hline
      [73,[0,5,13]]&[72,[0,6,33]]&[69,[0,6,50]]&[68,[0,6,53]]&[67,[0,6,57]]\\  \hline
      [66,[0,6,59]]&[65,[0,7,1]]&[64,[0,8,47]]&[63,[0,8,53]]&[62,[0,12,55]]\\  \hline
      [61,[0,14,47]]&[59,[0,14,59]]&[58,[0,20,20]]&[56,[0,21,22]]&[55,[0,26,21]]\\  \hline
    \end{array}
$
\end{scriptsize}
\end{enumerate}
\subsubsection{$\jslpc2$}
\begin{enumerate}
\item Initialization of the optimum to the sum of durations: Opt = 197
\item Very first optimum update:
  \begin{enumerate}
    \item Opt = 170, Time: [hours, minutes, seconds] = [0,4,17]
    \item $Sol = (X_0,X_1,X_2,\cdots ,X_{36})
	       = (0,115,145,148,154,161,164,102,$
    \item[] $110,115,125,135,145,64,69,84,92,101,102,42,54,59,64,67,75,$
    \item[] $30,39,42,47,51,54,0,3,6,15,25,29)$
  \end{enumerate}
\item Another optimum update:
  \begin{enumerate}
    \item Opt = 113, Time: [hours, minutes, seconds] = [0,26,49]
    \item $Sol =  (X_0,X_1,X_2,\cdots ,X_{36})
	      = (0,58,88,91,97,104,107,45,53,$
    \item[] $58,68,78,88,18,23,27,35,44,45,3,8,13,18,29,37,0,9,12,17,25,$
    \item[] $28,0,3,6,15,25,29)$
  \end{enumerate}
\item Very last optimum update:
  \begin{enumerate}
    \item Opt = 55, Time: [hours, minutes, seconds] = [134,7,0]
    \item $Sol =  (X_0,X_1,X_2,\cdots ,X_{36})
	       = (0,22,27,30,36,43,49,0,8,13,28,$
      \item[] $40,50,0,5,9,18,27,30,8,13,23,28,37,46,13,22,25,38,50,54,13,$
      \item[] $16,19,30,45,49)$
  \end{enumerate}
\item Run time: [hours, minutes, seconds] = [335,51,1] (14 days - 8 minutes - 59 seconds)
\item All optimum updates in the form [Opt,[hours, minutes, seconds]], without the corresponding solutions:\\
\begin{scriptsize}
  $
  \begin{array}{|l|l|l|l|l|}\hline
    [170,[0,4,17]]&[157,[0,4,31]]&[144,[0,4,41]&[141,[0,12,28]]&[138,[0,14,37]]\\  \hline
    [135,[0,17,4]]&[132,[0,18,24]]&[129,[0,18,36]]&[125,[0,21,0]]&[124,[0,22,33]]\\  \hline
    [121,[0,23,37]]&[118,[0,25,1]]&[115,[0,25,28]]&[113,[0,26,49]]&[111,[0,27,42]]\\  \hline
    [110,[0,31,8]]&[108,[0,42,8]]&[107,[0,42,29]]&[104,[0,44,6]]&[103,[0,46,42]]\\  \hline
    [102,[0,48,17]]&[101,[0,52,23]]&[96,[2,32,6]]&[95,[2,51,23]]&[94,[3,29,43]]\\  \hline
    [92,[3,40,26]]&[90,[7,11,21]]&[89,[7,37,12]]&[88,[9,21,21]]&[86,[10,55,3]]\\  \hline
    [85,[10,57,55]]&[84,[11,7,18]]&[81,[11,10,39]]&[80,[11,12,20]]&[79,[11,18,32]]\\  \hline
    [78,[17,6,26]]&[77,[17,39,2]]&[76,[17,42,53]]&[75,[35,35,56]]&[74,[35,39,5]]\\  \hline
    [73,[35,42,30]]&[72,[45,18,3]]&[69,[47,40,1]]&[68,[48,0,10]]&[67,[48,18,51]]\\  \hline
    [66,[48,29,24]]&[65,[48,42,44]]&[64,[57,9,0]]&[63,[57,35,34]]&[62,[76,5,14]]\\  \hline
    [61,[85,43,32]]&[59,[86,16,17]]&[58,[106,56,56]]&[56,[110,27,37]]&[55,[134,7,0]]\\  \hline
  \end{array}
  $
\end{scriptsize}
\end{enumerate}
\subsubsection{Discussion}
The following table summarizes the results, and provides for each solver $s$ the time $t_1(s)$ taken to reach the
final value 55 of the optimum of ft06, and the time $t_2(s)$ corresponding to the whole duration (run time, which includes the
time needed to reach the final value of the optimum and the additional time needed from then to realize that it's the final value):\\
\begin{scriptsize}
$
\begin{array}{|l||r|r|r|r|r|r|}\hline
solver\mbox{ }s     &t_1(s)&t_2(s)&\frac{t_1(\jswbdac3)}{t_1(\jslbdac3)}&\frac{t_2(\jswbdac3)}{t_2(\jslbdac3)}     &\frac{t_1(\jslpc2)}{t_1(\jslbdac3)}&\frac{t_2(\jslpc2)}{t_2(\jslbdac3)}\\  \hline\hline
\jswbdac3&4790&10349&&&&\\  \cline{1-3}
\jslbdac3&1581&3452&3,03&3,00&305,39&350,25\\  \cline{1-3}
\jslpc2&482820&1209061&&&&\\  \hline
\end{array}
$
\end{scriptsize}
The comparison clearly shows that the main point in \cite{SchwalbD97a} on the use of weak versions of path consistency as filtering procedures in $\tcsp$ solvers also holds for
the use of weak versions of \mbox{bdac-3} as filtering procedures in $\tcsp$ solvers: \mbox{lbdAC-3} (loose intersection instead of classical intersection) is more effective than
\mbox{wbdAC-3} (weak composition instead of classical composition). Indeed, as shown in the summarizing table, the time needed for $\jswbdac3$ to reach the final value of the
optimum is three times that of $\jslbdac3$; and the same applies for the whole run times of the two solvers.

The rest of the summarizing table shows that, as a filtering procedure in $\tcsp$ solvers, the best weak version of \mbox{bdAC-3}, namely \mbox{lbdAC-3}, outperforms the best
weak version of path consistency, namely \mbox{lPC-2}. Indeed, as the table shows, the time needed for $\jslpc2$ to reach the final value of the optimum is more than 305 times
that of $\jslbdac3$; and the same applies for the whole run times of the two solvers. This result confirms our expectation that the drastically advantageous performance of arc
consistency as a filtering procedure in discrete CSPs' solvers, compared to path consistency, would apply to (weak versions of) \mbox{bdAC-3} compared to (weak versions of) path
consistency as a filtering procedure in $\tcsp$ solvers.
\section{An incremental version of \mbox{bdAC-3}}\label{incv}
\begin{figure}
\begin{enumerate}
  \item[] {\bf Input: }The matrix representation of an n+1-variable bdArc-consistent $\tcsp$ $P=(X,C_1)$; and a set $C_2$ of new constraints of the form $(X_j-X_i)\in B_{ij}$, with $0\leq i<j\leq n$.
  \item[] {\bf Output: }False, indicating that inconsistency of $P$ augmented with $C_2$ has been detected; or true, indicating that $P$ augmented with $C_2$  has been made bdArc-consistent again.
  \item[] {\bf procedure }$\mbox{ibdAC-3}(P,C_1,C_2)$\{
  \item\label{ibdac31}\hskip 0.2cm $Q=\{(k,i): (k*i\not =0)\mbox{ and }(k\not =i)$
  \item[]\hskip 0.5cm $\mbox{ and }\mbox{ }(C_2\mbox{ has a binarized-domain constraint on }X_i)$
  \item[]\hskip 0.5cm $\mbox{ }and\mbox{ }(C_1\cup C_2\mbox{ has a non binarized-domain constraint on $X_k$ and $X_i$)}\}$;
  \item[]\hskip 0.2cm $Q=Q\cup\{(i,j),(j,i):\mbox{ }(i*j\not =0)\mbox{ and }(i\not =j)$
  \item[]\hskip 0.5cm $\mbox{ and }(C_2\mbox{ has a non binarized-domain constraint on $X_i$ and $X_j$)}\}$;
  \item\hskip 0.2cm {\bf for all }$(i,j)$ with ($0\leq i<j\leq n$)
  \item[]\hskip 0.5cm \mbox{ and }($C_2$ has a constraint $(X_j-X_i)\in B_{ij}$)\{
  \item[]\hskip 1.2cm $P_{ij}=P_{ij}\cap B_{ij}$; {\bf if}($P_{ij}=\emptyset$){\bf return false}; $P_{ji}=P_{ij}^\smile$\}
  \item\hskip 0.2cm $Empty\_domain=false$;
  \item\label{ibdac32}\hskip 0.2cm {\bf while }$(Q\not =\emptyset$\mbox{ and (not }Empty\_domain))\{
  \item\label{ibdac33}\hskip 0.8cm select and delete an arc $(k,m)$ from $Q$;
  \item\label{ibdac34}\hskip 0.8cm {\bf if }$\mbox{REVISE}(k,m)$
  \item\label{ibdac35}\hskip 1.2cm if $(P_{0k}=\emptyset)$ $Empty\_domain=true$
  \item\label{ibdac36}\hskip 1.2cm else $Q=Q\cup\{(i,k):\mbox{ $C_1\cup C_2$ has a constraint}$
   \item[]\hskip 2.0cm \mbox{on $X_i$ and $X_k$, $i\not =0$, $i\not =k$, $i\not =m$}\}\}\%endwhile
  \item\label{ibdac37}\hskip 0.4cm return(not $Empty\_domain$)\}\%end ibdAC3
\end{enumerate}
\caption{An incremental version, \mbox{ibdAC-3}, of \mbox{bdAC-3}. We assume that the set $C_2$ includes binarized-domain constraints and non binarized-domain constraints.}\label{ibdAC3}
\end{figure}
Figure \ref{ibdAC3} provides an incremental version of \mbox{bdAC-3}.

Incrementality is crucial for many applications of constraint-based reasoning in general, of constraint-based temporal reasoning in particular. It faces the challenging problem of processing
constraints in real-world situations where these do not come into the system all at once, in advance, and arrive, instead, per packs. When a new pack of constraints enters the system,
incrementality allows to treat it without having to redo the work that had been done on the packs arrived before it.

As seen in this work, $\stps$ were introduced in \cite{DechterMP91a}, where the more general framework of $\tcsps$ was introduced. $\tcsps$ are more general than $\stps$ in the sense that
constraints are not exclusively convex; they can, indeed,  be disjunctive while remaining binary. Another interesting generalization of $\stps$ is the framework of Disjunctive Temporal Networks
\cite{StergiouK00a}, or $\dtns$, where a constraint may be of any arity $k\leq n$, $n$ being the number of variables, and is of the form $(Y_1-X_1)\in A_1\vee ...\vee (Y_i-X_i)\in A_i$ where
$X_j$ and $Y_j$, $j$ from 1 to $i$, are variables and $A_1, ...,A_i$ convex subsets of $\BBR$. With the aim of representing real-world situations where the constraints cannot be known all at
once, $\stps$, in \cite{VidalF99a,MorrisMV01a}, then $\dtns$, in \cite{VenableY05a}, were generalized to $\stps$ with uncertainty, or $\stnus$, and $\dtns$ with uncertainty, or $\dtnus$. $\stnus$
and $\dtnus$ strengthened the importance of incrementality in constraint satisfaction in general, and in temporal constraint satisfaction in particular.

Incrementality is also used by constraint solvers using and applying, at each node of the search space, a filtering procedure such as \mbox{bdAC-3} or \mbox{PC-2}. Our
implemented job shop schedulers use \mbox{bdAC-3} or \mbox{PC-2}, or weak versions of them, as the filtering procedure during the search, and do indeed use incrementality
at each node, when initializing the queue of the filtering procedure. For instance, the solver with \mbox{bdAC-3} as the filtering procedure does it as follows:
\begin{enumerate}
  \item the very first filtering, at the root of the search space, known as the preprocessing step, initializes the \mbox{bdAC-3} queue Q to all pairs $(X_i,X_j)$, with $i*j\not =0$,
          on which there is a constraint;
  \item at every other node $N_2$, whose parent node is $N_1$, the filtering by \mbox{bdAC-3} is done on the bdArc consistent $\tcsp$ $P_1$ labeling $N_1$ in which an entry $(X_i,X_j)$,
           which was disjunctive in $P_1$, has been instantiated with one of its disjuncts. This means that the queue of the filtering at $N_2$ can be, and is, initialized to $\{(X_i,X_j),(X_j,X_i)\}$.
\end{enumerate}
\section{Summary and future work}\label{sfive}
We provided, and analysed the computational complexity of, an adaptation to $\tcsps$ \cite{DechterMP91a} of Mackworth's \cite{Mackworth77a} arc-consistency
algorithm $\mbox{AC-3}$. A $\tcsp$ is closed under classical node- and arc-consistencies \cite{Montanari74a,Mackworth77a}, and we argued that this was certainly what
made \cite{DechterMP91a} jump directly to the next local consistency algorithm, namely path-consistency. What made the adaptation possible was the look at the constraints
of a $\tcsp$ between the ''origin of the world'' variable and the other variables, as the binarized domains of these other variables.

For the convex part of $\tcsps$, namely $\stps$, we showed the equivalence between applying $\mbox{bdAC-3}$ to an $\stp$, on one hand, and applying a one-to-all
all-to-one shortest paths algorithm to its distance graph, on the other; this led to a generalization of Bellman-Ford-Moore's algorithm \cite{Bellman58a,FordF62a,Moore59a}.
We showed that the binarized domains of a connected bdArc-Consistent $\stp$ are minimal, and
provided a polynomial backtrack-free procedure computing a solution to such an $\stp$. Another polynomial backtrack-free procedure has been provided for the task of
extracting, from a bdArc-Consistent $\stp$, a connected bdArc-Consistent $\stp$ refinement, if the $\stp$ is consistent, or showing its inconsistency, otherwise.

We showed how to use $\mbox{wbdAC-3}$ as the filtering procedure of a general $\tcsp$ solver, and of a $\tcsp$-based job shop scheduler. We also showed that
bdArc Consistency is complete for a known class of $\tcsps$ named \mbox{STAR} $\tcsps$ \cite{SchwalbD97a}; as a consequence, \mbox{bdAC-3} can be used as the
filtering procedure of a $\tcsp$-based job shop scheduler, instead of \mbox{wbdAC-3}, with no risk of making its branching factor worsened by the fragmentation problem
\cite{SchwalbD97a}.

We discussed how to adapt to $\tcsps$ the other local-consistency algorithms in \cite{Mackworth77a}, namely, the arc-consistency algorithm \mbox{AC-1}, and
the path consistency algorithms \mbox{PC-1} and \mbox{PC-2}. In particular, this led us to the conclusion that an existing adaptation to $\tcsps$ of \mbox{PC-2}
\cite{DechterMP91a,SchwalbD97a}, contrary to the adaptation we provided, is not guaranteed to always terminate. Furthermore, for $\stps$, inspired by the idea in
Bellman-Ford-Moore's algorithm \cite{Bellman58a,FordF62a,Moore59a}, we also provided  \mbox{bdAC-1stp} and \mbox{bdAC-3stp}, simpler versions of the
algorithms \mbox{bdAC-1} and \mbox{bdAC-3}, respectively.

We also provided:
\begin{enumerate}
  \item an experimental comparison of \mbox{bdAC-3} with an existing arc-consistency
                algorithm, $\acstp$, restricted to $\stps$ \cite{KongLL18a};
  \item an experimental comparison of three $\tcsp$-based job shop schedulers, two of which use weak versions of \mbox{bdAC-3} as the
                 filtering procedure during the search, the other \cite{SchwalbD97a} a weak version of
                 path-consistency; and
  \item an incremental version of \mbox{bdAC-3}.
\end{enumerate}
To the best of our knowledge, the binarized-domains arc-consistency algorithm $\mbox{bdAC-3}$ we have investigated for general $\tcsps$ is the very first of its kind, as
the one in \cite{KongLL18a}, $\acstp$, was restricted to $\stps$. We hope and expect that it will have a positive impact on the near-future developments of the field. Points needing further investigation
include improvement of the lower and upper bounds $\mbox{path-lb}(P)$ and $\mbox{path-ub}(P)$, as defined in this work for a $\tcsp$ $P$, as they are crucial for the \mbox{bdAC-3}
algorithm's computational behavior, and for that of search algorithms using \mbox{bdAC-3} as the filtering procedure during the search.

Finally, as discussed in the introduction, the $\tcsp$ tools provided in this work, that, again, include the bdArc Consistency algorithm \mbox{bdAC-3}, a $\tcsp$ solver using \mbox{bdAC-3} as
the filtering procedure, and a job shop scheduler using also \mbox{bdAC-3} as the filtering procedure, together with available spatial systems and tool boxes implementing existing qualitative spatial calculi \cite{EgenhoferF91a,Frank92a,Freksa92a,IsliC00a,Ligozat98a,RandellCuiCohn92a}, such as the toolbox \mbox{SparQ} in \cite{DyllaFWW06a,WallgrunFWDF06a}
or the \mbox{RCC-8} query answering system in \cite{BennettIC98a}, pave the way for an extension of pure Logic Programming to Constraint Logic Programming over Spatial and
Temporal Domains.
%
\bibliographystyle{acm} 
\bibliography{biblio-c-maj}
\appendix
\section{Complexity of the three algebraic operations}\label{appendix1}
The worst-case computational complexity of the three operations of converse, intersection and composition is as follows:
\begin{enumerate}
  \item If $C_{ik}$ and $C_{kj}$ are convex sets (intervals) then their composition $C_{ikj}=C_{ik}\otimes C_{kj}$ is convex:
           \begin{enumerate}
             \item the lower (respectively, upper) bound of $C_{ikj}$ is the sum of the lower (respectively, upper) bounds of $C_{ik}$ and
           $C_{kj}$;
             \item $C_{ikj}$ is left-closed (respectively, right-closed) if both $C_{ik}$ and $C_{kj}$ are left-closed (respectively,
            right-closed), it is left-open (respectively, right-open) otherwise.
            \end{enumerate}
            As an example,
            $(-41,20]\otimes [55,60]=(-41+55,20+60]=(14,80]$. We need therefore two operations of addition to perfom the
            composition of two convex sets, whose worst-case computational complexity is thus $O(1)$.
  \item If $C_{ij}^1$ and $C_{ij}^2$ are convex sets then their intersection $A=C_{ij}^1\cap C_{ij}^2$ is obtained by comparing
           the bounds of $C_{ij}^1$ with those of $C_{ij}^2$: its worst-case computational complexity is thus $O(1)$.
  \item If $C_{ij}$ is a convex set then its converse $C_{ij}^\smile$ is obtained by negating the bounds of $C_{ij}$: its computational
           complexity is thus also $O(1)$.
  \item Therefore:
           \begin{enumerate}
             \item If $C_{ik}$ and $C_{kj}$ are unions of convex sets with $|\mbox{mPcs}(C_{ik})|=n_1$ and $|\mbox{mPcs}(C_{kj})|=n_2$,
                      the worst-case computational complexity of their composition is $O(n_1n_2)$;
             \item If $C_{ij}^1$ and $C_{ij}^2$ are unions of convex sets with $|\mbox{mPcs}(C_{ij}^1)|=n_1$ and $|\mbox{mPcs}(C_{ij}^2)|=n_2$,
                      the worst-case computational complexity of their intersection is $O(n_1+n_2)$;
             \item If $C_{ij}$ is a union of convex sets with $|\mbox{mPcs}(C_{ij})|=n$, the worst-case computational complexity of its converse is $O(n)$. \cqfd
            \end{enumerate}
\end{enumerate}
\section{\mbox{bdAC-3} applied to an $\stp$: illustration 1}\label{appendix6}
\begin{figure}
\centerline{\includegraphics[width=8.3cm]{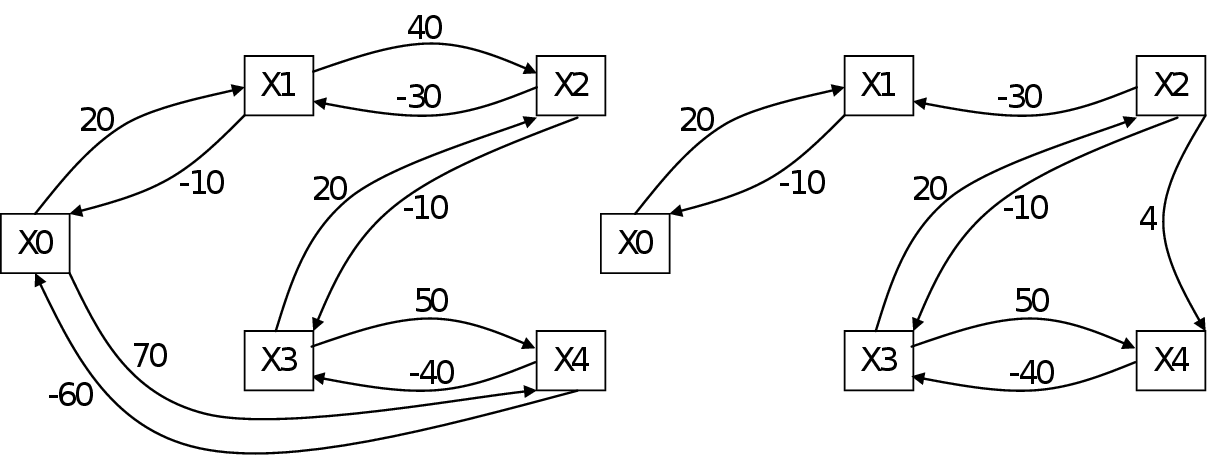}}
\caption{The distance graphs of the $\stps$ of \ref{appendix6} (left) and \ref{appendix7} (Right). The edges $(X_i,X_j)$ whose
               weights are infinite and the loops $(X_i,X_i)$ are not shown.}\label{stpdistancegraph}
\end{figure}

Let $P=(X,C)$ be the following $\stp$, whose distance graph is given by Figure \ref{stpdistancegraph}(Left):
\begin{enumerate}
  \item $X=\{X_0,X_1,X_2,X_3,X_4\}$ and
  \item $C=\{c_1:(X_1-X_0)\in [10,20],c_2:(X_4-X_0)\in [60,70],c_3:(X_2-X_1)\in [30,40],c_4:(X_3-X_2)\in [-20,-10],c_5:(X_4-X_3)\in [40,50]\}$.
\end{enumerate}
The $\stp$ is an $\stp$ refinement of a $\tcsp$ in \cite{DechterMP91a} (see Figure 1 in \cite{DechterMP91a}). We apply the bdArc-Consistency algorithm
\mbox{bdAC-3} to $P$. The initial binarized domains of the variables and the initialization of the queue $Q$ are as follows:
\begin{enumerate}
  \item $Q=\{(X_1,X_2),(X_2,X_1),(X_2,X_3),(X_3,X_2),(X_3,X_4),$ $(X_4,X_3)\}$
  \item $
           \begin{array}{l||l|l|l|l}
           X_i       &X_1     &X_2                     &X_3                      &X_4\\  \hline
           bd(X_i)&[10,20]&\BBR&\BBR&[60,70]
           \end{array}
           $
\end{enumerate}
We take in turn the pairs in $Q$ for propagation. When a pair $(X_i,X_j)$, with $ij\not =0$, is taken from $Q$ for propagation,
we apply the path-consistency operation on triangle $(X_0,X_j,X_i)$: $P_{0i}=P_{0i}\cap P_{0j}\otimes P_{ji}$. If $P_{0i}$ is
modified, the pairs $(X_k,X_i)$ such that $k\not =0$, $k\not =i$, $k\not =j$ and $P$ has a constraint on $X_k$ and $X_i$, that
are no longer in $Q$, reenter the queue. \mbox{bdAC-3} will terminate when a binarized domain becomes empty, indicating
inconsistency of the initial $\stp$, or the queue becomes empty, indicating that the $\stp$ has been made bdArc-Consistent.

Step 1: we take the pair $(X_1,X_2)$ from $Q$ and apply the path-consistency operation on triangle $(X_0,X_2,X_1)$:
\begin{enumerate}
  \item $P_{01}=P_{01}\cap P_{02}\otimes P_{21}=[10,20]\cap \BBR\otimes [-40,-30]=[10,20]$
  \item No modification, and the new configuration is:
  \item[$\bullet$] $Q=\{(X_2,X_1),(X_2,X_3),(X_3,X_2),(X_3,X_4),(X_4,X_3)\}$
  \item[$\bullet$] $
           \begin{array}{l||l|l|l|l}
                X_i       &X_1     &X_2                     &X_3                      &X_4\\  \hline
                bd(X_i)&[10,20]&\BBR&\BBR&[60,70]
           \end{array}
           $
\end{enumerate}

Step 2: we take the pair $(X_2,X_1)$:
\begin{enumerate}
  \item $P_{02}=P_{02}\cap P_{01}\otimes P_{12}=\BBR\cap [10,20]\otimes [30,40]=[40,60]$
  \item There is modification, and the new configuration is:
  \item[$\bullet$] $Q=\{(X_2,X_3),(X_3,X_2),(X_3,X_4),(X_4,X_3)\}$
  \item[$\bullet$]
           $
           \begin{array}{l||l|l|l|l}
           X_i       &X_1     &X_2       &X_3                     &X_4\\  \hline
           bd(X_i)&[10,20]&[40,60]&\BBR&[60,70]
           \end{array}
           $
\end{enumerate}

Step 3: we take the pair $(X_2,X_3)$:
\begin{enumerate}
  \item $P_{02}=P_{02}\cap P_{03}\otimes P_{32}=[40,60]\cap\BBR\otimes [10,20]=[40,60]$
  \item No modification, and the new configuration is:
  \item[$\bullet$] $Q=\{(X_3,X_2),(X_3,X_4),(X_4,X_3)\}$
  \item[$\bullet$]
           $
           \begin{array}{l||l|l|l|l}
           X_i       &X_1     &X_2       &X_3                     &X_4\\  \hline
           bd(X_i)&[10,20]&[40,60]&\BBR&[60,70]
           \end{array}
           $
\end{enumerate}

Step 4: we take the pair $(X_3,X_2)$:
\begin{enumerate}
  \item $P_{03}=P_{03}\cap P_{02}\otimes P_{23}=\BBR\cap [40,60]\otimes [-20,-10]=[20,50]$
  \item There is modification, and the new configuration is:
  \item[$\bullet$] $Q=\{(X_3,X_4),(X_4,X_3)\}$
  \item[$\bullet$]
           $
           \begin{array}{l||l|l|l|l}
           X_i       &X_1     &X_2       &X_3      &X_4\\  \hline
           bd(X_i)&[10,20]&[40,60]&[20,50]&[60,70]
           \end{array}
           $
\end{enumerate}

Step 5: we take the pair $(X_3,X_4)$:
\begin{enumerate}
  \item $P_{03}=P_{03}\cap P_{04}\otimes P_{43}=[20,50]\cap [60,70]\otimes [-50,-40]=[20,50]\cap [10,30]=[20,30]$
  \item There is modification, and the new configuration is:
  \item[$\bullet$] The pair $(X_2,X_3)$ reenters to $Q$, which becomes $Q=\{(X_4,X_3),(X_2,X_3)\}$
  \item[$\bullet$]
           $
           \begin{array}{l||l|l|l|l}
           X_i       &X_1     &X_2       &X_3      &X_4\\  \hline
           bd(X_i)&[10,20]&[40,60]&[20,30]&[60,70]
           \end{array}
           $
\end{enumerate}

Step 6: we take the pair $(X_4,X_3)$:
\begin{enumerate}
  \item $P_{04}=P_{04}\cap P_{03}\otimes P_{34}=[60,70]\cap [20,30]\otimes [40,50]=[60,70]\cap [60,80]=[60,70]$
  \item No modification, and the new configuration is:
  \item[$\bullet$] $Q=\{(X_2,X_3)\}$
  \item[$\bullet$]
           $
           \begin{array}{l||l|l|l|l}
           X_i       &X_1     &X_2       &X_3      &X_4\\  \hline
           bd(X_i)&[10,20]&[40,60]&[20,30]&[60,70]
           \end{array}
           $
\end{enumerate}

Step 7: we take the pair $(X_2,X_3)$:
\begin{enumerate}
  \item $P_{02}=P_{02}\cap P_{03}\otimes P_{32}=[40,60]\cap [20,30]\otimes [10,20]=[40,60]\cap [30,50]=[40,50]$
  \item There is modification, and the new configuration is:
  \item[$\bullet$] The pair $(X_1,X_2)$ reenters to $Q$, which becomes $Q=\{(X_1,X_2)\}$
  \item[$\bullet$]
           $
           \begin{array}{l||l|l|l|l}
           X_i       &X_1     &X_2       &X_3      &X_4\\  \hline
           bd(X_i)&[10,20]&[40,50]&[20,30]&[60,70]
           \end{array}
           $
\end{enumerate}

Step 8: we take the pair $(X_1,X_2)$:
\begin{enumerate}
  \item $P_{01}=P_{01}\cap P_{02}\otimes P_{21}=[10,20]\cap [40,60]\otimes [-40,-30]=[10,20]\cap [0,30]=[10,20]$
  \item No modification, and the new configuration is:
  \item[$\bullet$] $Q=\{\}$
  \item[$\bullet$]
           $
           \begin{array}{l||l|l|l|l}
           X_i       &X_1     &X_2       &X_3      &X_4\\  \hline
           bd(X_i)&[10,20]&[40,50]&[20,30]&[60,70]
           \end{array}
           $
\end{enumerate}
The queue $Q$ is empty, and no (binarized) domain has become empty. \mbox{bdAC-3} terminates and the resulting $\stp$ is bdArc-Consistent.
The minimal (binarized) domains of variables $X_1,X_2,X_3,X_4$ are, respectively, $[10,20]$, $[40,50]$, $[20,30]$ and $[60,70]$. Therefore,
if we consider the distance graph of the original $\stp$, we have the following:
\begin{enumerate}
  \item the distances of the shortest paths from $X_0$ to $X_1,X_2,X_3,X_4$ are, respectively, $20$, $50$, $30$ and $70$
  \item the distances of the shortest paths from each of $X_1,X_2,X_3,X_4$ to $X_0$ are, respectively, $-10$, $-40$, $-20$ and $-60$. \cqfd
\end{enumerate}

\section{\mbox{bdAC-3} applied to an $\stp$: illustration 2}\label{appendix7}
In the $\stp$ of the example in Appendix \ref{appendix6}, we replace constraint $c_3$ with $(X_2-X_1)\in [30,+\infty )$, remove constraint $c_2$,
and add the constraint $(X_4-X_2)\in (-\infty ,4]$. We get the $\stp$ $P=(X,C)$, whose distance graph is
given by Figure \ref{stpdistancegraph}(Right):
\begin{enumerate}
  \item $X=\{X_0,X_1,X_2,X_3,X_4\}$ and
  \item $C=\{c_1:(X_1-X_0)\in [10,20],c_2:(X_2-X_1)\in [30,+\infty ),c_3:(X_3-X_2)\in [-20,-10],c_4:(X_4-X_2)\in (-\infty ,4],c_5:(X_4-X_3)\in [40,50]\}$.
\end{enumerate}
The rooted distance graph of $P$ does have a negative ciruit that
is not reachable from $X_0$, but from which $X_0$ is reachable. The proposed algorithm, contrary to Bellman-Ford-Moore's, will detect the
negative circuit; this amounts to detecting the inconsistency of the $\stp$.

We apply the bdArc-Consistency algorithm \mbox{bdAC-3} to $P$. The initial binarized domains of the variables and the initialization of the queue $Q$ are as follows:
\begin{enumerate}
  \item $Q=\{(X_1,X_2),(X_2,X_1),(X_2,X_3),(X_3,X_2),(X_2,X_4),$ $(X_4,X_2),(X_3,X_4),$ $(X_4,X_3)\}$
  \item $
           \begin{array}{l||l|l|l|l}
           X_i       &X_1     &X_2                     &X_3                      &X_4\\  \hline
           bd(X_i)&[10,20]&\BBR&\BBR&\BBR
           \end{array}
           $
\end{enumerate}
Step 1: we take the pair $(X_2,X_1)$:
\begin{enumerate}
  \item $P_{02}=P_{02}\cap P_{01}\otimes P_{12}=\BBR\cap [10,20]\otimes [30,+\infty )=[40,+\infty )$
  \item There is modification, and the new configuration is:
  \item[$\bullet$] $Q=\{(X_1,X_2),(X_2,X_3),(X_3,X_2),(X_2,X_4),$ $(X_4,X_2),(X_3,X_4),$ $(X_4,X_3)\}$
  \item[$\bullet$] $
           \begin{array}{l||l|l|l|l}
           X_i       &X_1     &X_2                     &X_3               &X_4\\  \hline
           bd(X_i)&[10,20]&[40,+\infty )&\BBR&\BBR
           \end{array}
           $
\end{enumerate}

Step 2: we take the pair $(X_2,X_3)$:
\begin{enumerate}
  \item $P_{02}=P_{02}\cap P_{03}\otimes P_{32}=[40,+\infty )\cap\BBR\otimes [10,20]=[40,+\infty )$
  \item No modification, and the new configuration is:
  \item[$\bullet$] $Q=\{(X_1,X_2),(X_3,X_2),(X_2,X_4),$ $(X_4,X_2),(X_3,X_4),$ $(X_4,X_3)\}$
  \item[$\bullet$] $
           \begin{array}{l||l|l|l|l}
           X_i       &X_1     &X_2                     &X_3               &X_4\\  \hline
           bd(X_i)&[10,20]&[40,+\infty )&\BBR&\BBR
           \end{array}
           $
\end{enumerate}

Step 3: we take the pair $(X_3,X_2)$:
\begin{enumerate}
  \item $P_{03}=P_{03}\cap P_{02}\otimes P_{23}=\BBR\cap [40,+\infty )\otimes [-20,-10]=[20,+\infty )$
  \item There is modification, and the new configuration is:
  \item[$\bullet$] $Q=\{(X_1,X_2),(X_2,X_4),$ $(X_4,X_2),(X_3,X_4),$ $(X_4,X_3)\}$
  \item[$\bullet$] $
           \begin{array}{l||l|l|l|l}
           X_i       &X_1     &X_2                     &X_3        &X_4\\  \hline
           bd(X_i)&[10,20]&[40,+\infty )&[20,+\infty )&\BBR
           \end{array}
           $
\end{enumerate}

Step 4: we take the pair $(X_2,X_4)$:
\begin{enumerate}
  \item $P_{02}=P_{02}\cap P_{04}\otimes P_{42}=[40,+\infty )\cap\BBR\otimes [-4,+\infty )=[40,+\infty )$
  \item No modification, and the new configuration is:
  \item[$\bullet$] $Q=\{(X_1,X_2),(X_4,X_2),(X_3,X_4),$ $(X_4,X_3)\}$
  \item[$\bullet$] $
           \begin{array}{l||l|l|l|l}
           X_i       &X_1     &X_2                     &X_3        &X_4\\  \hline
           bd(X_i)&[10,20]&[40,+\infty )&[20,+\infty )&\BBR
           \end{array}
           $
\end{enumerate}

Step 5: we take the pair $(X_4,X_2)$:
\begin{enumerate}
  \item $P_{04}=P_{04}\cap P_{02}\otimes P_{24}=\BBR\cap [40,+\infty )\otimes (-\infty ,4]=\BBR$
  \item No modification, and the new configuration is:
  \item[$\bullet$] $Q=\{(X_1,X_2),(X_3,X_4),$ $(X_4,X_3)\}$
  \item[$\bullet$] $
           \begin{array}{l||l|l|l|l}
           X_i       &X_1     &X_2                     &X_3        &X_4\\  \hline
           bd(X_i)&[10,20]&[40,+\infty )&[20,+\infty )&\BBR
           \end{array}
           $
\end{enumerate}

Step 6: we take the pair $(X_3,X_4)$:
\begin{enumerate}
  \item $P_{03}=P_{03}\cap P_{04}\otimes P_{43}=[20,+\infty )\cap\BBR\otimes [-50,-40]=[20,+\infty )$
  \item No modification, and the new configuration is:
  \item[$\bullet$] $Q=\{(X_1,X_2),(X_4,X_3)\}$
  \item[$\bullet$] $
           \begin{array}{l||l|l|l|l}
           X_i       &X_1     &X_2                     &X_3        &X_4\\  \hline
           bd(X_i)&[10,20]&[40,+\infty )&[20,+\infty )&\BBR
           \end{array}
           $
\end{enumerate}

Step 7: we take the pair $(X_4,X_3)$:
\begin{enumerate}
  \item $P_{04}=P_{04}\cap P_{03}\otimes P_{34}=\BBR\cap [20,+\infty )\otimes [40,50]=[60,+\infty )$
  \item There is modification. $(X_2,X_4)$ reenters the queue, and the new configuration is:
  \item[$\bullet$] $Q=\{(X_1,X_2),(X_2,X_4)\}$
  \item[$\bullet$]
           $
           \begin{array}{l||l|l|l|l}
           X_i       &X_1     &X_2       &X_3      &X_4\\  \hline
           bd(X_i)&[10,20]&[40,+\infty )&[20,+\infty )&[60,+\infty )
           \end{array}
           $
\end{enumerate}

Step 8: we take the pair $(X_2,X_4)$:
\begin{enumerate}
  \item $P_{02}=P_{02}\cap P_{04}\otimes P_{42}=[40,+\infty )\cap [60,+\infty )\otimes [-4,+\infty )=[56,+\infty )$
  \item There is modification, and the new configuration is:
  \item[$\bullet$] The pair $(X_3,X_2)$ reenters the queue $Q$, which becomes $Q=\{(X_1,X_2),$ $(X_3,X_2)\}$
  \item[$\bullet$]
           $
           \begin{array}{l||l|l|l|l}
           X_i       &X_1     &X_2       &X_3      &X_4\\  \hline
           bd(X_i)&[10,20]&[56,+\infty )&[20,+\infty )&[60,+\infty )
           \end{array}
           $
\end{enumerate}

Step 9: we take the pair $(X_3,X_2)$:
\begin{enumerate}
  \item $P_{03}=P_{03}\cap P_{02}\otimes P_{23}=[20,+\infty )\cap [56,+\infty )\otimes [-20,-10]=[36,+\infty )$
  \item There is modification, and the new configuration is:
  \item[$\bullet$] The pair $(X_4,X_3)$ reenters the $Q$, which becomes $Q=\{(X_1,X_2),(X_4,X_3)\}$
  \item[$\bullet$]
           $
           \begin{array}{l||l|l|l|l}
           X_i       &X_1     &X_2       &X_3      &X_4\\  \hline
           bd(X_i)&[10,20]&[56,+\infty )&[36,+\infty )&[60,+\infty )
           \end{array}
           $
\end{enumerate}

Step 10: we take the pair $(X_4,X_3)$:
\begin{enumerate}
  \item $P_{04}=P_{04}\cap P_{03}\otimes P_{34}=[60,+\infty )\cap [36,+\infty )\otimes [40,50]=[76,+\infty )$
  \item There is modification, and new configuration is:
  \item[$\bullet$] The pair $(X_2,X_4)$ reenters the $Q$, which becomes $Q=\{(X_1,X_2),(X_2,X_4)\}$
  \item[$\bullet$]
           $
           \begin{array}{l||l|l|l|l}
           X_i       &X_1     &X_2       &X_3      &X_4\\  \hline
           bd(X_i)&[10,20]&[56,+\infty )&[36,+\infty )&[76,+\infty )
           \end{array}
           $
\end{enumerate}

Step 11: we take the pair $(X_2,X_4)$:
\begin{enumerate}
  \item $P_{02}=P_{02}\cap P_{04}\otimes P_{42}=[56,+\infty )\cap [76,+\infty )\otimes [-4,+\infty )=[72,+\infty )$
  \item There is modification, and the new configuration is:
  \item[$\bullet$] The pair $(X_3,X_2)$ reenters the queue $Q$, which becomes $Q=\{(X_1,X_2),$ $(X_3,X_2)\}$
  \item[$\bullet$]
           $
           \begin{array}{l||l|l|l|l}
           X_i       &X_1     &X_2       &X_3      &X_4\\  \hline
           bd(X_i)&[10,20]&[72,+\infty )&[36,+\infty )&[76,+\infty )
           \end{array}
           $
\end{enumerate}

Step 12: we take the pair $(X_3,X_2)$:
\begin{enumerate}
  \item $P_{03}=P_{03}\cap P_{02}\otimes P_{23}=[36,+\infty )\cap [72,+\infty )\otimes [-20,-10]=[52,+\infty )$
  \item There is modification, and the new configuration is:
  \item[$\bullet$] The pair $(X_4,X_3)$ reenters the queue $Q$, which becomes $Q=\{(X_1,X_2),$ $(X_4,X_3)\}$
  \item[$\bullet$]
           $
           \begin{array}{l||l|l|l|l}
           X_i       &X_1     &X_2       &X_3      &X_4\\  \hline
           bd(X_i)&[10,20]&[72,+\infty )&[52,+\infty )&[76,+\infty )
           \end{array}
           $
\end{enumerate}

Step 13: we take the pair $(X_4,X_3)$:
\begin{enumerate}
  \item $P_{04}=P_{04}\cap P_{03}\otimes P_{34}=[76,+\infty )\cap [52,+\infty )\otimes [40,50]=[92,+\infty )$
  \item There is modification of $P_{04}$ with $-lowerB(P_{04})=-92$ strictly smaller than $\mbox{path-lb}(P)=-90$.
	The conditions of Theorem \ref{circuitfreepath} get violated, and the situation corresponds to a detection of a negative circuit. The
	algorithm terminates with a negative answer to the consistency problem of $P$ (line \ref{revise42} of the REVISE procedure of Figure \ref{TheREVISEProcedure}). \cqfd
\end{enumerate}
\end{document}